\documentclass[10pt,twocolumn,letterpaper]{article}

\usepackage[pagenumbers]{cvpr}
\providecommand{\linenumbers}{}\providecommand{\nolinenumbers}{}      %

\usepackage{array}
\usepackage{float}
\usepackage{colortbl}
\usepackage{multirow}
\definecolor{oursband}{HTML}{EAF3F9}
\definecolor{groupband}{HTML}{D1E5F0}
\definecolor{groupink}{HTML}{053061}
\newcommand{\best}[1]{\textbf{#1}}
\newcommand{\second}[1]{\underline{#1}}
\newcommand{\oursrow}{\rowcolor{oursband}}
\newcommand{\groupband}[2]{\rowcolor{groupband}\multicolumn{#1}{@{}l}{\textcolor{groupink}{\textbf{#2}}}\\}

\usepackage{iftex}
\ifXeTeX
  \usepackage{fontspec}
\fi

\definecolor{cvprblue}{rgb}{0.21,0.49,0.74}
\usepackage{bibunits}  %
\usepackage[pagebackref,breaklinks,colorlinks,allcolors=cvprblue]{hyperref}
\title{CADENCE: Sensor-Grounded World-Consistent Gaussian SLAM\\for Collaborative Mobile Reconstruction}

\author{Reshad Ul Karim$^{1}$\qquad Abrar Samin$^{1}$\qquad Md Rashad Al Hasan Rony$^{2}$\\
$^{1}$BRAC University, Dhaka, Bangladesh\qquad $^{2}$AI Competence Center, Deutsche Telekom AG\\
{\tt\small reshad.ul.karim@g.bracu.ac.bd\quad abrar.samin@g.bracu.ac.bd\quad rashad.research@gmail.com}
}

\begin{document}
\maketitle
\begin{abstract}
Collaborative reconstruction with consumer phones promises faster and more complete indoor maps. However, heterogeneous devices introduce degraded observations and unreliable cross-device associations, so a single poor contributor or incorrect match can corrupt the shared Gaussian map. To address this, we propose \textbf{CADENCE}, a collaborative Gaussian simultaneous localization and mapping (SLAM) system that separates finding cross-device matches from deciding whether and how peer observations update the map.
To find matches, a sensor-grounded regional memory summarizes committed observations and ranks candidate peer regions using appearance, relative pose, gravity, barometric altitude, and observation quality.
Deterministic geometric verification and optional physical checks then decide which of the proposed matches are accepted.
Once a match is accepted, separate cue-based confidences control how its observations contribute to pose constraints, geometry, appearance, and memory.
As a result, useful depth can still contribute even when color is unreliable.
Throughout this process, local tracking remains isolated from mapping, verification, and networking. We evaluate CADENCE on eight Matterport3D houses that were not used to train the memory. On these houses, regional memory achieves an area under the receiver operating characteristic curve (AUROC) of $0.95$ to $0.97$ and thus outperforms appearance retrieval, while verification keeps false-link rates below $0.5\,\%$. Furthermore, soft confidence weighting reduces the mean receiver peak signal-to-noise ratio (PSNR) loss from $2.66$ to $1.05$\,dB across eight degradations while retaining useful peer depth. Finally, using inertial and platform motion, CADENCE tracks all twelve held-out iPhone scans with a mean absolute trajectory error of $6.98$\,cm, and its refined maps achieve better average rendering quality than the baselines.

\end{abstract}

\suppressfloats[t]%
\begin{figure}[t]
  \centering
  \includegraphics[width=\linewidth]{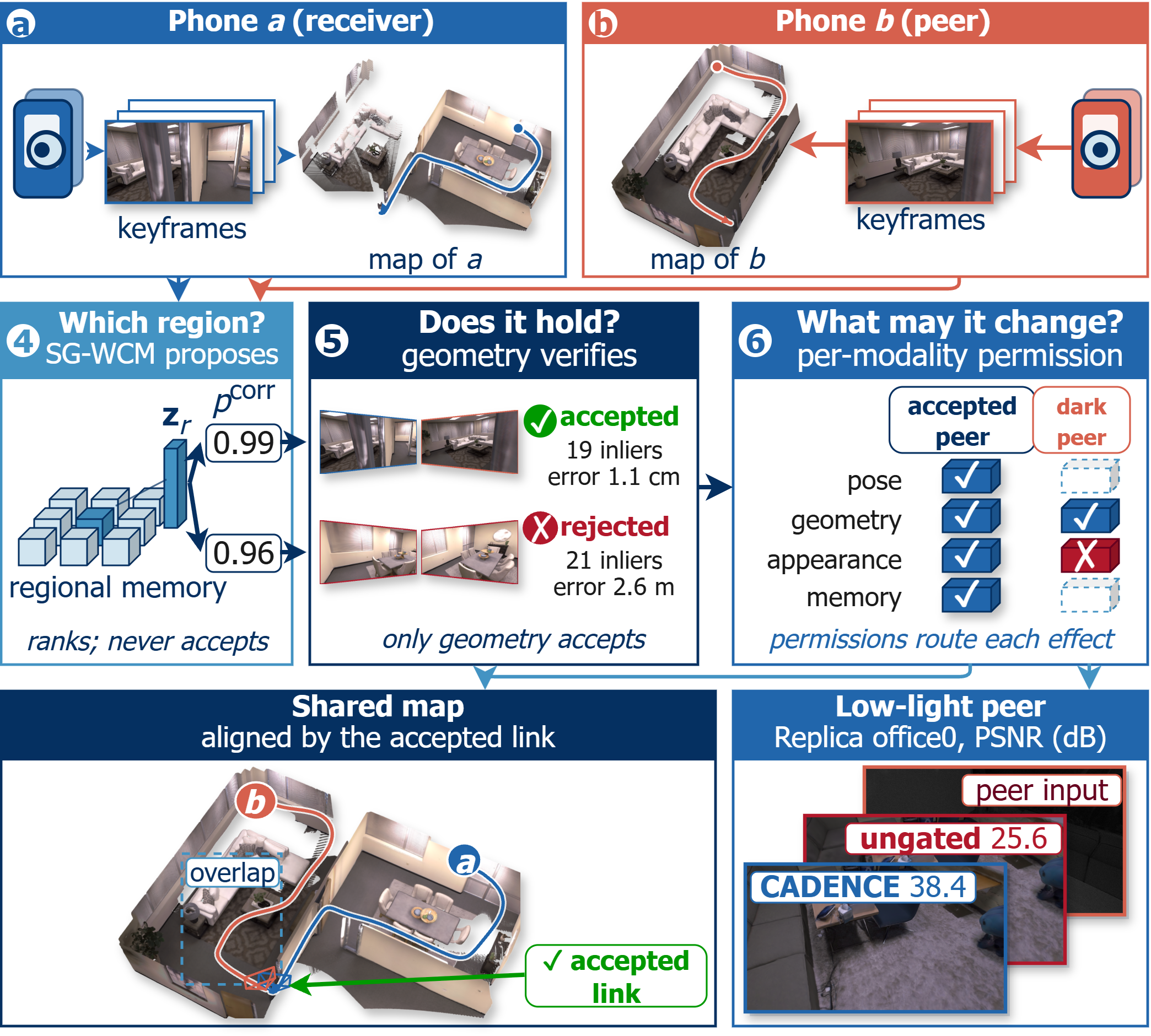}
  \caption{Three separate decisions govern what one phone takes from another. As two phones map one apartment, the sensor-grounded memory (SG-WCM) ranks candidate regions, only geometric verification accepts a relation, and per-modality permissions decide what an accepted peer may change: a low-light peer still contributes geometry while its appearance is withheld.}
  \label{fig:teaser}
\end{figure}

\section{Introduction}
\label{sec:intro}

When several phones scan the same environment, they can cover more areas than a single phone. However, different phones may capture different lighting conditions, exposure levels, motion blur, depth quality, and tracking reliability. As a result, one phone may correctly recognize a place but still provide unreliable data that reduces the quality of another phone's map. A collaborative system must therefore do two things: identify matching places and decide which information from another phone is reliable enough to add to the map.

3D Gaussian Splatting (3DGS) can create detailed 3D scene representations~\cite{kerbl20233d}. It has also been used in dense simultaneous localization and mapping (SLAM) systems~\cite{matsuki2024gaussian}. Similar approaches include SplaTAM~\cite{keetha2024splatam} and PhotoSLAM~\cite{huang2024photoslam}. Collaborative Gaussian SLAM systems combine observations from multiple devices using appearance matching and geometric or photometric registration. Examples include the method presented in~\cite{yugay2025magicslam, xu2025macego3d}. However, finding the correct place does not mean that every type of information in the matched observation is reliable. Research on collaborative perception also shows that differences between contributing devices can reduce the quality of a shared result~\cite{gao2025stamp}. Therefore, a collaborative reconstruction system must check how useful each type of incoming evidence is for the receiving map, while allowing the phone to continue tracking its own movement.

CADENCE addresses this problem by separating three steps: ranking possible matches, checking whether a match is geometrically valid, and deciding which matched information can update the map (Fig.~\ref{fig:teaser}). It uses a sensor-grounded world-consistency memory (SG-WCM) to summarize committed observations within spatial regions. This memory ranks candidate regions from other devices using appearance, relative pose, gravity, barometric altitude, and observation quality. Deterministic geometric tests decide whether a proposed match can be accepted, with optional physical checks providing additional verification. Separate confidence values then control how accepted peer evidence affects pose constraints, geometry, appearance, and memory. The explicit map remains the main source of truth, while local tracking proceeds independently of background collaboration. On unseen Matterport3D~\cite{chang2017matterport3d} houses, the regional memory improves region ranking over appearance retrieval, achieving an area under the receiver operating characteristic curve (AUROC) of $0.95$ to $0.97$, while geometric verification limits false associations. Soft confidence weighting removes the $3.73$\,dB own-view peak signal-to-noise ratio (PSNR) loss from a low-light peer while retaining its depth, and live iPhone sessions demonstrate collaboration with low tracking latency. The contributions of this paper are summarized as follows:
\begin{enumerate}
\item A sensor-grounded regional memory that ranks cross-device regions from committed appearance and sensor evidence, with capped embedding storage and single-stage training.

\item Collaborative fusion that verifies memory-ranked cross-device matches using feature correspondences and available depth before admitting peer updates, keeping false-link rates below $0.5\,\%$ on unseen Matterport3D houses.

\item Confidence-governed sharing that controls peer updates to pose constraints, geometry, appearance, and memory; soft weighting reduces mean receiver PSNR loss from $2.66$ to $1.05$\,dB across eight degradations.

\item A native mobile implementation that isolates tracking from background collaboration and validates accelerated operators against CPU references, maintaining a 95th-percentile pose time below $33.3$\,ms in live iPhone sessions.
\end{enumerate}

\section{Related Work}
\label{sec:related}

ORB-SLAM3~\cite{campos2021orb} provides feature-based visual-inertial localization, while iMAP~\cite{sucar2021imap} and Point-SLAM~\cite{sandstrom2023point} learn dense neural scene representations.
3D Gaussian Splatting~\cite{kerbl20233d} offers an explicit representation for dense mapping.
MonoGS~\cite{matsuki2024gaussian} and SplaTAM~\cite{keetha2024splatam} use photometric tracking, GS-ICP SLAM~\cite{ha2024rgbd} aligns depth geometrically, and Photo-SLAM~\cite{huang2024photoslam} runs sparse feature tracking alongside mapping.
Gaussian-SLAM~\cite{yugay2023gaussianslam} adds submaps, LoopSplat~\cite{zhu2025loopsplat} adds loop closure, and MM3DGS SLAM~\cite{sun2024mm3dgs} and VIGS-SLAM~\cite{zhu2026vigsslam} incorporate inertial constraints.
Extending these maps across agents requires place matching and alignment.
Sparse systems such as COVINS~\cite{schmuck2021covins} and Kimera-Multi~\cite{tian2022kimeramulti} exchange place descriptors and reject inconsistent inter-agent relations, including through pairwise consistency maximization~\cite{mangelson2018pairwise}.
Swarm-SLAM~\cite{lajoie2024swarmslam} prioritizes verification under a communication budget, while AnyLoc~\cite{keetha2024anyloc} uses frozen foundation features for place recognition.
Dense systems share neural points in CP-SLAM~\cite{hu2023cpslam} or Gaussians in MAC-Ego3D~\cite{xu2025macego3d} and OminiMAG-SLAM~\cite{ding2026ominimagslam}.
The closest systems, MAGiC-SLAM~\cite{yugay2025magicslam}, CoMA-SLAM~\cite{chen2026comaslam}, GRAND-SLAM~\cite{thomas2025grandslam} and MAGS-SLAM~\cite{cao2026magsslam}, retrieve candidates by appearance, verify them geometrically or photometrically, and align submaps.
CADENCE follows this structure while ranking regions using committed sensor evidence.
Its association study varies only the proposal policy, keeping the tracker, mapper and verifier fixed.

Even after a match is accepted, some shared observations may still fail to improve the receiving map.
Collaborative perception addresses this reliability problem: Where2comm~\cite{hu2022where2comm} weights shared features by spatial confidence, STAMP~\cite{gao2025stamp} studies compatibility between heterogeneous contributors, and consensus checks identify adversarial teammates~\cite{li2023amongus}.
In mapping, DGOMapping~\cite{li2026dgomapping} weights observations by quality, visibility and cross-agent depth consistency, while CG-SLAM~\cite{hu2024cgslam} reweights depth using per-pixel uncertainty.
CADENCE gives each modality of an accepted peer contribution a separate permission and measures its effect on the receiving Gaussian map, including negative transfer.
These peer-update decisions operate outside the pose path, allowing tracking to continue independently of collaboration.
Mobile deployment adds constraints on computation and depth sensing.
CGS-SLAM~\cite{deambrogi2026cgsslam} evaluates client reconstruction on a laptop graphics processing unit (GPU) and offloads metric depth estimation to a server.
CADENCE uses sensor depth and runs tracking and mapping natively on Apple devices.

\section{Method}
\label{sec:method}

\subsection{Problem Formulation}
\label{sec:method_overview}
\begin{figure*}[t]
\centering
\includegraphics[width=\textwidth]{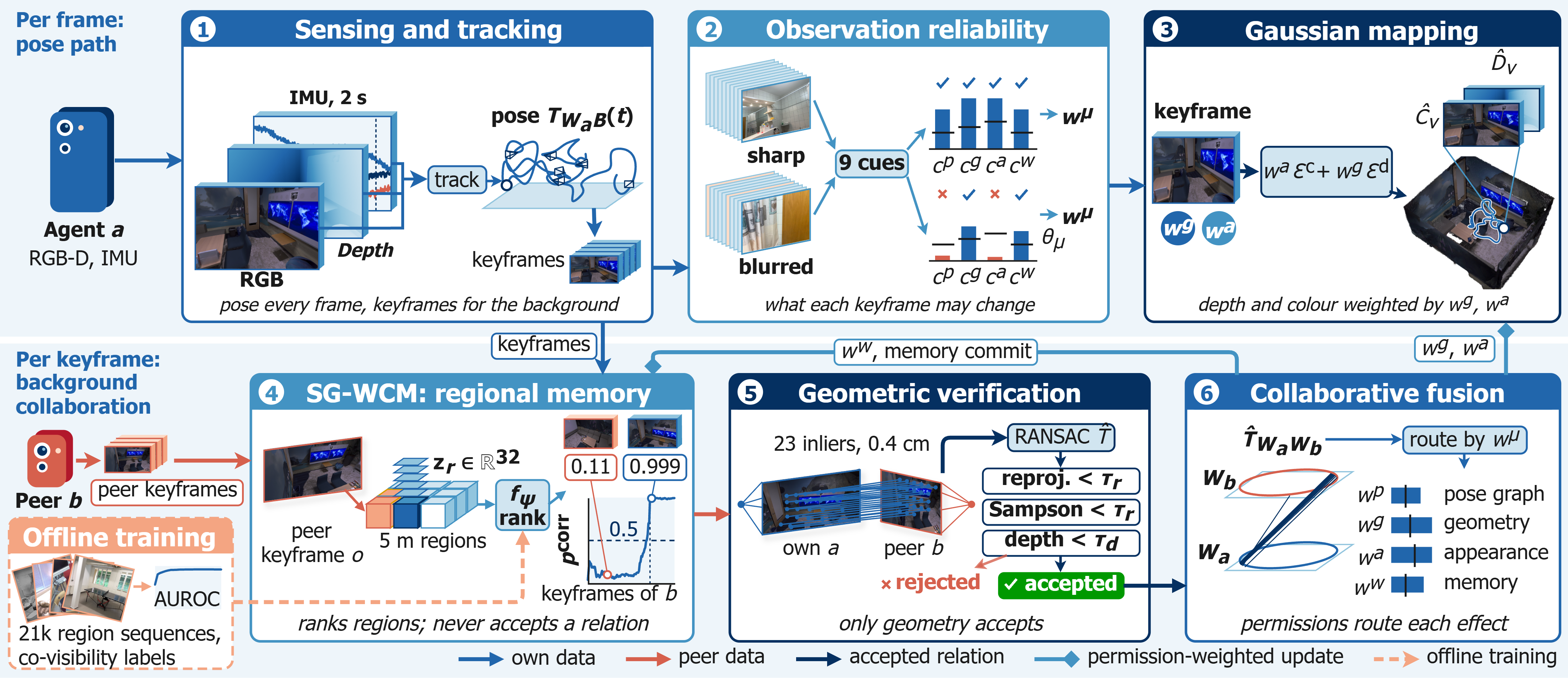}
\caption{Architecture of CADENCE. Each agent tracks its observations and builds a local Gaussian map, while regional memory, geometric verification and modality permissions govern collaboration. Dashed lines indicate offline training of $f_\psi$.}
\label{fig:method_overview}
\end{figure*}
CADENCE reconstructs a static indoor scene from calibrated devices with separate clocks, world frames and Gaussian maps.
Agent $a\in\mathcal A$ receives images and available inertial measurement unit (IMU), sensor-depth, barometric and exposure measurements.
The agent estimates a body trajectory $T_{W_aB}(t)$ and a local Gaussian map, where $B$ is the IMU frame and $T_{XY}$ maps frame $Y$ into frame $X$.
We assume known intrinsics and camera-to-IMU extrinsics, sensor-based metric geometry and cooperative peers that may contribute degraded observations.

The explicit world $\mathcal W=\{\mathcal G,\mathcal P,\mathcal A,\mathcal F\}$ contains Gaussians, a pose and submap graph, agents and optional floor hypotheses.
SG-WCM summarizes committed observations in memory $Z_W$ and ranks candidates, geometric verification decides acceptance, and modality permissions control updates.
Only the owning agent modifies its map.

\subsection{Reliability-Aware Local Reconstruction}
\label{sec:method_local}
\paragraph{Tracking and Keyframes.}
Sparse feature tracking~\cite{rosten2006machine,lucas1981iterative} refines an available preintegrated IMU prediction~\cite{forster2017manifold} through robust reprojection optimization.
Profiles with dense depth registration align against a separate surfel model~\cite{keller2013real} instead of the Gaussians.
Fixed-gain filters correct velocity and biases.
Without IMU data, the tracker omits the rotation prior, and recorded phones may use gated ARKit motion.
Keyframes carry pose, image, depth and quality to bounded background queues (\cref{sec:supp_local}).

\paragraph{Observation Reliability.}
\label{sec:method_reliability}
Nine cues $\mathbf y\in[0,1]^9$ characterize image quality, motion, texture, viewpoint, depth and pose reliability~\cite{crete2007blur,immerkaer1996fast}.
Their combinations yield confidences $c^\mu$ for pose, geometry, appearance and memory, $\mu\in\{p,g,a,w\}$ (Eq.~\ref{eq:method_conf}), and sensed depth can thus support geometry when color is poor.
The guard sets the permission weight $w^\mu=c^\mu\,\mathbf1[c^\mu\ge\theta_\mu]$ to zero below threshold and to the confidence otherwise.
A soft policy instead weights every update of a contribution by $\min_\mu c^\mu$.
\Cref{sec:supp_reliability,sec:exp_setup} specify the cue computation and thresholds.

\paragraph{Gaussian Mapping.}
\label{sec:method_mapping}
The transform $T_{W_aS_k}$ places each Gaussian submap $S_k$, and graph corrections change this transform without refitting the submap.
Sensed depth seeds provisional Gaussians where the render leaves pixels unexplained, and another keyframe must confirm their depth before verification.
Rendering views $\mathcal V$ yields color $\hat C_v$, opacity $\hat A_v$ and depth $\hat D_v$ for the objective
\begin{equation}
\label{eq:method_loss}
\mathcal L=\sum_{v\in\mathcal V}\Big[w^{a}_v\big((1{-}\lambda)\,\mathcal E^{c}_v+\lambda\,(1{-}\mathrm{SSIM}_v)\big)+w^{g}_v\,\mathcal E^{d}_v\Big].
\end{equation}
Here, $\mathcal E^c_v$ and $\mathcal E^d_v$ are mean squared color and depth errors, the depth term uses sensed pixels with $\hat A_v>0.3$, and $\lambda$ weights the structural similarity index measure (SSIM)~\cite{wang2004image}.
Independent weights allow depth to contribute while color is withheld, and these weights leave own-keyframe seeding and validation unchanged.
Sparse Adam updates affected Gaussians, and \cref{sec:exp_setup,sec:supp_mapping} describe the scheduling and offline refinement.

\subsection{Sensor-Grounded Regional Memory}
\label{sec:method_wcm}
Each cubic region $r$ stores a token $\mathbf z_r\in\mathbb R^{32}$, an evidence summary and a capped cache of frozen DINOv3 embeddings~\cite{simeoni2025dinov3}.
For observation $o$, the query $\mathbf x(o,r)\in\mathbb R^{95}$ combines this token with region-relative pose, quality, confidences, sensor evidence, availability masks and embedding-similarity statistics.
The region circumsphere radius normalizes position offsets, and the adopted model zeroes its inertial-increment block (\cref{sec:supp_wcm}).
For committed observations $\mathcal C_r$ with $n_r=|\mathcal C_r|>0$, the relative sensor evidence is
\begin{equation}
\label{eq:method_sensor_relative}
\begin{aligned}
\bar{\boldsymbol\gamma}_r&=\frac1{n_r}\sum_{j\in\mathcal C_r}\boldsymbol\gamma_j,
&\bar h_r&=\frac1{n_r}\sum_{j\in\mathcal C_r}h_j,\\
\mathbf e^{\mathrm{ga}}_{or}&=\big[h_o-\bar h_r,\ \sigma_{h,o},\ (\boldsymbol\gamma_o-\bar{\boldsymbol\gamma}_r)^\top\big]^\top,
\end{aligned}
\end{equation}
where $\boldsymbol\gamma_o$ is body-frame gravity divided by $9.81$\,m/s$^2$, $h_o$ is barometric altitude and $\sigma_{h,o}$ is its uncertainty.
The gravity difference encodes device tilt, and both differences are zero for an empty region.
A perceptron with two 128-unit rectified linear unit (ReLU) layers predicts logits $\boldsymbol\ell$ and an increment $\boldsymbol\delta$ as
\begin{equation}
\label{eq:method_wcm}
\begin{aligned}
(\boldsymbol\ell,\boldsymbol\delta)&=f_\psi\big(\mathbf x(o,r)\big),\\
\mathbf z_r&\leftarrow\operatorname{clip}\big(\mathbf z_r+\tanh\boldsymbol\delta,\,-4,\,4\big).
\end{aligned}
\end{equation}
At run time, CADENCE uses the correspondence head $p^{\mathrm{corr}}$, the floor-compatibility head $p^{\mathrm{floor}}$ and the peer-reliability head $p^{\mathrm{rel}}$, and \cref{sec:supp_wcm} lists auxiliary heads.
A query alone changes no regional state.
Initial own observations commit directly, whereas peer increments stay cached until verification and memory permission.
Only committed evidence enters the token, cache and sensor means, and an empty region returns no prediction.
\Cref{sec:supp_wcm} specifies the capacity and stale-increment rules of the memory.

\subsection{Verified Collaborative Fusion}
\label{sec:method_fusion}
\paragraph{Candidate Selection.}
Asynchronous relations support place recognition and frame and submap alignment rather than simultaneous stereo.
Announcements carry embeddings, sender-frame poses and quality and sensor summaries, whereas maps and tokens remain local.
The receiver retrieves keyframes by embedding distance $d_{ok}$ and, after alignment, adds candidates near the transformed peer pose.
For candidate $k$ in region $r_k$, the receiver computes
\begin{equation}
\label{eq:method_rank}
\eta_{ok}=p^{\mathrm{corr}}(o,r_k)\,p^{\mathrm{floor}}(o,r_k),\qquad p^{\mathrm{corr}}(o,r_k)\ge0.5,
\end{equation}
and breaks ties by embedding distance.
Thus, memory ranks regions and retrieval orders their keyframes.
Before alignment, the receiver masks peer pose, and distance screening handles first contact independently of memory.
Descriptor screening provides a fallback without embeddings, and a fixed exploration fraction bypasses the correspondence cutoff (\cref{sec:exp_setup,sec:supp_fusion}).

\paragraph{Verification.}
One evidence packet, reused across candidates, contains a thumbnail, sparse depth and binary robust independent elementary features (BRIEF)~\cite{calonder2010brief}.
Mutual Hamming matches produce a metric rigid proposal through three-point random sample consensus (RANSAC)~\cite{fischler1981random,umeyama1991least} and Perspective-n-Point (PnP) refinement.
The verifier tests re-tracked points and falls back to all mutual matches when support is insufficient
\begin{equation}
\label{eq:method_verify}
\begin{aligned}
&\operatorname*{med}_i\big\|\pi(\hat T\mathbf X_i)-\mathbf u_i\big\|<\tau_r,\qquad
\operatorname*{med}_i d_{\mathrm S}(\mathbf u_i,\mathbf u'_i)<\tau_r,\\
&\operatorname*{med}_j|z_j-\hat z_j|<\tau_d.
\end{aligned}
\end{equation}
Here, $\pi$ projects transformed 3D matches $\mathbf X_i$ to pixels $\mathbf u_i$, $d_{\mathrm S}$ is the square-root Sampson error~\cite{hartley2004multiple} between pixels $\mathbf u_i,\mathbf u'_i$, and $z_j,\hat z_j$ are measured and predicted unoccluded depths.
A failed test or insufficient pose confidence rejects a relation, and missing geometry defers the decision. In contrast, a relation can still be accepted when a test is merely unavailable.
Optional physical checks bound gravity tilt, barometric height disagreement under a shared reference and inconsistency with prior alignments. Memory-based deferral is off in reported runs and cannot accept a relation.
\Cref{sec:exp_setup,sec:supp_fusion} fix all thresholds and support requirements.

\paragraph{Multi-View Evidence.}
If either agent lacks usable metric depth, the system falls back to a monocular path.
In this case, it triangulates features at the tracked poses and filters them by reprojection error.
For matching, this path uses scale-invariant feature transform (SIFT) keypoints~\cite{lowe2004distinctive} with hashed RootSIFT descriptors~\cite{arandjelovic2012three}, which are exchanged in the same binary packet format.
Because metric depth is unavailable, a consistency test replaces the depth test.
Specifically, the forward and reverse pose estimates must agree in one direction, while reprojection verification remains in place.
To further stabilize the estimate, the path also applies a stronger gyroscope prior (\cref{sec:supp_round2_settings}).
Finally, since the triangulated depths inherit the scale of the tracked translation, the system trusts monocular translation only once that scale is metric (\cref{sec:supp_local}).

\paragraph{Alignment and Permitted Updates.}
An accepted relative body pose $\hat T_{B_iB_j}$ between keyframe $i$ of agent $a$ and keyframe $j$ of agent $b$ yields
\begin{equation}
\label{eq:method_alignment}
\hat T_{W_aW_b}=T_{W_aB_i}\,\hat T_{B_iB_j}\,T_{W_bB_j}^{-1}.
\end{equation}
The reported confidence $\tilde{\mathbf c}_o$, capped in pose and geometry by $p^{\mathrm{rel}}$ when active, updates the running summary of peer $b$ as
\begin{equation}
\label{eq:method_peer_conf}
\begin{aligned}
\bar{\mathbf c}_b&\leftarrow(1-\nu)\bar{\mathbf c}_b+\nu\tilde{\mathbf c}_o,\\
\mathbf c^{\mathrm{eff}}_{b,o}&=\min\big(\bar{\mathbf c}_b,\tilde{\mathbf c}_o\big),
\end{aligned}
\end{equation}
where the minimum operates elementwise and thresholding yields the permission weights.
By default, a missing pose permission prevents verification, whereas the verify-first phone variant reverses this order.
For accepted relations, $w^p$ weights the graph factor, and graph corrections align frames and submap origins rather than stored trajectories.
Merge inserts verified peer points in gray with opacity scaled by $w^g$, and own views can color them.
Share additionally uses aligned peer views with appearance weights scaled by $w^a$, and memory commits require $w^w$.

\subsection{Offline Training}
\label{sec:method_training}
We jointly train $f_\psi$ and four loss log-variances in one offline stage with DINOv3 frozen.
Each sequence commits the regional visit of one pseudo-agent and then queries positives and hard or random negatives under simulated alignments and modality dropout.
A query is positive when at least $30\,\%$ of its $5$\,cm depth voxels occur in the memory visit, negative at no more than $5\,\%$, and excluded otherwise.
Co-visibility, level identity and tracking accuracy supervise the heads (\cref{sec:supp_wcm}).
Learned log-variances $s_k$~\cite{kendall2018multi} balance the loss as
\begin{equation}
\label{eq:method_wcm_loss}
\mathcal L_\psi=\sum_{k\in\{\mathrm{corr},\mathrm{vis},\mathrm{floor},\mathrm{rel}\}}\big(e^{-s_k}\,\overline{\mathrm{BCE}}_k+s_k\big)+\tfrac12\,\mathcal L_{\mathrm{err}},
\end{equation}
where $\overline{\mathrm{BCE}}_k$ averages binary cross-entropy over labeled steps and $\mathcal L_{\mathrm{err}}$ supervises correspondence-decision errors at threshold $0.5$.
Gradients pass through token updates without token targets.
Labels govern training commits, whereas inference uses initial own commits and verified, permitted peer commits.
Weights stay fixed at inference, own queries are event-driven, and peer announcements trigger scoring.

\subsection{Tracking Isolation}
\label{sec:method_runtime}
The pose call neither accesses Gaussians nor invokes mapping, memory inference, verification, graph optimization or networking.
Full queues evict or refuse jobs instead of waiting.
Shared hardware can nevertheless increase the pose-call duration $\tau_t$, which we assess against the contract
\begin{equation}
\label{eq:method_contract}
\operatorname{p95}_t(\tau_t)\le\tau^\star=33.3\,\text{ms},
\end{equation}
where $\tau^\star$ equals the frame period of $30$\,Hz input at every input rate.
Under resource pressure, a controller reduces background work without changing tracker settings or verification thresholds.
Deterministic execution fixes scheduling, and paced execution uses idle time (\cref{sec:exp_setup,sec:supp_backend,sec:supp_modes}).
\label{sec:method_end}

\section{Experiments and Results}
\label{sec:exp}
\begin{figure}[t]

\centering
\captionsetup[sub]{labelformat=parens,font=footnotesize,skip=0pt}%
\begin{subfigure}[t]{0.45\columnwidth}\centering\includegraphics[width=\linewidth]{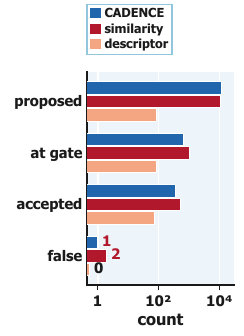}\caption{}\label{fig:exp_ev_a}\end{subfigure}%
\begin{subfigure}[t]{0.45\columnwidth}\centering\includegraphics[width=\linewidth]{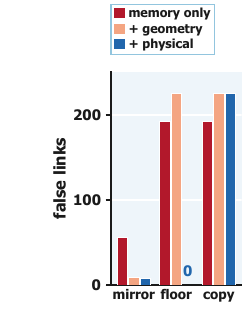}\caption{}\label{fig:exp_ev_b}\end{subfigure}\\[1pt]
\begin{subfigure}[t]{0.45\columnwidth}\centering\includegraphics[width=\linewidth]{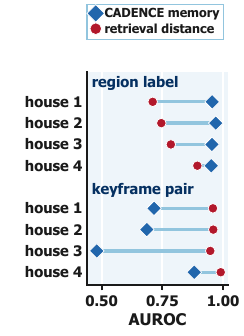}\caption{}\label{fig:exp_ev_c}\end{subfigure}%
\begin{subfigure}[t]{0.45\columnwidth}\centering\includegraphics[width=\linewidth]{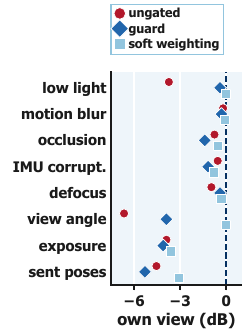}\caption{}\label{fig:exp_ev_d}\end{subfigure}
\caption{Results for (a)~association, (b)~injected structure, (c)~ranking AUROC and (d)~degraded peers.}
\label{fig:exp_ranking}
\label{fig:exp_peer_qual}

\end{figure}

\subsection{Setup}
\label{sec:exp_setup}
We evaluate local reconstruction on Replica-8~\cite{straub2019replica,sucar2021imap} and collaboration on MA-Replica~\cite{hu2023cpslam} and time-interleaved three-agent Matterport3D (MP3D) walks~\cite{chang2017matterport3d}.
ScanNet++~\cite{yeshwanth2023scannet} provides iPhone images, LiDAR depth, real IMU data and ARKit poses, whereas barometers and other inertial streams are synthetic.
We tune on the development split (DEV), select configurations on the validation split (VAL) and evaluate held-out splits once after freezing, with TEST3 as the last of three disjoint held-out phone splits.
We fixed acceptance thresholds before the collaboration studies, and memory training excludes the eight test houses and the held-out phone scenes (\cref{sec:supp_thresholds,sec:supp_review_protocol}).
Solo runs an agent alone, Online performs causal deterministic reconstruction rather than real-time reconstruction, Refined adds offline optimization, and Paced feeds $30$\,Hz input to the real-time configuration CADENCE-RT.
The absolute trajectory error (ATE) is the rigid-aligned position root mean square error in cm, and PSNR is in dB. In the tables, bold and underlined values mark the best and second-best results per group, and a dagger marks coverage below $95\,\%$.
PSNR, SSIM and learned perceptual image patch similarity (LPIPS)~\cite{zhang2018unreasonable} score renders at the estimated poses of each method, including mapping keyframes, and thus measure reconstruction consistency rather than novel-view synthesis.
Inputs are color and depth (RGB-D) images with raw IMU and ARKit motion where indicated, and Table~\ref{tab:supp_phone_collab} identifies the depth-hidden control.
Published collaborative systems use other inputs and protocols, and Table~\ref{tab:supp_extended_results} lists their reported values only as an external reference.

\paragraph{Implementation.}
The memory caps each $5$\,m runtime region at eight embeddings; sensitivity to the $2$\,m training size is reported in \cref{sec:supp_region}. Cue thresholds and peer summaries govern modality permissions. At most two retrieved or pose-near candidates undergo verification per peer keyframe, with $10\,\%$ exploration below the correspondence cutoff. Geometric thresholds are $\tau_r=2$\,px and $\tau_d=0.10$\,m (\cref{sec:supp_round2_settings,sec:supp_fusion}). Phone runs retain cue permissions but disable the learned reliability cap because sparse ground truth cannot reliably label motion (\cref{sec:supp_phone_collab_results}). Collaboration uses the guard; \cref{sec:exp_degraded} compares both sharing policies. Timings use precomputed embeddings; live sessions use handcrafted descriptors, with SG-WCM evaluated on replay. \Cref{sec:supp_mapping,sec:supp_backend,sec:supp_apple} describe mapping schedules, device budgets and parity bounds.

\begin{table*}[t]
\vspace{-1.0em}
\centering
\caption{Single-agent results on Replica-8 and on the final held-out iPhone split (TEST3), grouped by input.}
\label{tab:exp_replica}
\footnotesize
\setlength{\tabcolsep}{4pt}
\renewcommand{\arraystretch}{1.05}
\begin{tabular}{@{}l c c rrrr l rrrr@{}}
\toprule
& & & \multicolumn{4}{c}{Replica-8 (synthetic)} & \multicolumn{5}{c}{Held-out iPhone scans} \\
\cmidrule(lr){4-7}\cmidrule(l){8-12}
Method & Input & Mode & ATE$\downarrow$ & PSNR$\uparrow$ & SSIM$\uparrow$ & LPIPS$\downarrow$ & Split & ATE$\downarrow$ & PSNR$\uparrow$ & SSIM$\uparrow$ & LPIPS$\downarrow$ \\
\midrule
\groupband{12}{Full system: RGB-D, IMU and ARKit motion}
\oursrow\textcolor{cvprblue}{\textbf{CADENCE}} & D+I+P & online & -- & -- & -- & -- & TEST3 & 6.98 & \second{20.99} & \second{0.811} & \second{0.327} \\
\oursrow\textcolor{cvprblue}{\textbf{CADENCE refined}} & D+I+P & refined & -- & -- & -- & -- & TEST3 & \second{6.81} & \best{24.04} & \best{0.850} & \best{0.213} \\
ARKit (platform) & device & & -- & -- & -- & -- & TEST3 & \best{4.16} & -- & -- & -- \\
\groupband{12}{With the IMU}
\oursrow\textcolor{cvprblue}{\textbf{CADENCE}} & D+I & online & \best{0.26} & \best{37.40} & \best{0.970} & \best{0.042} & -- & -- & -- & -- & -- \\
ORB-SLAM3~\cite{campos2021orb} & D+I & & -- & -- & -- & -- & TEST3 & \best{5.82}$^\dagger$ & -- & -- & -- \\
\groupband{12}{RGB-D only: same input as the baselines}
\oursrow\textcolor{cvprblue}{\textbf{CADENCE}} & D & online & \second{0.41} & 38.50 & \second{0.975} & \second{0.038} & TEST3 & 41.04$^\dagger$ & 17.46 & 0.744 & \second{0.444} \\
\oursrow\textcolor{cvprblue}{\textbf{CADENCE refined}} & D & refined & \second{0.41} & \best{39.16} & \best{0.977} & \best{0.033} & -- & -- & -- & -- & -- \\
\oursrow\textcolor{cvprblue}{\textbf{CADENCE-RT}} & D & real time & \second{0.41} & 33.61 & 0.937 & 0.121 & -- & -- & -- & -- & -- \\
Photo-SLAM~\cite{huang2024photoslam} & D & & 0.60 & 34.96 & 0.942 & 0.059 & TEST3 & \best{18.31}$^\dagger$ & \best{22.87}$^\dagger$ & \best{0.840} & \best{0.279} \\
GS-ICP SLAM~\cite{ha2024rgbd} & D & & \best{0.16} & \second{38.83} & \second{0.975} & 0.041 & TEST3 & 83.47 & \second{17.97} & \second{0.751} & 0.544 \\
ORB-SLAM3 & D & & -- & -- & -- & -- & TEST3 & \second{19.57}$^\dagger$ & -- & -- & -- \\
\bottomrule
\end{tabular}
\end{table*}
\begin{figure*}[t]
\centering
\includegraphics[width=\textwidth,trim=0 0 0 70.5bp,clip]{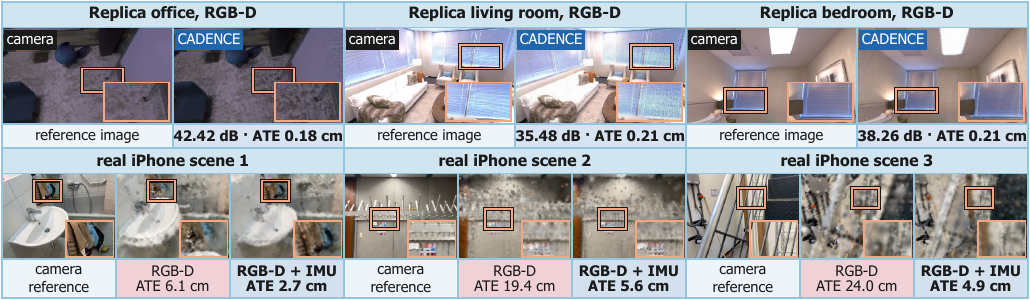}
\caption{Camera references and CADENCE maps of three VAL iPhone scenes tracked with RGB-D input alone and with the raw phone IMU, without the ARKit prior. Labels state the scene ATE, and insets enlarge the most textured window.}
\label{fig:exp_rgb_reconstruction}
\label{fig:exp_gain}
\label{fig:exp_phone_qual}

\end{figure*}

\subsection{World-Memory Ranking}
\label{sec:exp_ranking}
With memory commits taken from ground-truth labels, SG-WCM raises the ranking AUROC over embedding similarity on five of six fixed evaluation splits and at first contact before any peer commit (MP3D $k{=}0$, Table~\ref{tab:exp_wcm}).
The splits include 7-Scenes~\cite{shotton2013scene} and TUM~\cite{sturm2012benchmark}, with a standard deviation of at most $0.006$. A0 uses raw DINOv3 similarity, A1 a learned visual memory, A2 adds region-relative pose, and the adopted A5rel adds gravity, altitude and quality (Table~\ref{tab:supp_wcm_full}).
Region-relative pose contributes the largest step ($+0.043$ on the memory test, 95\,\% interval from $+0.031$ to $+0.053$).
The gain is largest on hard pairs, where the low-overlap AUROC rises from $0.926$ to $0.968$ on MP3D and from $0.815$ to $0.895$ on 7-Scenes.
Deployed on the four linked unseen houses, the memory separates consistent regions better than retrieval distance, with a region AUROC between $0.951$ and $0.970$ compared with $0.710$ to $0.894$ for retrieval (Fig.~\ref{fig:exp_ranking}c, per house for region labels and keyframe pairs).
The two signals complement each other, because retrieval orders keyframe pairs at an AUROC of $0.947$ to $0.990$ compared with $0.480$ to $0.881$ for the memory, and Eq.~\ref{eq:method_rank} therefore selects regions by memory and keyframes by retrieval. First contact without an alignment prior relies on retrieval, where the memory raises AUROC from $0.940$ to $0.953$ (\cref{sec:supp_region}).

\begin{table}[t]
\centering
\footnotesize
\setlength{\tabcolsep}{3.2pt}
\renewcommand{\arraystretch}{1.05}
\caption{Memory ranking ablation in AUROC with label-committed memory on fixed splits, averaged over three seeds.}
\label{tab:exp_wcm}
\begin{tabular}{@{}lccccccc@{}}
\toprule
& \multicolumn{6}{c}{Frozen split} & MP3D \\
\cmidrule(lr){2-7}
Model & Mem.\ test & 7-Sc. & Degr. & MP3D & TUM & TUM d. & $k{=}0$ \\
\midrule
A0 & 0.913 & 0.876 & 0.960 & 0.939 & 0.955 & 0.985 & 0.879 \\
A1 & 0.921 & 0.904 & 0.966 & 0.950 & 0.960 & \best{0.987} & 0.877 \\
A2 & \best{0.964} & 0.926 & \best{0.975} & 0.972 & 0.942 & 0.962 & 0.933 \\
\oursrow A5rel & 0.946 & \best{0.931} & 0.974 & \best{0.977} & \best{0.969} & 0.971 & \best{0.948} \\
\bottomrule
\end{tabular}
\end{table}

\subsection{Verified Collaborative Fusion}
\label{sec:exp_verification}
On eight unseen MP3D houses with the same verifier and physical checks disabled (Table~\ref{tab:exp_assoc}, Fig.~\ref{fig:exp_ranking}a), hybrid association accepts $345$ links with one false link, compared with $503$ links and two false links for embedding similarity. Both false-link rates remain below $0.5\,\%$, where a false link deviates from ground truth by more than $0.30$\,m or $5^\circ$. The memory improves region ranking, while both policies link the same four houses with similar map quality (\cref{sec:supp_assoc_detail}). Memory guides candidate selection, while geometric verification checks the proposed relations. After alignment, nearby memory-ranked keyframes provide $328$ of the $345$ links from $9\,\%$ of verification attempts. The single false hybrid link occurs before alignment, with a $55$\,cm error (\cref{sec:supp_assoc_detail}). Accepting memory predictions without geometry admits five false links among $76$ on VAL and all $192$ injected floor copies, whereas geometric verification accepts no false links among $29$ VAL links. 

Physical checks additionally reject all $225$ floor copies accepted by geometry, although the gravity bound also rejects twelve of the $29$ true links (Fig.~\ref{fig:exp_ranking}b, Table~\ref{tab:supp_injected}). On phone data, verification accepts $671$ links among three agents cut from one VAL iPhone scan with overlapping frames; none of the $54$ scored links is false. Four live two-phone Wi-Fi sessions across three iPhones without LiDAR verify $535$ links using descriptor proposals and the multi-view path. For a rigidly joined pair, estimated separation remains between $7.8$ and $11.1$\,cm over $117$\,s, with a 90th-percentile height difference of $2.0$\,cm. Replaying the recordings with SG-WCM and Core ML embeddings verifies $206$ links, compared with $189$ for handcrafted descriptors. In a separate walk, learned retrieval proposes candidates for all $330$ exchanged keyframes, compared with one for handcrafted descriptors (Table~\ref{tab:supp_phone_collab}, Fig.~\ref{fig:supp_phone_live}).
\begin{table}[t]
\centering
\footnotesize
\setlength{\tabcolsep}{2.6pt}
\caption{Association policies under one verifier on MP3D with three agents per house and physical checks disabled.}
\label{tab:exp_assoc}
\begin{tabular}{@{}lcccccc@{}}
\toprule
 & \multicolumn{3}{c}{VAL (4 houses)} & \multicolumn{3}{c}{Unseen (8 houses)} \\
\cmidrule(lr){2-4}\cmidrule(l){5-7}
Policy & Attempts & Links & False & Attempts & Links & False \\
\midrule
Descriptor retrieval & 9 & 5 & 0 & 87 & 73 & 0 \\
Embedding similarity & 4527 & 43 & 1 & 10351 & 503 & 2 \\
\oursrow\textcolor{cvprblue}{\textbf{CADENCE (hybrid)}} & 6041 & 29 & 0 & 11840 & 345 & 1 \\
\oursrow\quad without IMU & 5470 & 20 & 0 & 11444 & 128 & 0 \\
\bottomrule
\end{tabular}
\end{table}
\label{sec:exp_collab}
\label{sec:exp_single}
Merged MA-Replica maps reach a keyframe ATE (ATE-KF) of $0.35$\,cm and a PSNR of $37.26$\,dB, whereas published collaborative Gaussian systems report $0.14$ to $0.25$\,cm and $34.26$ to $41.35$\,dB under their own protocols. \Cref{sec:supp_extended_results} reports the local, rotation, keyframe-budget and multi-floor results in detail.

\subsection{Degraded Peers}
\label{sec:exp_degraded}

On Replica VAL scenes office0 and office1, we compare the guard, soft weighting, ungated sharing, equal weights and frame rejection under one geometric verifier (Table~\ref{tab:exp_degraded}). Own-view luma PSNR change against Solo measures receiver harm outside the peer window; peer-only-region change against Merge measures sharing benefit. Across eight degradations, soft weighting reduces mean own-view loss from $2.66$ to $1.05$\,dB, improves on ungated sharing in seven conditions and retains peer-region gains in six (Fig.~\ref{fig:exp_peer_qual}d, two-scene means). Frame rejection causes less receiver harm but largely forfeits these gains. Under low light, soft weighting removes the receiver loss while retaining useful depth. The guard limits this loss to $0.40$\,dB and reduces peer-region depth error by $17.39$\,cm while withholding unreliable color (Table~\ref{tab:supp_scalar_guard}). This supports treating appearance and geometry separately: poor color need not invalidate useful depth. Soft weighting nevertheless causes less mean receiver harm than the guard in all eight conditions. Most effects arise in office0; office1 establishes its first link late and integrates at most two guarded peer keyframes in seven conditions (Table~\ref{tab:supp_degraded_scenes}). Exposure mismatch and noisy transmitted poses still pass the confidence tests, showing that observation quality alone does not establish compatibility with the receiver map. An additional depth check renders peer keyframes against that map and reduces the guard's mean own-view loss under noisy transmitted poses from $5.29$ to $1.25$\,dB, while retaining a peer-region gain. The improvement is concentrated in office0, and the check remains outside the default system and misses IMU-corrupted peer poses (\cref{sec:supp_consistency}).

\subsection{Held-Out Phone Sequences}
\label{sec:exp_phone}
In Table~\ref{tab:exp_replica}, D denotes RGB-D input, I denotes raw IMU input (synthetic on Replica-8) and P denotes the ARKit prior. Baselines run on our phone replays, and their Replica-8 entries are published values (Table~\ref{tab:supp_extended_results}). On Replica-8, the median online ATE of CADENCE is $0.21$\,cm, and one scene dominates the mean. On twelve TEST3 iPhone scenes, adding raw IMU and ARKit motion reduces the mean online ATE from $41.04$ to $6.98$\,cm. Refinement adds $3.05$\,dB PSNR at a median extra cost of $262$\,s, and both improvements hold in every scene. On VAL, raw IMU alone reduces ATE from $13.76$ to $5.61$\,cm and removes drift blur (Fig.~\ref{fig:exp_rgb_reconstruction}). With these inputs, CADENCE tracks every scene with full coverage and renders better on average than GS-ICP SLAM. ARKit and inertial ORB-SLAM3, which completes $11$ of $12$ scenes, track more accurately, whereas neither renders a map. CADENCE relies on visual-inertial input, and without the IMU and platform motion ($41.04$\,cm), ORB-SLAM3 and Photo-SLAM track better on all three evaluated splits (\cref{sec:supp_extended_results}). Refined CADENCE exceeds the saved Photo-SLAM map by $1.17$\,dB PSNR on average and in eight of twelve scenes, and lowers LPIPS by $0.066$ (paired $95\,\%$ interval excluding zero, nine of twelve scenes).

\subsection{Runtime and Deployment}
\label{sec:exp_runtime}
\begin{figure}[t]
\centering
\includegraphics[width=\columnwidth]{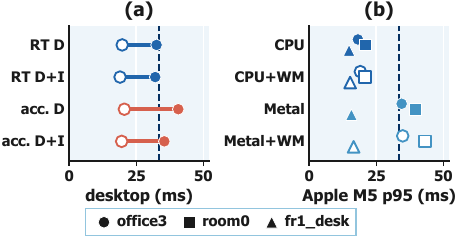}
\caption{Pose-call time on (a)~the desktop paced at $30$\,Hz and (b)~the Apple M5 against the $33.3$\,ms contract.}
\label{fig:exp_runtime}

\end{figure}

Tracking meets the $33.3$\,ms pose-call contract of Eq.~\ref{eq:method_contract} in every real-time desktop and iPhone configuration. Paced on the desktop, CADENCE-RT records a p95 pose time of $32.5$\,ms with $3.3\,\%$ late frames and reaches $33.61$\,dB PSNR on Replica-8 (Table~\ref{tab:exp_replica}, Fig.~\ref{fig:exp_runtime}). The accuracy-first configuration reaches $38.50$\,dB with a larger mapping budget in non-real-time execution (Table~\ref{tab:supp_resources}). These configurations provide operating points that balance reconstruction quality and processing time. Fig.~\ref{fig:exp_runtime}a compares their median (open) and worst-scene p95 (filled) timings. On the Apple M5, the default configuration with CPU mapping meets the contract, including with active memory, which adds $0.35$\,ms to p95 pose time on average (+WM in Fig.~\ref{fig:exp_runtime}b). With precomputed embeddings, memory inference therefore introduces little tracking overhead. Contract exceedances occur in two Replica sequences under the separate configuration where Metal mapping and depth registration share the GPU. The iPhone~11 Pro Max replays a $15$\,Hz ScanNet++ LiDAR sequence at a p95 pose time of $20.3$\,ms without dropping a frame; six short live stereo-depth sessions also meet the contract (Table~\ref{tab:exp_device}). With IMU input and Metal mapping, tracking remains responsive alongside on-device map updates. Table~\ref{tab:exp_device} reports pose-time distributions and the fraction of calls above the contract, with device profiles in Table~\ref{tab:supp_device_runtime}. Across three live two-phone sessions, p95 pose times stay within budget at $10.0$ to $11.3$\,ms on the iPhone~15 and $17.0$ to $26.7$\,ms on the iPhone~11 Pro Max (\cref{sec:supp_runtime_detail,sec:supp_comm}). The tested collaboration workload thus runs alongside tracking within the pose-time budget.

\begin{table}[t]
\centering
\footnotesize
\setlength{\tabcolsep}{2.2pt}
\renewcommand{\arraystretch}{1.05}
\caption{Pose-call time in ms on iPhones for one LiDAR replay, six live stereo-depth sessions and three live two-phone sessions.}
\label{tab:exp_device}
\begin{tabular}{@{}lccccc@{}}
\toprule
Run & p50 & p95 & Over & Memory & Gauss. \\
\midrule
\oursrow LiDAR replay & $14.3$ & $20.3$ & $1.0$\,\% & $1.0$\,GB & $144$\,k \\
\oursrow Live stereo ($6$) & $12.3$--$16.1$ & $19.5$--$32.2$ & $0.0$--$4.5$\,\% & -- & -- \\
\oursrow Two-phone, A16 & $6.5$--$6.9$ & $10.0$--$11.3$ & $0.0$--$0.3$\,\% & $2.6$\,GB & ${\le}590$\,k \\
\oursrow Two-phone, A13 & $10.2$--$11.1$ & $17.0$--$26.7$ & $0.9$--$4.4$\,\% & $1.8$\,GB & ${\le}147$\,k \\
\bottomrule
\end{tabular}
\end{table}

\subsection{Qualitative and Error Analysis}
\label{sec:exp_qual}
In Fig.~\ref{fig:exp_rgb_reconstruction}, drift under RGB-D input alone blurs textured regions, and the raw IMU removes this blur. Fig.~\ref{fig:supp_qual} compares the renders of all methods on the best, median and worst TEST3 scenes. In two low-light office0 views, ungated sharing costs $13.4$ and $13.8$\,dB against Solo, whereas the guard changes PSNR by $+0.4$ and $-1.0$\,dB (Fig.~\ref{fig:supp_degraded}). The analyzed failures arise from geometrically consistent yet wrong structure, since verification accepts $8$ of $56$ mirrored look-alike views, every exact copy placed $30$\,m away, and $9$ false links among $413$ accepted links on 7-Scenes.
Moreover, exposure mismatch and noisy transmitted poses pass the confidence tests (\cref{sec:exp_degraded}), because the cues measure observation quality rather than map compatibility.

\section{Conclusion}
\label{sec:conclusion}
We presented CADENCE, a collaborative Gaussian SLAM system that separates finding correspondences from deciding which aligned evidence may affect another agent's map.
To this end, it combines sensor-grounded regional ranking, geometric verification, and per-modality confidences.
At the same time, local tracking stays independent of the background collaboration, and the Gaussian map and the verified graph remain the authoritative state.
On unseen houses, the regional memory ranks peer regions with an AUROC of $0.95$ to $0.97$, and verification keeps false links below $0.5\,\%$.
Moreover, confidence-governed sharing allows a peer in a dark scene to contribute its depth while its color is withheld.
The system also runs on commodity hardware.
On iPhones without LiDAR, live two-phone collaboration verifies $535$ links at a 95th-percentile pose time of $10.0$ to $11.3$\,ms.
Taken together, these results suggest that collaborative reconstruction should be evaluated by both the evidence it accepts and the effect of that evidence on the receiving map.

\paragraph{Limitations and Future Work.}
\label{sec:exp_limits}
All experiments are indoor. Collaboration is evaluated in replay with synthetic barometers (and a synthetic IMU outside ScanNet++) and in live iPhone sessions, where a rigidly joined phone pair replaces ground truth.
Verification may accept geometrically consistent copies of a place unless a physical check separates them, and cue-based confidences miss a peer whose evidence is self-consistent but contradicts the receiver's map.
Future work will address map-aware confidence estimation and validation on independent multi-phone recordings with real sensors and external ground truth.

\clearpage
\section*{LLM Usage Statement}

The authors developed the research problem, conceptual framework, hypotheses, methodology, theoretical results and benchmark design independently before using generative AI tools. Claude and OpenAI Codex subsequently assisted implementation, routine programming, plotting scripts, supplementary controls, dataset cleaning and refinement of author-drafted text. The authors supervised tool use, reviewed all assisted outputs, and verified citations against original sources. All reported numerical results were obtained using the authors’ implementation. The code will be made publicly available.

{
    \small
    \bibliographystyle{ieeenat_fullname}
    \bibliography{main}
}

\begin{bibunit}[ieeenat_fullname]
\makeatletter
\let\Hy@backout\@gobble
\def\hyper@natlinkstart#1{\hyper@linkstart{cite}{cite.supp.#1}\def\hyper@nat@current{supp.#1}}
\def\hyper@natlinkbreak#1#2{\hyper@linkend#1\hyper@linkstart{cite}{cite.supp.#2}}
\def\hyper@natanchorstart#1{\Hy@raisedlink{\hyper@anchorstart{cite.supp.#1}}}
\def\backref#1{}\def\backrefalt#1#2#3#4{}
\makeatother
\clearpage
\setcounter{page}{1}
\gdef\thepage{S\arabic{page}}%
\maketitlesupplementary
\setcounter{section}{0}
\renewcommand{\thesection}{\Alph{section}}
\renewcommand{\theHsection}{supp.\Alph{section}}
\renewcommand{\theHsubsection}{\theHsection.\arabic{subsection}}
\setcounter{table}{0}\setcounter{figure}{0}\setcounter{equation}{0}
\renewcommand{\thetable}{S\arabic{table}}\renewcommand{\thefigure}{S\arabic{figure}}\renewcommand{\theequation}{S\arabic{equation}}
\renewcommand{\theHtable}{supp.\arabic{table}}\renewcommand{\theHfigure}{supp.\arabic{figure}}\renewcommand{\theHequation}{supp.\arabic{equation}}

\setcounter{dbltopnumber}{4}\renewcommand{\dbltopfraction}{0.98}\renewcommand{\dblfloatpagefraction}{0.92}
\setcounter{topnumber}{4}\renewcommand{\topfraction}{0.98}\renewcommand{\floatpagefraction}{0.92}
\setcounter{totalnumber}{8}\renewcommand{\textfraction}{0.02}
\clubpenalty=50 \widowpenalty=50 \displaywidowpenalty=50
\setlength{\floatsep}{4pt plus 1pt minus 1pt}\setlength{\textfloatsep}{6pt plus 1pt minus 1pt}
\setlength{\intextsep}{4pt plus 1pt minus 1pt}\setlength{\dblfloatsep}{4pt plus 1pt minus 1pt}
\setlength{\dbltextfloatsep}{6pt plus 1pt minus 1pt}
\setlength{\abovecaptionskip}{3pt}\setlength{\belowcaptionskip}{0pt}
\setlength{\abovedisplayskip}{4pt plus 2pt minus 2pt}\setlength{\belowdisplayskip}{4pt plus 2pt minus 2pt}
\setlength{\abovedisplayshortskip}{2pt plus 2pt minus 1pt}\setlength{\belowdisplayshortskip}{2pt plus 2pt minus 1pt}
\setlength{\aboverulesep}{1pt}\setlength{\belowrulesep}{1pt}
\makeatletter
\renewcommand\paragraph{\@startsection{paragraph}{4}{\z@}%
  {1.6ex \@plus 0.6ex \@minus 0.2ex}{-1em}{\normalfont\normalsize\bfseries}}
\makeatother

This supplement provides the evidence and details behind the main paper.
\Cref{sec:supp_glance} summarizes how CADENCE works, collects the notation (Table~\ref{tab:supp_notation}) and shows the system running in the iPhone app.
\Cref{sec:supp_results} reports the complete results behind \cref{sec:exp}: association and verification, world-memory ranking, degraded peers, collaboration on real phones, communication cost, single-agent accuracy, runtime, and studies of independent agents, camera rigs and multiple floors.
\Cref{sec:supp_review_protocol} defines the data splits, metrics, statistics and baselines, and \cref{sec:supp_impl} gives the implementation details needed to reproduce the system, in the order of \cref{sec:method}.
\Cref{sec:supp_scope} closes with the caveats that apply when reading the results.

\section{Overview}
\label{sec:supp_glance}
\label{sec:supp_app_flow}
CADENCE lets several phones, called agents, reconstruct one static indoor scene together while each phone keeps tracking in real time (Fig.~\ref{fig:method_overview}).
Every agent keeps its own world frame and Gaussian map, and only the owning agent modifies its map.
Collaboration therefore decides three things separately: which peer keyframes may show the same place (ranking), whether a proposed match is geometrically correct (verification), and which kinds of peer evidence may change the receiver's map (permissions).

\paragraph{Local Reconstruction (\cref{sec:method_local}; \cref{sec:supp_local,sec:supp_mapping}).}
A visual-inertial tracker estimates the pose of every frame from image features, the IMU and, where available, sensor depth.
Its pose call never waits for mapping or collaboration, which run in the background, and is measured against a $33.3$\,ms real-time contract.
Keyframes enter bounded queues, and background mapping jobs grow the agent's Gaussian map within a compute budget.

\paragraph{Observation Reliability (\cref{sec:method_reliability}; \cref{sec:supp_reliability}).}
Nine cues describe the image quality, motion, texture, viewpoint, depth and pose reliability of each keyframe and yield four confidences, for pose, geometry, appearance and memory.
The guard turns each confidence into a permission with a threshold, whereas soft weighting scales every update by the smallest confidence.

\paragraph{Regional Memory (\cref{sec:method_wcm}; \cref{sec:supp_wcm}).}
The sensor-grounded world-consistency memory (SG-WCM) summarizes the committed observations of each spatial region from appearance and sensor evidence: region-relative pose, gravity, barometric altitude and observation quality.
It scores how likely a peer keyframe shows each region, and it never changes a map.

\paragraph{Verified Fusion (\cref{sec:method_fusion}; \cref{sec:supp_fusion}).}
Agents broadcast small announcements for their keyframes.
A receiver ranks candidates by memory score and embedding retrieval (hybrid association), requests one evidence packet, and accepts a relation only if geometric tests on matched features and depth pass, optionally followed by gravity, barometric and consistency checks.
Accepted relations align the peer's frame in a pose graph, and the permissions then decide how strongly the peer's pose constraints, geometry, appearance and memory updates enter the receiver.

\paragraph{Execution (\cref{sec:method_runtime}; \cref{sec:supp_controller,sec:supp_backend,sec:supp_modes,sec:supp_apple}).}
A controller sheds background work under latency, memory, thermal or battery pressure without changing the tracker or the acceptance thresholds.
CPU, CUDA and Metal backends optimize the same objective and are tested against CPU references.
Deterministic runs give bit-identical results on one binary, and paced runs feed frames in real time.

\nolinenumbers
\begin{table*}[t]
\centering
\footnotesize
\setlength{\tabcolsep}{3pt}
\renewcommand{\arraystretch}{1.0}
\caption{Symbols of \cref{sec:method} and \cref{sec:supp_impl} with the location of each definition.}
\label{tab:supp_notation}
\begin{tabular}{@{}>{\raggedright\arraybackslash}p{0.13\textwidth}>{\raggedright\arraybackslash}p{0.27\textwidth}l@{\hspace{10pt}}>{\raggedright\arraybackslash}p{0.15\textwidth}>{\raggedright\arraybackslash}p{0.27\textwidth}l@{}}
\toprule
Symbol & Meaning & Defined & Symbol & Meaning & Defined \\
\midrule
$a,b\in\mathcal A$ & agents; set of agents & \S\ref{sec:method_overview} & $r$, $\mathbf z_r$ & cubic region; its 32-d token & \S\ref{sec:method_wcm} \\
$W_a$, $B$, $C$ & world frame of $a$; IMU body; camera & \S\ref{sec:method_overview} & $Z_W$ & learned memory (all tokens) & \S\ref{sec:method_overview} \\
$T_{XY}$ & rigid transform, frame $Y$ to $X$ & \S\ref{sec:method_overview} & $\mathbf x(o,r)$ & 95-d query of observation $o$ & \S\ref{sec:method_wcm} \\
$T_{W_aB}(t)$ & body trajectory of agent $a$ & \S\ref{sec:method_overview} & $\mathcal C_r$, $n_r$ & committed observations of $r$; count & Eq.~\ref{eq:method_sensor_relative} \\
$\mathcal G,\mathcal P,\mathcal F$ & Gaussians; pose and submap graph; floors & \S\ref{sec:method_overview} & $\boldsymbol\gamma_o$, $h_o$, $\sigma_{h,o}$ & body gravity$/9.81$; altitude; its uncertainty & Eq.~\ref{eq:method_sensor_relative} \\
$S_k$, $T_{W_aS_k}$ & Gaussian submap; its placement & \S\ref{sec:method_mapping} & $f_\psi$; $\boldsymbol\ell$, $\boldsymbol\delta$ & memory perceptron; logits, token increment & Eq.~\ref{eq:method_wcm} \\
$\mathbf y$, $m$ & nine quality cues; motion-blur risk & Eq.~\ref{eq:method_conf} & $p^{\mathrm{corr}},p^{\mathrm{floor}},p^{\mathrm{rel}}$ & correspondence, floor, reliability heads & \S\ref{sec:method_wcm} \\
$c^\mu$, $\mu\in\{p,g,a,w\}$ & pose, geometry, appearance, memory confidence & Eq.~\ref{eq:method_conf} & $\eta_{ok}$ & region score of candidate $k$ for peer observation $o$ & Eq.~\ref{eq:method_rank} \\
$w^\mu$, $\theta_\mu$ & permission weight; its threshold & \S\ref{sec:method_reliability} & $\hat T$, $\pi$ & proposed relative pose; projection & Eq.~\ref{eq:method_verify} \\
$\hat C_v,\hat A_v,\hat D_v$ & rendered color, opacity, depth of view $v$ & Eq.~\ref{eq:method_loss} & $\mathbf X_i$; $\mathbf u_i,\mathbf u'_i$ & 3-D match; its pixels in both keyframes & Eq.~\ref{eq:method_verify} \\
$\mathcal V$, $\lambda$ & views of a mapping job; SSIM weight & Eq.~\ref{eq:method_loss} & $d_{\mathrm S}$; $z_j,\hat z_j$ & Sampson error; measured, predicted depth & Eq.~\ref{eq:method_verify} \\
$\mathcal E^c_v$, $\mathcal E^d_v$ & mean squared color, depth error & Eq.~\ref{eq:method_loss} & $\tau_r$, $\tau_d$ & verification thresholds ($2$\,px, $0.10$\,m) & Eq.~\ref{eq:method_verify} \\
$\mathbf a_v$, $\mathbf b_v$ & affine exposure gain, offset (Refined) & \S\ref{sec:exp_setup} & $\hat T_{B_iB_j}$, $\hat T_{W_aW_b}$ & accepted relative pose; frame alignment & Eq.~\ref{eq:method_alignment} \\
$\mathcal G_n$ & Gaussian $(\boldsymbol\mu_n,\mathbf q_n,\mathbf s_n,o_n,\mathbf c_n)$ & Eq.~\ref{eq:supp_render} & $\tilde{\mathbf c}_o$, $\bar{\mathbf c}_b$, $\mathbf c^{\mathrm{eff}}_{b,o}$ & reported, running, effective peer confidence & Eq.~\ref{eq:method_peer_conf} \\
$\alpha_i$, $\beta_i$, $\Sigma'_i$ & opacity, transmittance, image covariance & Eq.~\ref{eq:supp_render} & $\nu$ & running-summary rate ($0.2$) & Eq.~\ref{eq:method_peer_conf} \\
$\varphi_k$ & keyframe sampling importance & Eq.~\ref{eq:supp_importance} & $d_{ok}$, $r_k$ & embedding distance; candidate's region & Eq.~\ref{eq:method_rank} \\
$R,\mathbf v,\mathbf p,\mathbf g$ & IMU rotation, velocity, position; gravity & Eq.~\ref{eq:supp_predict} & $\mathbf e^{\mathrm{ga}}_{or}$ & relative altitude and gravity evidence & Eq.~\ref{eq:method_sensor_relative} \\
$\rho$, $\lambda_R$ & Huber penalty; rotation-prior weight & Eq.~\ref{eq:supp_pnp} & $\mathcal L_\psi$, $s_k$ & memory loss; learned log-variance & Eq.~\ref{eq:method_wcm_loss} \\
$\mathbf q_j$; $\mathbf m_j,\mathbf n_j$ & depth point; surfel point and normal & Eq.~\ref{eq:supp_reg} & $\overline{\mathrm{BCE}}_k$, $\mathcal L_{\mathrm{err}}$ & mean binary cross-entropy; error-head loss & Eq.~\ref{eq:method_wcm_loss} \\
$\chi_j$, $\sigma_p$ & surfel confidence; residual scale & Eq.~\ref{eq:supp_reg} & $\tau_t$, $\tau^\star$ & pose-call time; $33.3$\,ms contract & Eq.~\ref{eq:method_contract} \\
$w^p_{ij}$, $\Lambda_{ij}$, $\mathbf r_{ij}$ & factor weight, information, residual & Eq.~\ref{eq:supp_graph} & $\ell_t$ & controller level & \S\ref{sec:supp_controller} \\
$\varrho$, $\Lambda_0$, $\mathbf r_n$ & graph Huber kernel; node prior & Eq.~\ref{eq:supp_graph} & $\Phi$, $\sigma$, $b$ & run map; job schedule; backend & \S\ref{sec:supp_modes} \\
\bottomrule
\end{tabular}
\end{table*}
\linenumbers

\begin{figure*}[t]
\centering
\includegraphics[width=\textwidth]{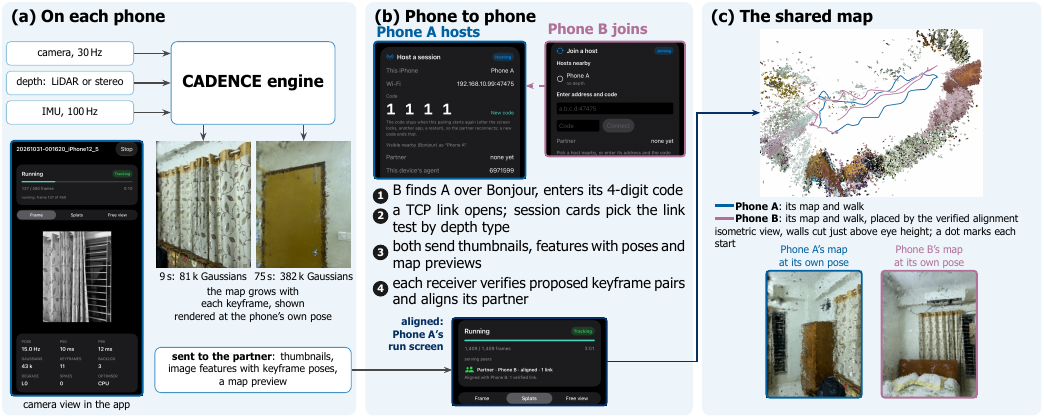}
\caption{iPhone application flow. (a)~One phone with camera view and growing map. (b)~Pairing over Bonjour with a code and exchange of evidence. (c)~Both maps and walks in the frame of Phone~A (isometric view).}
\label{fig:supp_app_flow}
\end{figure*}

\paragraph{iPhone App.}
Figure~\ref{fig:supp_app_flow} shows this pipeline in the iPhone app.
Each phone records color images with depth (LiDAR through ARKit or, without LiDAR, the dual-camera stereo pair) and Core Motion inertial samples, and the C++ engine tracks, selects keyframes and maps with the Metal optimizer.
To collaborate, one phone hosts a session on the local Wi-Fi network (Bonjour) and shows a four-digit code; the partner enters it, and the phones connect over the Transmission Control Protocol (TCP).
A short session card states each phone's depth type, which selects the link test: one-frame for metric (LiDAR) depth, multi-view for stereo depth.
The phones then exchange gray thumbnails, image features with keyframe poses and a colored map preview; the partner's preview and path are drawn for display only, and only verified links enter the pose graph.

In Fig.~\ref{fig:supp_app_flow}, the screens come from two iPhone~17 simulators that replay two walks recorded on an iPhone~11 Pro Max. An Apple M5 builds the maps from the same walks, and Phone~B is placed by the alignment that the app verified. The $382$\,k Gaussians of panel (a) fit the memory tier of the M5; a $4$\,GB phone keeps a smaller map (Table~\ref{tab:supp_apple}a).

\section{Extended Results}
\label{sec:supp_results}

\subsection{Association and Verification}
This subsection extends \cref{sec:exp_verification}.
Table~\ref{tab:exp_assoc} and Fig.~\ref{fig:exp_ranking}a report, under one shared verifier, how much verification each association policy costs and how many of its links are correct. Table~\ref{tab:supp_injected} tests each check separately by injecting false structure (Fig.~\ref{fig:exp_ranking}b).
On the unseen houses, $664$ hybrid and $1037$ similarity proposals obtain a relative pose (feature matching and RANSAC on the peer's evidence packet) and reach the geometric gate ($100$ vs.\ $172$ on VAL).
In $24$ network-fault cases, none of the $375$ pose-graph constraints is false.
\begin{table}[htb]
\centering
\footnotesize
\setlength{\tabcolsep}{3.5pt}
\caption{Injected false structure on four MP3D VAL houses with hybrid association, reported as false links among accepted links.}
\label{tab:supp_injected}
\begin{tabular}{@{}lccc@{}}
\toprule
Condition & Unverified & Geometric & + physical \\
\midrule
Mirrored look-alike views & 56 of 56 & 9 of 9 & 8 of 8 \\
Floor copy 3 or 6\,m above & 192 of 192 & 225 of 225 & 0 of 0 \\
Exact copy 30\,m away & 192 of 192 & 225 of 225 & 225 of 225 \\
Normal walks & 5 of 76 & 0 of 29 & 0 of 17 \\
\bottomrule
\end{tabular}
\end{table}

In Table~\ref{tab:supp_injected}, the Unverified column accepts every pair whose predicted memory correspondence is at least $0.5$, without any geometric check; as a different gate, it can accept fewer links than the geometric one. The + physical column adds the gravity, barometric-height and pairwise-consistency checks.

The barometric height check rejects all floor copies, which the learned floor head alone does not separate (median $p^{\mathrm{floor}}$ $0.986$ over the $225$ geometrically accepted copies, minimum $0.754$).
An exact copy of a place $30$\,m away matches in geometry, gravity and height and therefore passes every check; this is a limit of verification from local sensor evidence.
The gravity check is conservative: it rejects twelve true links whose measured tilt is a constant $3.12^\circ$, just above the $3^\circ$ bound, because the verifier acts on the available sensor evidence rather than on whether a link is truly correct.

\label{sec:supp_thresholds}
We fixed all acceptance thresholds ($\theta_\mu$, $\tau_r$, $\tau_d$ and the $0.2$ pose-confidence floor) and the physical bounds ($3^\circ$ tilt, half a story) before the collaboration studies, and did not change them even after the gravity check rejected true links.
Only the embedding screens were chosen, on the four VAL houses.
With these values, false links are $0$ of $29$ on VAL, $1$ of $345$ on the unseen MP3D houses and $9$ of $413$ on 7-Scenes, with exact (Clopper-Pearson) $95\,\%$ upper bounds of $11.9$, $1.6$ and $4.1\,\%$; similarity retrieval gives $2$ of $503$ (bound $1.4\,\%$) on the unseen houses.
The 7-Scenes and Replica collaboration scenes were used to train the memory, whereas the unseen houses were not.

\label{sec:supp_assoc_detail}
We judge association against ground truth in two ways.
Two keyframes are co-visible if their depth maps, placed at the ground-truth poses, overlap in $5$\,cm voxels (overlap $\ge0.30$ is positive, $\le0.05$ negative), and a relation is false if its pose is off by more than $0.30$\,m or $5^\circ$.
\nolinenumbers
The one false hybrid link on the unseen houses (house 4, agents one and three) was the first link in that house, $5.5$\,s into the replay.
Before any alignment, the embedding screen picked this pair (distance $0.33$, memory correspondence only $0.024$), and PnP accepted it with $16$ inliers (minimum $12$), placing the keyframes $4.6$\,m apart with $0.55$\,m error.
With the physical checks off and no earlier alignment to compare against, no check could reject this link. The $35$ later links of that house ($34$ of them memory-ranked) all lie within $5$\,cm.
\linenumbers

The unseen houses need $11840$ hybrid verifications in total.
The embedding screen before the first alignment uses $10677$ of them and yields seven links (the false one included), memory-ranked candidates after alignment use $1060$ and yield $328$ links, and exploration ($10\,\%$ of peer keyframes, checked regardless of the memory score) uses $103$ and yields ten.
Most of the verification cost is therefore spent on finding the first contact.
Table~\ref{tab:supp_assoc_detail} compares hybrid and similarity association per linked house.
\begin{table}[htb]
\centering
\footnotesize
\setlength{\tabcolsep}{2.5pt}
\caption{Association per linked house for hybrid (H) and embedding similarity (S) under the same verifier.}
\label{tab:supp_assoc_detail}
\begin{tabular}{@{}lccccccc@{}}
\toprule
 & \multicolumn{2}{c}{Links} & \multicolumn{2}{c}{False} & \multicolumn{2}{c}{First link (s)} & Align. \\
\cmidrule(lr){2-3}\cmidrule(lr){4-5}\cmidrule(lr){6-7}
House & H & S & H & S & H & S & (cm) \\
\midrule
\groupband{8}{VAL}
VAL 1 & 12 & 26 & 0 & 1 & 20.1 & 20.1 & 12.0 \\
VAL 2 & 5 & 5 & 0 & 0 & 22.8 & 22.8 & 20.8 \\
VAL 3 & 12 & 12 & 0 & 0 & 18.3 & 18.3 & 9.9 \\
\groupband{8}{Unseen houses}
house 1 & 20 & 55 & 0 & 0 & -- & -- & -- \\
house 2 & 247 & 334 & 0 & 0 & -- & -- & -- \\
house 3 & 42 & 66 & 0 & 0 & -- & -- & -- \\
house 4 & 36 & 48 & 1 & 2 & -- & -- & -- \\
\bottomrule
\end{tabular}
\end{table}

Table~\ref{tab:supp_assoc_detail} lists accepted and false links, the time of the first verified link (replay seconds) and the frame-alignment error of hybrid association (Align.). The rotation part of this error is $1.7$ to $2.9^\circ$, and on VAL similarity retrieval aligns within $1.2$\,cm and $0.13^\circ$ of hybrid association. Dashes mark values not reported for the unseen houses. VAL house 4 has no co-visible keyframe pair (Table~\ref{tab:supp_region}), so no policy can link it, and four unseen houses link under no policy. The MP3D identifiers of VAL houses 1 to 4 are 2azQ1b91cZZ, 7y3sRwLe3Va, XcA2TqTSSAj and pa4otMbVnkk, and the identifiers of unseen houses 1 to 4 are 29hnd4uzFmX, 5ZKStnWn8Zo, 759xd9YjKW5 and SN83YJsR3w2.

\begin{table}[htb]
\centering
\footnotesize
\setlength{\tabcolsep}{5pt}
\caption{Map-level depth check under noisy sent poses, as receiver own-view PSNR change vs.\ Solo (dB).}
\label{tab:supp_map_checks}
\begin{tabular}{@{}lcc@{}}
\toprule
DEV scene & CADENCE guard & + map verification \\
\midrule
Replica room0 & $-4.11$ & $+0.12$ \\
Replica office3 & $-9.47$ & $-0.19$ \\
\bottomrule
\end{tabular}
\end{table}
Similarity retrieval finds more links, but they connect no additional house and arrive no earlier (descriptor retrieval links only two VAL houses).
Per-agent ATE is identical under every policy, because a relation moves whole frames and submap origins, not the trajectory inside them.
Mean own-view PSNR differs by at most $0.04$\,dB on VAL and $0.02$\,dB on the held-out houses (per agent, at most $0.14$ and $0.35$\,dB, respectively).

\label{sec:supp_consistency}
\begin{table*}[t]
\centering
\scriptsize
\setlength{\tabcolsep}{3pt}
\renewcommand{\arraystretch}{0.9}
\caption{World-memory ranking ablations and memory cost. (a)~Teacher-forced (label-committed) AUROC on frozen splits. (b)~Memory policies on two Replica rooms in the cache harness (offline, no encoder latency).}
\label{tab:supp_wcm_full}
\label{tab:supp_memory_budget}
\begin{tabular}{@{}lcccccccc@{}}
\multicolumn{9}{@{}l}{\textbf{(a) Ranking ablations on frozen splits}} \\
\toprule
Model & WM-test & 7-Scenes & Degraded & MP3D & TUM & TUM degr. & MP3D $k=0$ & Floor \\
\midrule
A0 & 0.913 & 0.876 & 0.960 & 0.939 & 0.955 & \second{0.985} & 0.879 & -- \\
A1 & $0.921 \pm 0.000$ & $0.904 \pm 0.001$ & $0.966 \pm 0.000$ & $0.950 \pm 0.000$ & $0.960 \pm 0.000$ & \best{$0.987 \pm 0.000$} & $0.877 \pm 0.000$ & 0.769 \\
A2 & \best{$0.964 \pm 0.000$} & $0.926 \pm 0.005$ & $0.975 \pm 0.001$ & $0.972 \pm 0.000$ & $0.942 \pm 0.002$ & $0.962 \pm 0.002$ & $0.933 \pm 0.001$ & 0.884 \\
A3 & $0.961 \pm 0.001$ & $0.923 \pm 0.004$ & \second{$0.978 \pm 0.002$} & \second{$0.976 \pm 0.000$} & $0.948 \pm 0.001$ & $0.963 \pm 0.003$ & \second{$0.947 \pm 0.001$} & 0.922 \\
A4 & \second{$0.962 \pm 0.002$} & $0.926 \pm 0.003$ & $0.973 \pm 0.002$ & $0.972 \pm 0.000$ & $0.947 \pm 0.005$ & $0.953 \pm 0.000$ & $0.937 \pm 0.001$ & 0.921 \\
A5 & $0.943 \pm 0.003$ & \second{$0.929 \pm 0.004$} & \best{$0.979 \pm 0.001$} & $0.975 \pm 0.001$ & $0.953 \pm 0.003$ & $0.967 \pm 0.006$ & $0.944 \pm 0.002$ & \second{0.926} \\
A3rel & $0.958 \pm 0.001$ & $0.927 \pm 0.002$ & \best{$0.979 \pm 0.000$} & $0.975 \pm 0.000$ & $0.964 \pm 0.003$ & $0.975 \pm 0.001$ & $0.946 \pm 0.001$ & \best{\textbf{0.928}} \\
A4rel & $0.958 \pm 0.002$ & \best{$0.931 \pm 0.004$} & \second{$0.978 \pm 0.001$} & $0.975 \pm 0.001$ & \second{$0.965 \pm 0.001$} & $0.975 \pm 0.001$ & $0.944 \pm 0.001$ & 0.924 \\
A5rel & $0.946 \pm 0.001$ & \best{$0.931 \pm 0.006$} & $0.974 \pm 0.001$ & \best{$0.977 \pm 0.001$} & \best{$0.969 \pm 0.001$} & $0.971 \pm 0.003$ & \best{$0.948 \pm 0.002$} & 0.921 \\
\bottomrule
\end{tabular}

\vspace{0.4em}
{\setlength{\tabcolsep}{4pt}%
\begin{tabular}{@{}l@{}}
\textit{(b) Memory-policy controls in the cache harness} \\
\begin{tabular}{@{}llrrrrr@{}}
\toprule
Scene & Memory policy & Size (KB) & Query p95 (ms) & AUROC & Reference AUROC & Labeled queries \\
\midrule
office0 & region & 49.53 & 0.038 & 1.000 & 1.000 & 229 \\
office0 & full history & 1143.81 & 0.175 & 0.177 & 0.747 & 373 \\
office0 & fixed window & 25.95 & 0.052 & -- & 0.756 & 376 \\
office0 & global cache & 12.38 & 0.042 & 1.000 & 0.949 & 245 \\
office1 & region & 12.38 & 0.042 & 0.984 & 0.984 & 248 \\
office1 & full history & 982.38 & 0.210 & -- & 1.000 & 323 \\
office1 & fixed window & 25.95 & 0.054 & -- & 0.701 & 323 \\
office1 & global cache & 12.38 & 0.042 & 0.984 & 0.984 & 248 \\
\bottomrule
\end{tabular}
\end{tabular}}
\end{table*}
\begin{figure*}[t]
\centering
\includegraphics[width=0.85\textwidth]{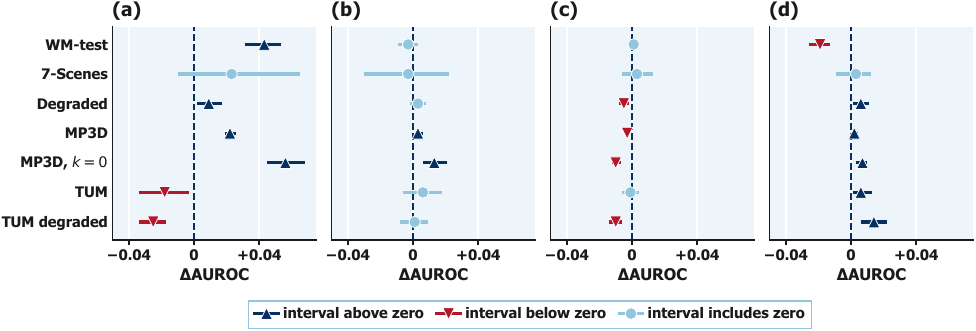}
\caption{AUROC change of each absolute-encoding step with $95$\,\% paired cluster-bootstrap intervals for (a)~+ region-relative pose, (b)~+ IMU increments and VIO poses, (c)~+ gravity and altitude and (d)~+ observation quality.}
\label{fig:supp_ranking_ladder}
\end{figure*}
Pairwise consistency compares a new relation with the peer's earlier alignments.
It therefore cannot reject an exact copy (whose $225$ links all agree with each other) or a false first contact, which has nothing earlier to compare against. On 7-Scenes, it lowered false links from $14$ of $426$ to $9$ of $413$.
A map-level depth check, which renders peer keyframes against the receiver's map, is not part of the default system and is off in every reported run, as is a photometric gain test.
On the two Replica VAL scenes, the depth check reduces the guard's mean own-view loss under noisy sent poses from $5.29$ to $1.25$\,dB, mainly in office0 ($-9.07$ to $-0.99$\,dB, with the peer-region gain rising from $+2.35$ to $+4.80$\,dB); it does not detect peer poses corrupted through the IMU, and it leaves runs with a clean peer unchanged.

Table~\ref{tab:supp_map_checks} shows the same effect on the two DEV scenes, with the sent poses perturbed by $0.10$\,m and $3^\circ$ over the whole window; photometric degradations are left nearly unchanged.

\subsection{World-Memory Ranking}
This subsection extends \cref{sec:exp_ranking} and Table~\ref{tab:exp_wcm}.
Table~\ref{tab:supp_wcm_full} and Fig.~\ref{fig:supp_ranking_ladder} add the memory's evidence one kind at a time (encoding in Eq.~\ref{eq:method_sensor_relative} and \cref{sec:supp_wcm}).
In these offline tests, the ground-truth correspondence labels decide which observations are committed to the memory (teacher forcing), so AUROC measures ranking alone, independent of runtime verification.

Part (a) of Table~\ref{tab:supp_wcm_full} reports the mean and standard deviation over three seeds where available. Each step of the ladder adds one kind of evidence: A0 is raw DINOv3 embedding similarity, A1 the learned visual memory, A2 adds region-relative pose, A3 adds IMU increments (which also changes the tracker poses), A4 adds gravity and altitude, and A5 adds observation quality. The suffix rel expresses gravity and altitude relative to the region's committed observations and drops the IMU increments; A5rel is the adopted model. WM-test is the world-memory (WM) test split, MP3D $k{=}0$ is first contact pooled over alignment regimes, and Floor is the floor-head AUROC. Bold and underline mark the two best means per column and do not imply statistical significance.

All panels of Fig.~\ref{fig:supp_ranking_ladder} share one scale, and VIO denotes visual-inertial odometry. Dark blue marks an interval above zero, red marks an interval below zero, and light blue marks an interval that includes zero.

\linenumbers
We adopted A5rel based on validation AUROC averaged over domains, not on any single split.
On the memory test split alone, A2 is higher, because inertial increments and quality cues lower that split by $0.003$ and $0.019$; the adopted encoding is therefore relative and drops the inertial increments.
Adding inertial evidence (A2$\to$A3) gains $+0.003$ on Matterport3D, with an interval above zero, and raises floor AUROC, although this step also changes the tracker poses.
Plain similarity stays ahead of A5rel on degraded TUM, and on VAL a descriptor ranks keyframe pairs best; association therefore combines retrieval with the memory (hybrid).
Without an alignment prior (MP3D no-prior stratum), the memory raises AUROC from $0.940$ to $0.953$, and retrieval handles first contact (\cref{sec:method_fusion}).
On the low-overlap pairs of the fixed splits, the memory raises AUROC over embedding similarity from $0.926$ to $0.968$ on Matterport3D and from $0.815$ to $0.895$ on 7-Scenes (A0 to A5rel).
\begin{table*}[t]
\centering
\scriptsize
\setlength{\tabcolsep}{4.2pt}
\renewcommand{\arraystretch}{0.88}
\setcounter{table}{6}%
\caption{Sharing with a degraded peer as the mean of office0 and office1 (per scene in Table~\ref{tab:supp_degraded_scenes}), with own-view and peer-region luma PSNR change (dB) per policy.}
\label{tab:exp_degraded}
\begin{tabular}{@{}lrrrrrrrrrr@{}}
\toprule
 & \multicolumn{2}{c}{Ungated} & \multicolumn{2}{c}{Equal weight} & \multicolumn{2}{c}{Frame rejection} & \multicolumn{2}{c}{CADENCE soft weight} & \multicolumn{2}{c}{CADENCE guard} \\
\cmidrule(lr){2-3}\cmidrule(lr){4-5}\cmidrule(lr){6-7}\cmidrule(lr){8-9}\cmidrule(l){10-11}
Peer degradation & own & peer & own & peer & own & peer & own & peer & own & peer \\
\midrule
\groupband{11}{Small own-view loss}
Low light & $-3.73$ & $-0.96$ & $-4.86$ & $-2.08$ & \best{$+0.01$} & \second{$-0.02$} & \best{$+0.01$} & \best{$+1.05$} & \second{$-0.40$} & $-0.47$ \\
Motion blur & $-0.20$ & \best{$+10.15$} & $-0.30$ & \second{$+10.08$} & \best{$0.00$} & $-0.01$ & \second{$-0.07$} & $+5.24$ & $-0.31$ & $+3.21$ \\
\groupband{11}{Moderate own-view loss}
Partial occlusion & $-0.76$ & $+2.01$ & $-1.06$ & $+2.18$ & \second{$-0.71$} & \best{$+2.99$} & \best{$-0.54$} & $+1.95$ & $-1.39$ & \second{$+2.94$} \\
IMU corruption & \second{$-0.55$} & \best{$+10.15$} & $-0.96$ & \second{$+9.30$} & \best{$0.00$} & $+0.02$ & $-0.81$ & $+5.91$ & $-1.18$ & $+4.69$ \\
\groupband{11}{Benefit withheld}
Defocus ($\sigma=4$\,px) & $-0.96$ & \best{$+9.64$} & $-0.92$ & \second{$+9.59$} & $-0.57$ & $0.00$ & \best{$-0.31$} & $+2.49$ & \second{$-0.40$} & $+0.10$ \\
\groupband{11}{Harm not prevented}
Bad view angle & $-6.65$ & \second{$-1.52$} & $-7.80$ & $-2.64$ & \best{$0.00$} & \best{$-0.01$} & \second{$-0.01$} & \best{$-0.01$} & $-3.89$ & $-1.97$ \\
Exposure mismatch & $-3.89$ & $-6.12$ & $-8.44$ & $-7.86$ & \best{$-1.90$} & \best{$-2.49$} & \second{$-3.58$} & $-6.09$ & $-4.10$ & \second{$-5.97$} \\
Noisy sent poses ($0.10$\,m, $3^\circ$) & $-4.54$ & \best{$+2.37$} & $-7.79$ & $+1.02$ & \best{$-0.26$} & $-0.13$ & \second{$-3.08$} & \second{$+2.19$} & $-5.29$ & $+1.75$ \\
\bottomrule
\end{tabular}
\end{table*}
\begin{figure*}[t]
\centering
\includegraphics[width=\textwidth]{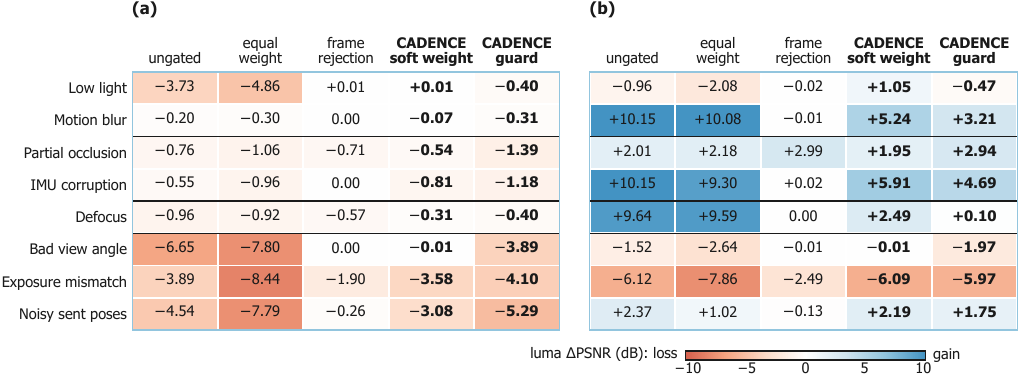}
\caption{Sharing outcome per policy and degradation, as luma PSNR change of (a)~the receiver vs.\ Solo and (b)~the peer-only region relative to Merge.}
\label{fig:supp_reliability}
\end{figure*}
Deployed on the four linked unseen houses ($5$\,m regions, commits only after verification), the memory ranks region labels at AUROC $0.951$ to $0.970$ (Fig.~\ref{fig:exp_ranking}c), against $0.710$ to $0.894$ for retrieval distance.
The two signals complement each other: keyframes of a region share its score, so retrieval orders keyframe pairs better ($0.947$ to $0.990$, against $0.480$ to $0.881$), and Eq.~\ref{eq:method_rank} selects regions by memory and keyframes by retrieval.
The region memory needs $49.5$\,KB (full history: $1.1$\,MB) and answers in under $0.1$\,ms (Table~\ref{tab:supp_memory_budget}); the policies answer different labeled queries, so their AUROC values do not rank them.
Query scheduling leaves all $168$ single-agent PSNR, ATE and Gaussian-count metrics unchanged, because the memory never modifies a map.

\label{sec:supp_region}
Collaboration runs use $5$\,m regions, whereas the memory learned on $2$\,m regions, so Table~\ref{tab:supp_region} tests $2$\,m regions against a criterion fixed beforehand (pair AUROC up in at least two houses, region-label AUROC down by at most $0.02$ in each).
\begin{table}[htb]
\centering
\footnotesize
\setlength{\tabcolsep}{3pt}
\setcounter{table}{5}%
\caption{Region size of $5$\,m (default) vs.\ $2$\,m (training) on four MP3D VAL houses with hybrid association.}
\label{tab:supp_region}
\begin{tabular}{@{}lcccccc@{}}
\toprule
 & \multicolumn{2}{c}{Pair} & \multicolumn{2}{c}{Region label} & \multicolumn{2}{c}{Links} \\
\cmidrule(lr){2-3}\cmidrule(lr){4-5}\cmidrule(l){6-7}
House & $5$\,m & $2$\,m & $5$\,m & $2$\,m & $5$\,m & $2$\,m \\
\midrule
VAL 1 & 0.890 & 0.949 & 0.981 & 0.990 & 12 & 9 \\
VAL 2 & 0.995 & 0.995 & 0.991 & \textbf{0.903} & 5 & 5 \\
VAL 3 & 0.636 & 0.919 & 0.978 & 0.978 & 12 & 15 \\
VAL 4 & -- & -- & -- & -- & 0 & 0 \\
\midrule
All four & & & & & 29 & 29 \\
\bottomrule
\end{tabular}
\end{table}

Pair and Region label are AUROC values, bold marks a value outside the $0.02$ tolerance, Links counts accepted links (none false), and a dash marks a house without a co-visible keyframe pair.
Regions of $2$\,m improve pair ranking in the two weakest houses but fail the region-label condition in VAL 2, so $5$\,m stays the default; trajectories and renders are identical at both sizes, and $2$\,m regions would add $0.4$\,\% PnP attempts.

\begin{table*}[t]
\centering
\scriptsize
\setlength{\tabcolsep}{3.2pt}
\renewcommand{\arraystretch}{0.9}
\setcounter{table}{8}%
\caption{Degraded peers per scene, as own-view and peer-region luma PSNR change (dB) for ungated sharing (U), the guard (G) and soft weighting (S).}
\label{tab:supp_degraded_scenes}
\begin{tabular}{@{}l rrr rrr r rrr rrr r@{}}
\toprule
 & \multicolumn{7}{c}{office0} & \multicolumn{7}{c}{office1} \\
\cmidrule(lr){2-8}\cmidrule(l){9-15}
 & \multicolumn{3}{c}{Own} & \multicolumn{3}{c}{Peer region} & & \multicolumn{3}{c}{Own} & \multicolumn{3}{c}{Peer region} & \\
\cmidrule(lr){2-4}\cmidrule(lr){5-7}\cmidrule(lr){9-11}\cmidrule(lr){12-14}
Peer degradation & U & G & S & U & G & S & KF & U & G & S & U & G & S & KF \\
\midrule
Low light & $-6.75$ & $-0.79$ & $-0.48$ & $-1.49$ & $-0.95$ & $+2.07$ & 39 & $-0.72$ & $0.00$ & $+0.49$ & $-0.43$ & $0.00$ & $+0.03$ & 0 \\
Motion blur & $-1.11$ & $-0.62$ & $-0.59$ & $+9.98$ & $+6.41$ & $+8.68$ & 47 & $+0.72$ & $0.00$ & $+0.45$ & $+10.32$ & $0.00$ & $+1.79$ & 0 \\
Partial occlusion & $-1.68$ & $-2.78$ & $-1.37$ & $+5.17$ & $+5.87$ & $+6.42$ & 109 & $+0.16$ & $0.00$ & $+0.29$ & $-1.14$ & $0.00$ & $-2.52$ & 0 \\
IMU corruption & $-1.72$ & $-2.35$ & $-1.70$ & $+9.84$ & $+9.38$ & $+9.10$ & 113 & $+0.62$ & $0.00$ & $+0.08$ & $+10.47$ & $0.00$ & $+2.73$ & 0 \\
Defocus & $-2.29$ & $-0.80$ & $-0.97$ & $+9.70$ & $+0.19$ & $+0.13$ & 72 & $+0.38$ & $0.00$ & $+0.35$ & $+9.58$ & $0.00$ & $+4.86$ & 0 \\
Bad view angle & $-10.03$ & $-7.79$ & $-0.03$ & $-0.68$ & $-3.94$ & $-0.02$ & 40 & $-3.26$ & $0.00$ & $0.00$ & $-2.36$ & $0.00$ & $0.00$ & 0 \\
Exposure mismatch & $-5.80$ & $-6.78$ & $-3.84$ & $-6.21$ & $-6.62$ & $-5.97$ & 111 & $-1.98$ & $-1.43$ & $-3.33$ & $-6.03$ & $-5.32$ & $-6.21$ & 50 \\
Noisy sent poses & $-7.54$ & $-9.07$ & $-4.71$ & $+3.76$ & $+2.35$ & $+3.93$ & 111 & $-1.54$ & $-1.52$ & $-1.46$ & $+0.99$ & $+1.15$ & $+0.45$ & 2 \\
\bottomrule
\end{tabular}
\end{table*}
\begin{figure*}[t]
\centering
\includegraphics[width=\textwidth]{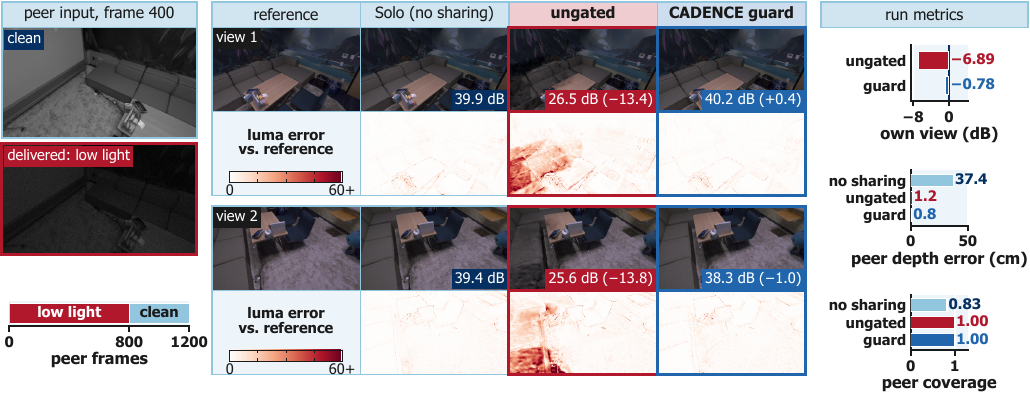}
\caption{A low-light peer in one Replica run (office0). Left, peer frame 400 clean and as delivered. Middle, two receiver views by Solo, ungated sharing and the guard. Right, own-view and peer-region metrics.}
\label{fig:supp_degraded}
\label{fig:supp_degraded_metrics}
\label{fig:supp_peer_input}
\end{figure*}
Each region stores at most eight observations, a cap reached by $7$ of $20$ verified VAL peer observations at $5$\,m ($8$ of $21$ at $2$\,m).
Caps of $16$ or $32$ let peers commit $18$ instead of $13$ observations but lower mean pair AUROC ($0.840\to0.815$ at $5$\,m, $0.954\to0.938$ at $2$\,m, nearly all in one house) and links ($29$ to $27$ and $26$); four leave no room for peer commits ($32$ and $27$ links). No cap beats eight at both sizes, and no link is false in any setting.

\subsection{Degraded Peers}
This subsection extends \cref{sec:exp_degraded} with five sharing policies on two Replica VAL scenes (Table~\ref{tab:exp_degraded}, Fig.~\ref{fig:supp_reliability}): own is the receiver's PSNR change against Solo on views outside the peer's window, and peer is the change against Merge in the region that only the peer observed.
Rows are grouped by the guard's own-view loss, bold and underline mark the two best values among the own and among the peer columns, and the figure uses one diverging scale (red loss, blue gain).
Photometric degradations begin after a clean overlap, whereas IMU corruption and sent-pose noise last the whole window.
A healthy peer adds $6.42$\,dB in the peer region at a cost of $0.62$\,dB own-view PSNR (office0: $+10.32$ and $-1.51$\,dB; office1: $+2.53$ and $+0.27$\,dB).

\begin{table}[htb]
\centering
\footnotesize
\setlength{\tabcolsep}{2.4pt}
\setcounter{table}{7}%
\caption{Guard (G) vs.\ soft weighting (S) in the three separating conditions, two Replica VAL scenes pooled.}
\label{tab:supp_scalar_guard}
\scriptsize
\begin{tabular}{@{}llrrrrr@{}}
\toprule
 & & \multicolumn{2}{c}{Own view} & \multicolumn{3}{c}{Peer region} \\
\cmidrule(lr){3-4}\cmidrule(l){5-7}
Condition & & PSNR & Depth & PSNR & Depth & Cov. \\
\midrule
Low light & G & $-0.40$ & $+0.01$ & $-0.47$ & $-17.39$ & $+0.083$ \\
 & S & $+0.01$ & $+0.01$ & $+1.05$ & $-16.18$ & $+0.078$ \\
Bad view angle & G & $-3.89$ & $+2.97$ & $-1.97$ & $-6.08$ & $+0.079$ \\
 & S & $-0.01$ & $0.00$ & $-0.01$ & $0.00$ & $0.000$ \\
Noisy sent poses & G & $-5.29$ & $+2.88$ & $+1.75$ & $-37.13$ & $+0.205$ \\
 & S & $-3.08$ & $+1.51$ & $+2.19$ & $-43.11$ & $+0.237$ \\
\bottomrule
\end{tabular}
\end{table}

Soft weighting harms the receiver's own view less than ungated sharing in seven of eight conditions (mean $-1.05$ against $-2.66$\,dB) and less than the guard in all eight, in five of them in both scenes.
The guard reduces own-view harm against ungated sharing under low light, defocus and bad view angle, and adds $0.11$ to $0.75$\,dB of loss in the other five conditions.
Office1 offers little shared evidence: its first link forms after $24$\,s, and in seven conditions the guard integrates at most two peer keyframes (column KF of Table~\ref{tab:supp_degraded_scenes}), so office0 carries most of the comparison.
\label{sec:supp_reliability_detail}
\begin{table*}[t]
\centering
\setcounter{table}{9}%
\caption{Collaboration on real phones as links accepted at the fixed thresholds, with pictures of (a)~three agents on the ground-truth path, (b)~multi-view links on the same scan and (c)~the live app.}
\label{tab:supp_phone_collab}
{\scriptsize\setlength{\tabcolsep}{0pt}%
\begin{tabular}{@{}c@{\hspace{8pt}}c@{\hspace{8pt}}c@{}}
\includegraphics[height=1.25in]{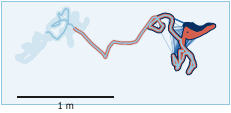} & \includegraphics[height=1.25in]{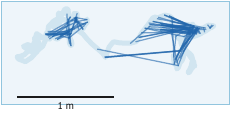} & \includegraphics[height=1.25in]{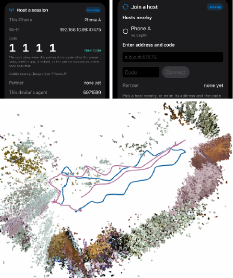} \\[1pt]
(a) three agents, LiDAR scan (VAL) & (b) multi-view links, same scan & (c) live, two iPhones \\
\end{tabular}}\par\medskip
\footnotesize
\setlength{\tabcolsep}{4pt}
\renewcommand{\arraystretch}{1.08}
\begin{tabular}{@{}>{\raggedright\arraybackslash}p{0.17\textwidth}>{\raggedright\arraybackslash}p{0.17\textwidth}llrr>{\raggedright\arraybackslash}p{0.3\textwidth}@{}}
\toprule
Setting & Device / data & Depth & Link test & Links & False & Alignment / note \\
\midrule
\groupband{7}{(a) Recorded LiDAR iPhone scans, three agents cut from one scan}
Reliability cap on & ScanNet++ iPhone, VAL & LiDAR & one-frame & $0$ & -- & $3161$ proposals: $1360$ without a relative pose, $1801$ denied by the pose permission \\
Cap off & ScanNet++ iPhone, DEV & LiDAR & one-frame & $238$ & $0$ of $29$ & six agent pairs aligned to $7.4$\,cm, $9.0^\circ$ \\
Cap off, verify first & ScanNet++ iPhone, DEV & LiDAR & one-frame & $264$ & $0$ of $31$ & six agent pairs aligned to $7.4$\,cm, $9.0^\circ$ \\
Cap off, verify first & ScanNet++ iPhone, VAL & LiDAR & one-frame & $671$ & $0$ of $54$ & six agent pairs aligned to $2.1$\,cm, $2.4^\circ$; the $700$-frame cuts share frames \\
\groupband{7}{(b) Offline revisits, DINOv3 proposals at least $5$\,s apart}
Multi-view rule & ScanNet++ iPhone, VAL & LiDAR & multi-view & $107$ & $0$ & \\
 & ScanNet++ iPhone, DEV & LiDAR & multi-view & $69$ & $1$ & \\
 & Replica room0 & synthetic & multi-view & $302$ & $0$ & \\
 & Replica office3 & synthetic & multi-view & $219$ & $0$ & \\
 & ScanNet++ iPhone, VAL & none & multi-view & $81$ & $2$ & depth hidden from the tracker (camera and IMU only) \\
One-frame path & TUM fr1\_desk & Kinect & one-frame & $51$ & $0$ & metric-depth devices keep this path \\
\groupband{7}{(c) Phones without LiDAR}
Engine replay, three walks & iPhone~11 Pro Max & stereo & one-frame & $0$ & -- & stereo depth fails the depth test ($307$ proposals) \\
 & iPhone~11 Pro Max & stereo & multi-view & $13$ & -- & all within $5^\circ$ of accelerometer gravity \\
Live Wi-Fi session, two minutes, memory off & iPhone~13 (joins) & stereo & multi-view & $67$ & -- & aligned after $17$\,s \\
 & iPhone~11 Pro Max (hosts) & stereo & multi-view & $45$ & -- & aligned after $21$\,s; rotation agrees within $3.6^\circ$; the alignments compose to $79.5$\,cm in translation \\
Live Wi-Fi, $2$\,min, memory off & iPhone~15 / iPhone~11 Pro Max & stereo & multi-view & $114$ / $51$ & -- & aligned after $19$\,s; first alignment within $3.7$\,cm, $0.74^\circ$ of the final \\
Live Wi-Fi, $2.5$\,min, phones held together & iPhone~15 / iPhone~11 Pro Max & stereo & multi-view & $40$ / $110$ & -- & aligned after $27$\,s; estimated separation median $12.3$\,cm, height difference median $1.2$\,cm \\
Live Wi-Fi, $2$\,min, phones rigidly joined & iPhone~15 / iPhone~11 Pro Max & stereo & multi-view & $54$ / $54$ & -- & aligned after $38$\,s; estimated separation $7.8$--$11.1$\,cm, height difference ${\le}\,2.0$\,cm (p90), tilt difference $3.0^\circ$ \\
Replay with the memory and Core ML embeddings & iPhone~15 / iPhone~11 Pro Max, held-together recordings & stereo & multi-view & $112$ / $94$ & -- & every keyframe embedded and scored by the memory; $125$ memory commits \\
\quad same replay, handcrafted descriptor & iPhone~15 / iPhone~11 Pro Max & stereo & multi-view & $71$ / $118$ & -- & $189$ links with the same verifier \\
\bottomrule
\end{tabular}
\end{table*}
The two policies differ in two ways, which these complete-system comparisons do not separate: soft weighting scales every permitted update by the smallest of the four confidences of each raw frame report, whereas the guard thresholds each modality and caps each report by a running summary of the peer.
Table~\ref{tab:supp_scalar_guard} shows the three conditions that separate them (PSNR in dB, depth error in cm, own-view depth over all own views, Cov.\ = coverage change).
Under low light, both keep the peer's depth, but the guard's threshold blocks the color that soft weighting admits at a small weight.
Figure~\ref{fig:supp_degraded} follows the office0 low-light run, whose delivered peer frames stay dark for the first 800 frames. Its middle panels give the frame PSNR, its change against Solo and the luma error (white none, dark red at least $60$ gray levels); its right panels plot the own-view PSNR change against Solo and the peer-region depth error and coverage against no sharing.
On the DEV scenes room0 and office3, with the peer at $10\,\%$ illumination, the guard changes own views by $-0.32$\,dB and keeps $+6.58$\,dB of peer-region gain over no sharing (ungated: $-0.74$ and $-5.02$\,dB); at $50\,\%$ illumination, the peer-region gains are $+6.74$ and $+1.58$\,dB.
Under bad view angle and noisy sent poses, the guard admits geometry that clears $\theta_g$ but disagrees with the receiver's map, which no cue measures, so the receiver's own depth error rises; soft weighting ties this geometry to the weakest cue.
Likewise, no cue measures a global gain change, so exposure mismatch remains open for every policy that admits color ($3.58$ to $8.44$\,dB own-view loss).

\subsection{Collaboration on Real Phones}
\label{sec:supp_phone_collab_results}
Table~\ref{tab:supp_phone_collab} collects the results; a false link lies beyond $0.30$\,m or $5^\circ$ from ground truth, and a dash marks a run without ground truth.

\paragraph{Recorded LiDAR Scans.}
\begin{figure*}[t]
\centering
\captionsetup[sub]{labelformat=parens,font=footnotesize,skip=0pt}%
\includegraphics[width=\textwidth]{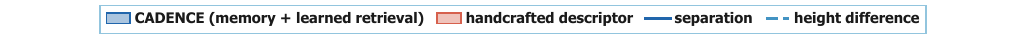}\\[1pt]
\begin{subfigure}[t]{0.2\textwidth}\centering\includegraphics[width=\linewidth]{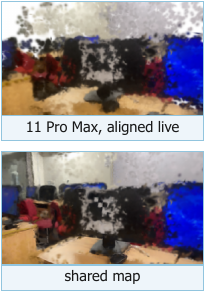}\caption{}\end{subfigure}%
\begin{subfigure}[t]{0.2\textwidth}\centering\includegraphics[width=\linewidth]{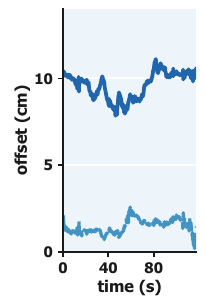}\caption{}\end{subfigure}%
\begin{subfigure}[t]{0.2\textwidth}\centering\includegraphics[width=\linewidth]{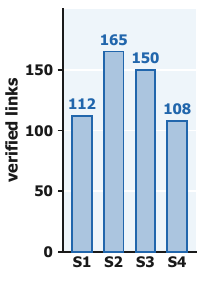}\caption{}\end{subfigure}%
\begin{subfigure}[t]{0.2\textwidth}\centering\includegraphics[width=\linewidth]{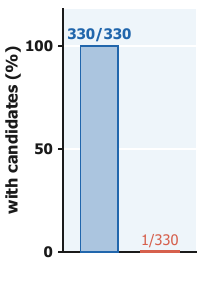}\caption{}\end{subfigure}%
\begin{subfigure}[t]{0.2\textwidth}\centering\includegraphics[width=\linewidth]{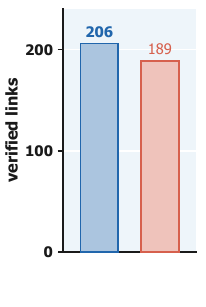}\caption{}\end{subfigure}
\caption{Live collaboration of iPhones without LiDAR, with (a)~maps placed by the live alignment, (b)~estimated offset of rigidly joined phones, (c)~verified links per session, (d)~received keyframes with retrieval candidates and (e)~verified links in replay.}
\label{fig:supp_phone_live}
\end{figure*}
On phones, the learned reliability cap is off, because sparse ScanNet++ ground truth cannot label motion reliability; cue permissions and geometric verification stay on.
The reliability label needs ground truth within $30$\,ms at both ends of a $0.5$\,s motion, but ScanNet++ poses come at $3$ to $6$\,Hz, so only $75$ of the $40{,}555$ ScanNet++ training and validation keyframes can be scored, none of them reliable.
\begin{table*}[t]
\centering
\footnotesize
\setlength{\tabcolsep}{3.5pt}
\caption{Collaboration traffic. (a)~Encoded message sizes. (b)~Traffic per agent in replays on the four MP3D VAL houses and in live Wi-Fi sessions.}
\label{tab:supp_comm}
\scriptsize
\begin{tabular}{@{}l@{\hspace{1.4em}}l@{}}
\begin{tabular}[t]{@{}llr@{}}
\multicolumn{3}{@{}l}{\textbf{(a) Message sizes}} \\
\toprule
Message & Part & KB \\
\midrule
Announcement & descriptor/observation fields & $0.9$ \\
 & float16 embedding block & $1.6$ \\
Evidence packet & thumbnail ($160\times120$ gray) & $19.2$ \\
 & $300$ depth points & $3.6$ \\
 & $500$ features with depth, descriptors & $22$ \\
 & embedded announcement & $2.5$ \\
 & bound & ${\approx}48$ \\
\bottomrule
\end{tabular}
&
\begin{tabular}[t]{@{}>{\raggedright\arraybackslash}p{0.17\textwidth}>{\raggedright\arraybackslash}p{0.2\textwidth}>{\raggedright\arraybackslash}p{0.2\textwidth}@{}}
\multicolumn{3}{@{}l}{\textbf{(b) Traffic per agent}} \\
\toprule
Setting & Sent & Messages / losses \\
\midrule
Replay, one peer & $0.12$--$0.21$\,MB/s & -- \\
Replay, two peers & $0.26$--$0.56$\,MB/s (mean $0.38$; $6.9$--$16.9$\,MB, ${\approx}82$\,KB per own keyframe) & announcements ${\approx}5$\,KB per own keyframe \\
Descriptor-only, two peers & $0.006$--$0.016$\,MB/s & almost no packets requested \\
Disconnect window & $+18$--$23$\,\% (descriptor resync) & no false constraint \\
$10$\,\% message loss & $2.4$--$3.0$\,MB dropped per house & no false constraint \\
Live, phone + simulator, $104$\,s & $25$\,KB/s ($2.60$\,MB); received $0.57$\,MB & $498$ sent ($328$ announcements, $51$ evidence packets), $307$ received; none lost or shed \\
Live, two iPhones, first session, $2$\,min & $64$ and $31$\,KB/s ($7.8$ and $3.7$\,MB) & -- \\
Live, iPhone~15 + iPhone~11 Pro Max, three sessions of $2$--$2.5$\,min & $35$--$109$\,KB/s per phone and direction & tracking keeps its pose time \\
\bottomrule
\end{tabular}
\end{tabular}
\end{table*}
Unscorable keyframes count as unreliable, so all $31{,}109$ phone training and validation queries carry this label, and the head scores phone keyframes near zero (median $0.0003$, against a cue-based geometry confidence of ${\approx}0.7$); with the cap on, no link passes (part a), although each agent alone tracks to between $1.9$ and $3.4$\,cm ATE.
Anchoring the label at the nearest ground-truth sample, with motion in the IMU frame, would mark $56.5\,\%$ of $6{,}721$ eligible keyframes reliable; retraining with it is future work.
With the cap off and verification first, none of the $54$ scored VAL links is false (worst $2.3$\,cm and $0.4^\circ$; exact $95\,\%$ upper bound $6.6\,\%$); picture (a) plots the keyframes of the three agents (navy, red, light blue) with all $671$ links.

\paragraph{Phones without LiDAR.}
These phones use the multi-view path (\cref{sec:method_fusion}), because their stereo depth disagrees between views by $0.23$ to $2.6$\,m, against $0.016$ to $0.15$\,m for LiDAR at the same baselines; its stronger gyroscope weighting lowers the end-of-walk rotation error from $39$--$48^\circ$ to $3$--$4^\circ$.
Picture (b) plots its $107$ links on the VAL scan, with false links in red.
Four live sessions of about two minutes ran without the memory (Fig.~\ref{fig:supp_phone_live}): S1 pairs the iPhone~11 Pro Max with an iPhone~13, and S2 to S4 pair the iPhone~15 with the iPhone~11 Pro Max, held together by hand in S3 and rigidly joined in S4.
They verified $535$ links over Wi-Fi (panel c); each pair aligned after $17$ to $38$\,s, and in S2 to S4 the first alignment lies within $6.0$\,cm and $3.1^\circ$ of the final one.
Phones held together give an external check, because their true offset is small and constant: rigidly joined, the estimated separation stays between $7.8$ and $11.1$\,cm (median $9.8$\,cm), the height difference within $2.0$\,cm (90th percentile) and the tilt difference at $3.0^\circ$ (median) for $117$\,s (panels a and b); held by hand, the median separation is $12.3$\,cm and the median height difference $1.2$\,cm.
S1 had no such reference; there, the two phones' alignments to each other agree in rotation but compose to $79.5$\,cm in translation, an offset we could not check.
Panel (a) also renders both merged maps from one pose on the iPhone~15 path in S4, and picture (c) of Table~\ref{tab:supp_phone_collab} shows both app screens during pairing.

\paragraph{Memory on the Device.}
Exported to Core ML in half precision, the DINOv3 embedder reproduces the reference embeddings (cosine similarity at least $0.99996$ on $30$ phone frames) and takes $10.4$\,ms per keyframe on the Apple M5 Neural Engine, which runs $448$ of the model's $449$ operations.
Replayed on the M5 under one verifier, learned retrieval with these embeddings proposes candidates for all $330$ keyframes exchanged on a separate walk, against $1$ of $330$ with the handcrafted descriptor (panel d), and the complete pipeline verifies $206$ links on the held-together recordings, against $189$ (panel e).

\subsection{\texorpdfstring{Communication}{Communication}}
\label{sec:supp_comm}
Table~\ref{tab:supp_comm} counts the size of every encoded collaboration message at the application layer.
A receiver requests at most one evidence packet per peer keyframe and checks all selected candidates of that keyframe against this one packet.

In Table~\ref{tab:supp_comm}, feature descriptors have $256$ bits, and replays use three agents, walks of $27$ to $30$\,s at $30$\,Hz, $3.3$ to $6.3$ keyframes per second per agent and a loopback connection. A dash marks unrecorded or inapplicable values.

Every setting stays far below the nominal budget of $2$\,MB/s, and evidence packets make up over $90$\,\% of replay traffic.
Total team traffic grows as $N(N-1)$ for $N$ agents: with two peers, an agent sends $1.9$ to $2.6\times$ as much as with one, because each peer receives every announcement and requests its own packets.
Because packets are requested before verification, traffic does not depend on whether links are accepted; the house without an accepted link exchanges $0.26$ to $0.44$\,MB/s per agent.

\subsection{Single-Agent Results on Phones and Replica}
\label{sec:supp_extended_results}
Table~\ref{tab:supp_extended_results} gives the full results behind Table~\ref{tab:exp_replica} (\cref{sec:exp_single}): every CADENCE row, every baseline run on the same replays and all eleven published Replica-8 systems.
The published MA-Replica rows average the ATE of both agents over the four scenes and score rendered training views; MAGiC-SLAM and CoMA-SLAM render after $3000$ refinement iterations on the merged map, MAC-Ego3D renders per agent, and CP-SLAM reports no rendering.
The CADENCE renders on MA-Replica are instead the online training views that each agent observes during mapping.
\begin{table*}[t]
\centering
\footnotesize
\setlength{\tabcolsep}{2.6pt}
\setlength{\fboxsep}{1pt}
\renewcommand{\arraystretch}{0.9}
\caption{Extended results of CADENCE, same-replay baselines and published systems under their own protocols.}
\label{tab:supp_extended_results}
\begin{tabular*}{\textwidth}{@{\extracolsep{\fill}}llrrcccc@{}}
\toprule
Method & Input & ATE$\downarrow$ & med.$\downarrow$ & PSNR$\uparrow$ & SSIM$\uparrow$ & LPIPS$\downarrow$ & Dataset \\
\midrule
\groupband{8}{(a) Replica-8, 8 scenes, no refinement unless noted}
\oursrow\textbf{CADENCE online} & D & 0.41 & -- & 38.50 & 0.975 & 0.038 & Replica-8 \\
\oursrow CADENCE online, final poses & D & 0.37 & -- & 38.96 & 0.977 & 0.036 & Replica-8 \\
\oursrow CADENCE refined ($30$k it.) & D & 0.41 & -- & 39.16 & 0.977 & 0.033 & Replica-8 \\
\oursrow CADENCE online & D+I (synth.) & 0.26 & -- & 37.40 & 0.970 & 0.042 & Replica-8 \\
\oursrow CADENCE-RT paced & D & 0.41 & -- & 33.61 & 0.937 & 0.121 & Replica-8 \\
\groupband{8}{published, own protocols}
Point-SLAM~\cite{sandstrom2023point} & D & 0.52 & -- & 35.17 & 0.975 & 0.124 & Replica-8 \\
SplaTAM~\cite{keetha2024splatam} & D & 0.36 & -- & 34.11 & 0.97 & 0.10 & Replica-8 \\
MonoGS~\cite{matsuki2024gaussian} & D & 0.58 & -- & 37.50 & 0.960 & 0.070 & Replica-8 \\
Photo-SLAM~\cite{huang2024photoslam} & D & 0.60 & -- & 34.96 & 0.942 & 0.059 & Replica-8 \\
RTG-SLAM~\cite{peng2024rtgslam} & D & 0.18 & -- & 35.43 & 0.982 & 0.109 & Replica-8 \\
LoopSplat~\cite{zhu2025loopsplat} & D & 0.26 & -- & 36.63 & 0.985 & 0.112 & Replica-8 \\
GS-ICP SLAM~\cite{ha2024rgbd} & D & 0.16 & -- & 38.83 & 0.975 & 0.041 & Replica-8 \\
SEGS-SLAM~\cite{wen2025segsslam}$^\ddagger$ & D & 0.43 & -- & 39.42 & 0.975 & 0.021 & Replica-8 \\
GauS-SLAM~\cite{su2026gausslam} & D & 0.06 & -- & 40.25 & 0.991 & 0.027 & Replica-8 \\
Gaussian-SLAM~\cite{yugay2023gaussianslam} & D & 0.31 & -- & 42.08 & 0.996 & 0.018 & Replica-8 \\
SGAD-SLAM~\cite{hu2026sgadslam} & D & 0.16 & -- & 44.87 & 0.998 & 0.021 & Replica-8 \\
\midrule
\groupband{8}{(b) ScanNet++ iPhone, VAL: decision split, 6 scenes}
\oursrow CADENCE online & D+I+P & \second{5.51} & \second{3.79} & \second{21.75} & \second{0.804} & \second{0.302} & VAL \\
\oursrow CADENCE refined & D+I+P & \best{\textbf{5.25}} & \best{\textbf{3.63}} & \best{\textbf{25.01}} & \best{\textbf{0.855}} & \best{\textbf{0.192}} & VAL \\
\oursrow CADENCE online & D+I & 5.61 & 4.43 & 21.30 & 0.795 & 0.322 & VAL \\
\oursrow CADENCE online & D & 13.76$^\dagger$ & 13.22 & 18.66 & 0.741 & 0.413 & VAL \\
ARKit & device & 1.75 & 1.72 & -- & -- & -- & VAL \\
ORB-SLAM3~\cite{campos2021orb} & D & 3.38 & 2.58 & -- & -- & -- & VAL \\
ORB-SLAM3 (1/6) & D+I & 3.83$^\dagger$ & 3.83 & -- & -- & -- & VAL \\
GS-ICP SLAM & D & 78.06 & 75.59 & 18.52 & 0.718 & 0.569 & VAL \\
Photo-SLAM final map & D & 3.94 & 2.74 & 24.40 & 0.858 & 0.229 & VAL \\
Photo-SLAM own KF$^*$ & D & -- & -- & 25.69 & 0.875 & 0.201 & VAL \\
\bottomrule
\end{tabular*}
\end{table*}

In Table~\ref{tab:supp_extended_results}, ATE is the mean in cm (median where reported), and PSNR, SSIM and LPIPS use shared evaluation frames. Inputs: D is RGB-D, I raw IMU and P the ARKit prior. Bold and underline mark the best and second-best values within each block of equal input. A dagger marks a mean that includes a scene with coverage below $95\,\%$; an asterisk marks Photo-SLAM scores on its own sparse keyframe (KF) views, which are not directly comparable; a double dagger marks novel views averaged over five runs; and ($n$/$N$) counts completed scenes.

\begin{table*}[t]
\centering
\footnotesize
\setlength{\tabcolsep}{2.6pt}
\setlength{\fboxsep}{1pt}
\renewcommand{\arraystretch}{0.9}
{\small Table~\ref{tab:supp_extended_results} (continued).}\par\smallskip
\begin{tabular*}{\textwidth}{@{\extracolsep{\fill}}llrrcccc@{}}
\toprule
Method & Input & ATE$\downarrow$ & med.$\downarrow$ & PSNR$\uparrow$ & SSIM$\uparrow$ & LPIPS$\downarrow$ & Dataset \\
\midrule
\groupband{8}{(c) ScanNet++ iPhone, TEST3: final held-out split, 12 scenes}
\oursrow CADENCE online & D+I+P & \second{6.98} & \second{5.63} & \second{20.99} & \second{0.811} & \second{0.327} & TEST3 \\
\oursrow CADENCE refined & D+I+P & \best{\textbf{6.81}} & \best{\textbf{5.29}} & \best{\textbf{24.04}} & \best{\textbf{0.850}} & \best{\textbf{0.213}} & TEST3 \\
\oursrow CADENCE online & D & 41.04$^\dagger$ & 29.77 & 17.46 & 0.744 & 0.444 & TEST3 \\
ARKit & device & 4.16 & 3.16 & -- & -- & -- & TEST3 \\
ORB-SLAM3 & D & 19.57$^\dagger$ & 13.34 & -- & -- & -- & TEST3 \\
ORB-SLAM3 (11/12) & D+I & 5.82$^\dagger$ & 3.76 & -- & -- & -- & TEST3 \\
GS-ICP SLAM & D & 83.47 & 89.03 & 17.97 & 0.751 & 0.544 & TEST3 \\
Photo-SLAM final map & D & 18.31$^\dagger$ & 13.31 & 22.87$^\dagger$ & 0.840 & 0.279 & TEST3 \\
Photo-SLAM own KF$^*$ & D & -- & -- & 24.78$^\dagger$ & 0.872 & 0.229 & TEST3 \\
\midrule
\groupband{8}{(d) ScanNet++ iPhone, TEST2: held-out split, 12 scenes}
\oursrow CADENCE online & D+I+P & 4.99 & 3.03 & 20.18 & 0.792 & 0.341 & TEST2 \\
\oursrow CADENCE online & D+P & 11.70$^\dagger$ & 3.68 & 19.06 & 0.772 & 0.379 & TEST2 \\
ARKit & device & 1.88 & 1.54 & -- & -- & -- & TEST2 \\
ORB-SLAM3 (11/12) & D & 4.21$^\dagger$ & 2.73 & -- & -- & -- & TEST2 \\
ORB-SLAM3 (3/12) & D+I & 1.00$^\dagger$ & 0.96 & -- & -- & -- & TEST2 \\
GS-ICP SLAM & D & 95.94 & 89.32 & 18.24 & 0.743 & 0.521 & TEST2 \\
Photo-SLAM final map & D & 3.55$^\dagger$ & 2.85 & 22.70$^\dagger$ & 0.833 & 0.286 & TEST2 \\
Photo-SLAM own KF$^*$ & D & -- & -- & 24.84$^\dagger$ & 0.861 & 0.242 & TEST2 \\
\midrule
\groupband{8}{(e) ScanNet++ iPhone, TEST: first held-out split, 11 valid scenes}
\oursrow CADENCE online & D+I & 14.41 & 7.84 & 17.89 & 0.728 & 0.437 & TEST \\
\oursrow CADENCE online & D & 85.02$^\dagger$ & 64.10 & 14.49 & 0.671 & 0.545 & TEST \\
ARKit & device & 2.30 & 2.47 & -- & -- & -- & TEST \\
ORB-SLAM3 & D & 17.25$^\dagger$ & 7.61 & -- & -- & -- & TEST \\
ORB-SLAM3 (4/11) & D+I & 3.90$^\dagger$ & 2.08 & -- & -- & -- & TEST \\
GS-ICP SLAM & D & 129.34 & 102.49 & 17.50 & 0.739 & 0.544 & TEST \\
Photo-SLAM final map & D & 22.21$^\dagger$ & 3.85 & 21.67$^\dagger$ & 0.815 & 0.331 & TEST \\
Photo-SLAM own KF$^*$ & D & -- & -- & 23.73$^\dagger$ & 0.847 & 0.274 & TEST \\
\midrule
\groupband{8}{(f) MA-Replica, two agents, four scenes; CADENCE online, no refinement}
\oursrow CADENCE Merge (ATE-KF) & D & 0.35 & -- & 37.26 & 0.957 & 0.07 & MA-Replica \\
\oursrow CADENCE Merge (ATE-KF) & D+I (synth.) & 0.20 & -- & 37.39 & 0.965 & 0.063 & MA-Replica \\
\groupband{8}{published, own protocols}
CP-SLAM~\cite{hu2023cpslam} & D & 1.23 & -- & -- & -- & -- & MA-Replica \\
MAGiC-SLAM~\cite{yugay2025magicslam} & D & 0.25 & -- & 34.26 & 0.97 & 0.12 & MA-Replica \\
MAC-Ego3D~\cite{xu2025macego3d} & D & 0.14 & -- & 40.04 & 0.977 & 0.050 & MA-Replica \\
CoMA-SLAM~\cite{chen2026comaslam} (3/4) & D & 0.18 & -- & 38.38 & 0.99 & 0.05 & MA-Replica \\
OminiMAG-SLAM~\cite{ding2026ominimagslam} & D & 0.20 & -- & 36.94 & 0.98 & 0.11 & MA-Replica \\
GRAND-SLAM~\cite{thomas2025grandslam} & D & 0.25 & -- & 41.35 & 0.99 & 0.04 & MA-Replica \\
\midrule
\groupband{8}{(g) Matterport3D, four VAL houses: independent agents, $60$ KF in total (no link); Share at full budget}
\oursrow CADENCE, 1 agent & D+I & -- & -- & 28.04 & -- & -- & MP3D \\
\oursrow CADENCE, 2 agents & D+I & -- & -- & \second{30.83} & -- & -- & MP3D \\
\oursrow CADENCE, 3 agents & D+I & -- & -- & \best{\textbf{30.84}} & -- & -- & MP3D \\
\oursrow CADENCE Share, full budget & D+I & -- & -- & \begin{tabular}[t]{@{}c@{}}$-0.45$ (2-ag.) /\\ $-0.70$ (3-ag.)\\ dB vs.\ Merge\end{tabular} & -- & -- & MP3D \\
\bottomrule
\end{tabular*}
\end{table*}

\begin{figure*}[t]
\centering
\includegraphics[width=\textwidth,trim=0 73bp 0 0,clip]{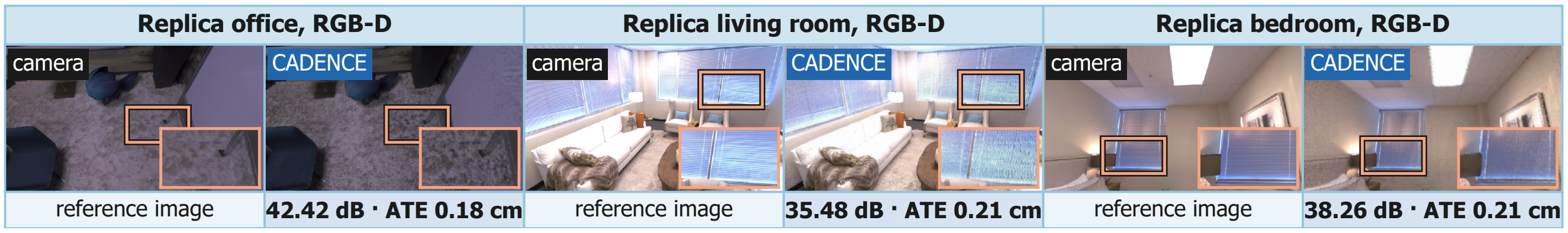}
\caption{Color reconstruction on Replica, as camera image and online RGB-D map of three rooms with scene PSNR and ATE.}
\label{fig:supp_replica_recon}
\end{figure*}

Figure~\ref{fig:supp_replica_recon} shows three Replica maps, with insets enlarging the most textured window; Fig.~\ref{fig:exp_rgb_reconstruction} shows the phone scenes. CADENCE returns a trajectory and a map in every phone scene.
With the IMU and the ARKit prior, CADENCE has a lower trajectory error than Photo-SLAM in $7$ of $12$ TEST3 scenes.
Its error stays above that of ARKit, the phone's own tracker, in all twelve scenes, because CADENCE uses ARKit only as a gated frame-to-frame motion prior during tracking.
In the RGB-D-only control, CADENCE tracks fewer than $95$\,\% of frames in four TEST3, three TEST2, seven TEST and one VAL scene ($^\dagger$ in Tables~\ref{tab:supp_extended_results} and~\ref{tab:supp_phone_scenes}). With this input, CADENCE has a lower ATE than Photo-SLAM in $0$ of $6$, $0$ of $11$ and $1$ of $12$ scenes on VAL, TEST and TEST3, ORB-SLAM3 also has a lower ATE on all three splits, and CADENCE is lower than GS-ICP SLAM in $6$ of $6$, $8$ of $11$ and $11$ of $12$ scenes.
\begin{table}[htb]
\centering
\footnotesize
\setlength{\tabcolsep}{2.6pt}
\caption{Inertial input per phone split, as CADENCE online mean ATE (cm) without and with the IMU of the phone.}
\label{tab:supp_imu_reads}
\scriptsize
\setlength{\tabcolsep}{2.6pt}
\begin{tabular}{@{}lllc>{\raggedright\arraybackslash}p{0.33\columnwidth}@{}}
\toprule
Split & Without & With & B/W & Factor changed \\
\midrule
VAL (6) & D $13.76^\dagger$ & D+I $5.61$ & 6/0 & IMU alone, no prior \\
TEST (11) & D $85.02^\dagger$ & D+I $14.41$ & 10/1 & IMU alone, no prior, first frozen configuration \\
TEST2 (12) & D+P $11.70^\dagger$ & D+I+P $4.99$ & 8/3 & IMU alone, prior in both arms \\
TEST3 (12) & D $41.04^\dagger$ & D+I+P $6.98$ & 12/0 & IMU and prior together \\
\bottomrule
\end{tabular}
\end{table}

In Table~\ref{tab:supp_imu_reads}, parentheses hold the scene counts, and B/W counts the scenes that get better or worse (ties below $0.05$\,cm excluded). A dagger marks coverage below $95\,\%$.

\begin{figure*}[t]
\centering
\captionsetup[sub]{labelformat=parens,font=footnotesize,skip=0pt}%
\includegraphics[width=\textwidth]{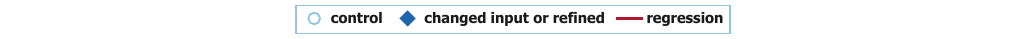}\\[1pt]
\begin{subfigure}[t]{0.2\textwidth}\centering\includegraphics[width=\linewidth]{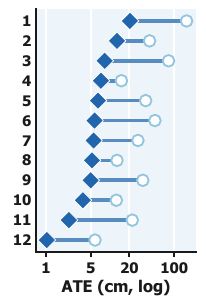}\caption{}\end{subfigure}%
\begin{subfigure}[t]{0.2\textwidth}\centering\includegraphics[width=\linewidth]{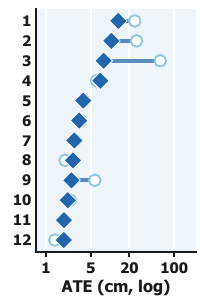}\caption{}\end{subfigure}%
\begin{subfigure}[t]{0.2\textwidth}\centering\includegraphics[width=\linewidth]{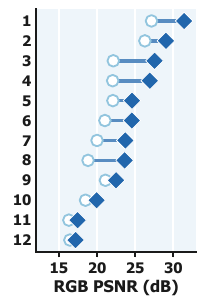}\caption{}\end{subfigure}%
\begin{subfigure}[t]{0.2\textwidth}\centering\includegraphics[width=\linewidth]{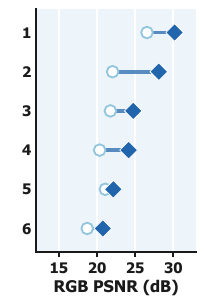}\caption{}\end{subfigure}%
\begin{subfigure}[t]{0.2\textwidth}\centering\includegraphics[width=\linewidth]{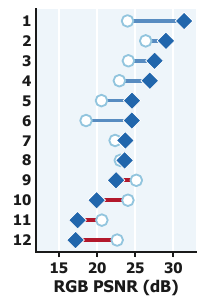}\caption{}\end{subfigure}
\caption{Paired per-scene phone effects. (a)~TEST3 ATE without and with raw IMU and ARKit prior. (b)~TEST2 without and with raw IMU. (c)~TEST3 and (d)~VAL online vs.\ refined PSNR. (e)~TEST3 refined CADENCE vs.\ Photo-SLAM.}
\label{fig:supp_input_metrics}
\label{fig:supp_refinement}
\end{figure*}

Adding the IMU lowers mean ATE on every split (Table~\ref{tab:supp_imu_reads}).
VAL, TEST and TEST2 change the IMU alone, whereas TEST3 adds the IMU and the ARKit prior together, so the larger TEST3 gain is not due to the IMU alone.
Refinement changes poses, color and exposure together; on TEST3 it raises PSNR with a paired interval of $[+2.19,+3.92]$\,dB, at $1.63\times$ the cost of the online run.
Table~\ref{tab:supp_phone_scenes} and Fig.~\ref{fig:supp_input_metrics} give the per-scene values and paired effects.
\begin{table*}[t]
\centering
\scriptsize
\setlength{\tabcolsep}{4.2pt}
\renewcommand{\arraystretch}{0.82}
\caption{Per-scene phone results, sorted by CADENCE online PSNR.}
\label{tab:supp_phone_scenes}
\label{fig:exp_grid}
\label{fig:supp_grids}
\label{fig:supp_grid_test}
\label{fig:supp_grid_val}
\begin{tabular}{@{}rrrrr@{\hspace{9pt}}rrrrrrrr@{}}
\toprule
 & \multicolumn{4}{c}{RGB PSNR (dB)\,$\uparrow$} & \multicolumn{8}{c}{ATE (cm)\,$\downarrow$} \\
\cmidrule(lr){2-5}\cmidrule(l){6-13}
\# & \textcolor{cvprblue}{\textbf{online}} & \textcolor{cvprblue}{\textbf{refined}} & Photo & GS-ICP & \textcolor{cvprblue}{\textbf{online}} & \textcolor{cvprblue}{\textbf{refined}} & \textcolor{cvprblue}{\textbf{control}} & ARKit & ORB-I & ORB-D & Photo & GS-ICP \\
\midrule
\groupband{13}{TEST3: online = D+I+P, control = D, 12 scenes}
1 & \textbf{27.1} & \textbf{31.4} & 24.0 & 22.4 & \textbf{2.3} & \textbf{2.4} & 22.2$^\dagger$ & 1.5 & 1.8$^\dagger$ & 63.0 & 63.2 & 23.6 \\
2 & \textbf{26.3} & \textbf{29.0} & 26.4 & 16.8 & \textbf{1.0} & \textbf{1.2} & 5.8 & 0.7 & 1.2$^\dagger$ & 2.2 & 3.1 & 90.4 \\
3 & \textbf{22.1} & \textbf{27.5} & 24.1 & 16.6 & \textbf{5.7} & \textbf{5.7} & 50.0 & 2.6 & 3.3 & 20.4 & 17.7 & 73.8 \\
4 & \textbf{22.1} & \textbf{24.6} & 20.6$^\dagger$ & 19.4 & \textbf{12.8} & \textbf{12.8} & 41.3$^\dagger$ & 12.4 & 15.2$^\dagger$ & 31.4$^\dagger$ & 31.9$^\dagger$ & 42.8 \\
5 & \textbf{22.0} & \textbf{26.9} & 22.9 & 17.4 & \textbf{3.8} & \textbf{3.8} & 12.5 & 1.5 & 1.8$^\dagger$ & 5.1 & 3.3 & 51.1 \\
6 & \textbf{21.1} & \textbf{22.5} & 25.2 & 17.1 & \textbf{5.5} & \textbf{4.8} & 27.2 & 1.9 & 4.4 & 4.1 & 3.6 & 145.4 \\
7 & \textbf{21.0} & \textbf{24.6} & 18.6$^\dagger$ & 18.2 & \textbf{8.2} & \textbf{8.0} & 81.5$^\dagger$ & 4.6 & n.c. & 38.1$^\dagger$ & 37.9$^\dagger$ & 99.3 \\
8 & \textbf{20.0} & \textbf{23.7} & 22.4$^\dagger$ & 18.1 & \textbf{20.4} & \textbf{20.2} & 155.8$^\dagger$ & 9.7 & 21.6$^\dagger$ & 19.5$^\dagger$ & 18.8$^\dagger$ & 87.7 \\
9 & \textbf{18.8} & \textbf{23.6} & 23.0 & 18.3 & \textbf{5.0} & \textbf{4.2} & 32.4 & 4.8 & 2.6$^\dagger$ & 7.1 & 8.9 & 54.8 \\
10 & \textbf{18.5} & \textbf{20.0} & 24.0 & 17.8 & \textbf{5.2} & \textbf{4.8} & 12.8 & 3.7 & 4.0 & 3.5 & 3.7 & 112.7 \\
11 & \textbf{16.4} & \textbf{17.2} & 22.6 & 17.0 & \textbf{6.5} & \textbf{6.5} & 35.9 & 3.2 & 4.3 & 36.5$^\dagger$ & 24.1 & 99.7 \\
12 & \textbf{16.3} & \textbf{17.5} & 20.6 & 16.4 & \textbf{7.2} & \textbf{7.2} & 15.0 & 3.2 & 3.8 & 3.8 & 3.7 & 120.3 \\
\midrule
\groupband{13}{TEST2: online = D+I+P, control = D+P, 12 scenes}
1 & \textbf{22.4} & -- & 25.8 & 18.1 & \textbf{1.9} & -- & 1.4 & 0.9 & 1.4$^\dagger$ & 2.0 & 2.0 & 27.7 \\
2 & \textbf{22.0} & -- & 25.6 & 18.7 & \textbf{2.2} & -- & 2.4 & 1.4 & n.c. & 2.7 & 2.5 & 33.3 \\
3 & \textbf{21.9} & -- & 25.5 & 19.4 & \textbf{1.9} & -- & 1.9 & 1.5 & n.c. & 1.8 & 1.8 & 122.5 \\
4 & \textbf{21.4} & -- & 24.6 & 16.9 & \textbf{2.5} & -- & 5.8 & 1.1 & n.c. & 1.5 & 1.5 & 84.2 \\
5 & \textbf{21.3} & -- & 24.6 & 21.3 & \textbf{7.0} & -- & 6.2 & 2.2 & n.c. & 4.0 & 4.4 & 54.0 \\
6 & \textbf{20.9} & -- & 23.9 & 16.6 & \textbf{2.8} & -- & 2.9$^\dagger$ & 2.4 & n.c. & 2.4 & 7.2 & 33.2 \\
7 & \textbf{20.7} & -- & 21.9 & 18.4 & \textbf{3.3} & -- & 3.5 & 1.3 & n.c. & 3.1 & 2.8 & 181.4 \\
8 & \textbf{20.5} & -- & 25.2 & 16.5 & \textbf{2.6} & -- & 2.0 & 1.2 & n.c. & 1.7 & 1.9 & 21.8 \\
9 & \textbf{19.8} & -- & 23.5 & 19.3 & \textbf{3.8} & -- & 3.9$^\dagger$ & 2.5 & 1.0$^\dagger$ & 4.2 & 4.7 & 94.5 \\
10 & \textbf{19.7} & -- & 22.2$^\dagger$ & 21.2 & \textbf{7.9} & -- & 60.4$^\dagger$ & 1.5 & 0.7$^\dagger$ & 3.5$^\dagger$ & 2.9$^\dagger$ & 95.2 \\
11 & \textbf{16.3} & -- & 16.4 & 16.7 & \textbf{10.5} & -- & 25.9 & 3.4 & n.c. & n.c. & 5.3 & 247.0 \\
12 & \textbf{15.3} & -- & 13.3 & 15.8 & \textbf{13.4} & -- & 24.3 & 3.2 & n.c. & 19.3 & 5.6 & 156.6 \\
\midrule
\groupband{13}{TEST: online = D+I, control = D, 11 scenes}
1 & \textbf{21.2} & -- & 20.4 & 20.2 & \textbf{5.8} & -- & 173.2$^\dagger$ & 3.5 & n.c. & 7.6 & 3.9 & 72.8 \\
2 & \textbf{21.0} & -- & 25.5 & 17.0 & \textbf{7.4} & -- & 9.0 & 1.2 & n.c. & 2.7 & 2.5 & 102.5 \\
3 & \textbf{19.5} & -- & 21.9 & 17.5 & \textbf{5.8} & -- & 40.7$^\dagger$ & 1.8 & 1.8$^\dagger$ & 35.9$^\dagger$ & 2.3 & 97.2 \\
4 & \textbf{19.4} & -- & 22.8$^\dagger$ & 18.6 & \textbf{6.0} & -- & 64.1$^\dagger$ & 2.7 & n.c. & 11.4$^\dagger$ & 10.3$^\dagger$ & 182.2 \\
5 & \textbf{18.5} & -- & 21.8 & 18.6 & \textbf{7.8} & -- & 230.4$^\dagger$ & 1.9 & 2.4$^\dagger$ & 53.4$^\dagger$ & 45.1 & 157.0 \\
6 & \textbf{18.4} & -- & 23.0$^\dagger$ & 15.3 & \textbf{3.5} & -- & 20.8$^\dagger$ & 1.2 & n.c. & 1.5$^\dagger$ & 1.5$^\dagger$ & 60.4 \\
7 & \textbf{18.4} & -- & 22.8 & 17.3 & \textbf{10.8} & -- & 12.4 & 0.9 & n.c. & 3.1 & 2.5 & 78.5 \\
8 & \textbf{17.6} & -- & 22.6$^\dagger$ & 17.7 & \textbf{11.8} & -- & 153.6$^\dagger$ & 2.6 & 1.7$^\dagger$ & 2.8$^\dagger$ & 2.7$^\dagger$ & 60.2 \\
9 & \textbf{16.2} & -- & 18.9 & 16.6 & \textbf{24.3} & -- & 14.7 & 2.5 & n.c. & 5.0 & 4.9 & 188.9 \\
10 & \textbf{14.7} & -- & 19.3 & 16.4 & \textbf{50.5} & -- & 146.8 & 3.7 & n.c. & 9.4 & 111.0 & 268.4 \\
11 & \textbf{11.9} & -- & 19.3$^\dagger$ & 17.3 & \textbf{25.0} & -- & 69.6$^\dagger$ & 3.4 & 9.8$^\dagger$ & 56.8$^\dagger$ & 57.6$^\dagger$ & 154.6 \\
\midrule
\groupband{13}{VAL: online = D+I+P, control = D, 6 scenes}
1 & \textbf{26.6} & \textbf{30.2} & 26.2 & 18.9 & \textbf{0.6} & \textbf{0.4} & 3.5 & 0.3 & n.c. & 1.7 & 1.6 & 34.5 \\
2 & \textbf{22.0} & \textbf{28.1} & 24.7 & 19.4 & \textbf{1.8} & \textbf{1.5} & 6.1 & 1.3 & n.c. & 1.5 & 1.5 & 88.3 \\
3 & \textbf{21.7} & \textbf{24.7} & 24.6 & 18.1 & \textbf{10.1} & \textbf{9.5} & 19.4 & 3.0 & 3.8$^\dagger$ & 2.7 & 3.0 & 150.6 \\
4 & \textbf{21.1} & \textbf{22.1} & 25.0 & 17.0 & \textbf{1.5} & \textbf{1.8} & 7.0 & 1.4 & n.c. & 4.0 & 4.1 & 43.7 \\
5 & \textbf{20.3} & \textbf{24.1} & 25.4 & 19.3 & \textbf{13.4} & \textbf{12.9} & 22.4 & 2.5 & n.c. & 2.5 & 2.5 & 63.5 \\
6 & \textbf{18.7} & \textbf{20.8} & 20.5 & 18.5 & \textbf{5.8} & \textbf{5.5} & 24.0$^\dagger$ & 2.0 & n.c. & 7.9 & 11.0 & 87.7 \\
\bottomrule
\end{tabular}
\end{table*}

In Table~\ref{tab:supp_phone_scenes}, the online and refined columns are CADENCE with the input stated in each block header, and control is CADENCE with the reduced input stated there. Photo is the saved Photo-SLAM map, and ORB-I and ORB-D are inertial and RGB-D ORB-SLAM3. A dagger marks coverage below $95\,\%$, n.c. marks a run without a completed result, and a dash marks a run that we did not perform.

Each panel of Fig.~\ref{fig:supp_input_metrics} sorts its rows separately, so row positions are ranks, not the scene numbers of Table~\ref{tab:supp_phone_scenes}. Open markers show runs without IMU and prior in (a), online runs in (c) and (d) and the saved Photo-SLAM map in (e). Filled markers show runs with IMU and prior in (a), refined runs in (c) and (d) and refined CADENCE in (e). Both arms of panel (b) use the prior, and red connectors mark scenes that get worse from one arm to the other.

\begin{figure*}[t]
\centering
\includegraphics[width=\textwidth]{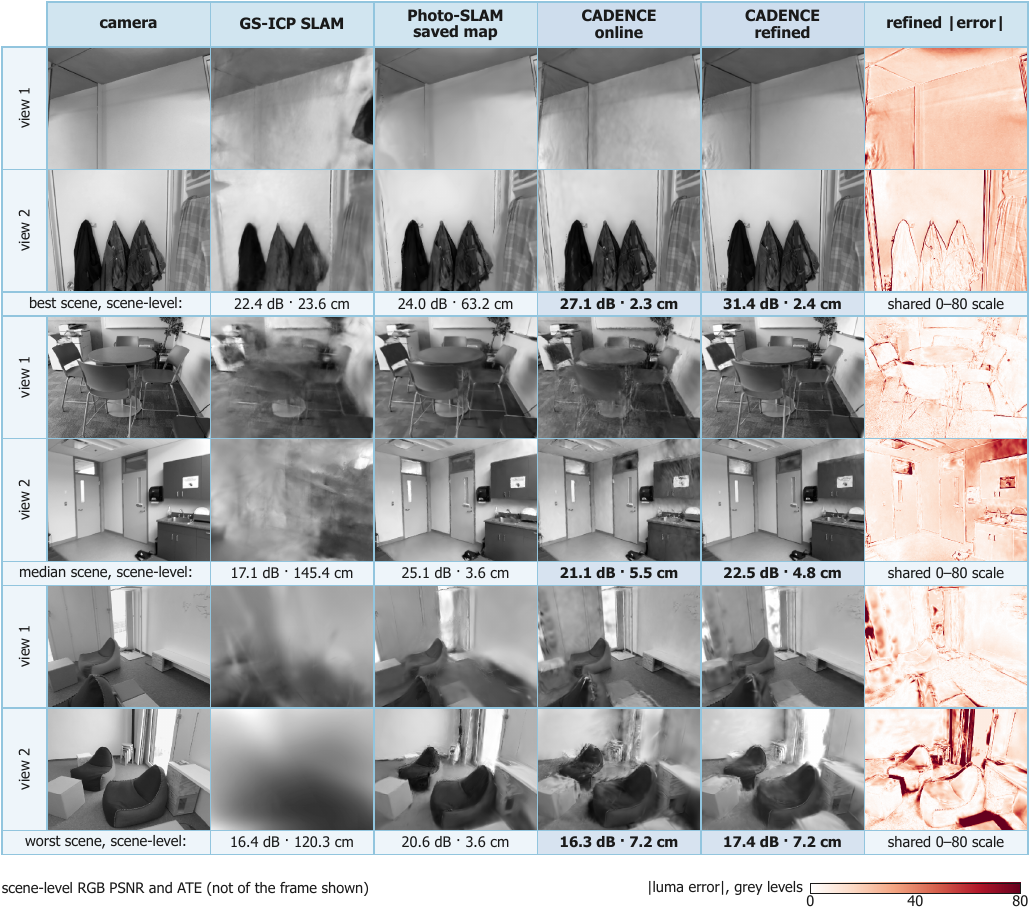}
\caption{Matched views on the final held-out split (TEST3) for the best, median and worst scenes by online PSNR, two views each, with refined luma error.}
\label{fig:supp_qual}
\label{fig:exp_qual}
\end{figure*}

Figure~\ref{fig:supp_qual} compares matched views of the best, median and worst TEST3 scenes.
Photo-SLAM renders better on the median and worst scenes and tracks better on the worst one.
Across scenes, render quality and trajectory error do not always agree (Table~\ref{tab:supp_phone_scenes}), so both should be read together.

All methods in Fig.~\ref{fig:supp_qual} are shown in luma, because the CADENCE panels contain luma only; the panel labels give the scene PSNR and ATE.

\begin{figure*}[t]
\centering
\includegraphics[width=\textwidth]{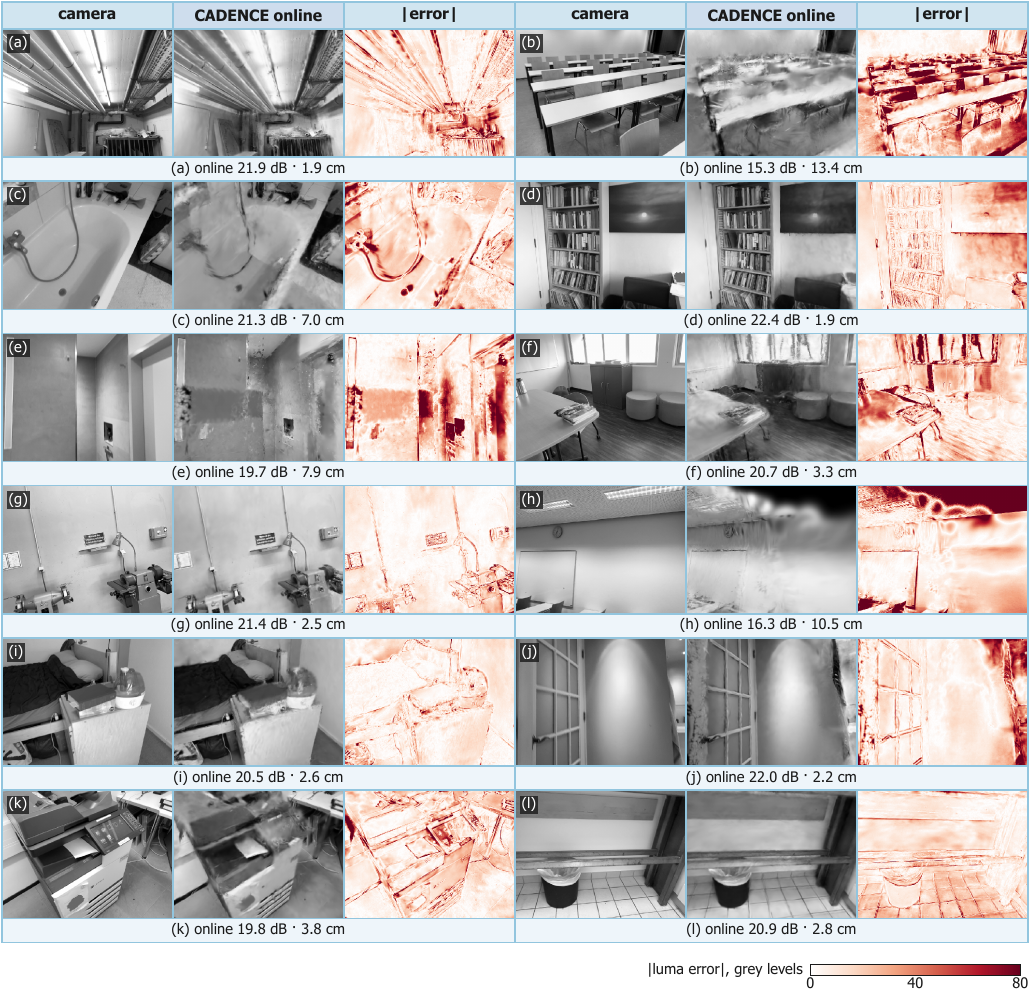}
\caption{All twelve scenes of the second held-out split (TEST2), with camera, CADENCE online, luma error, scene PSNR and ATE.}
\label{fig:supp_contact}
\end{figure*}

\begin{figure*}[t]
\centering
\includegraphics[width=\textwidth]{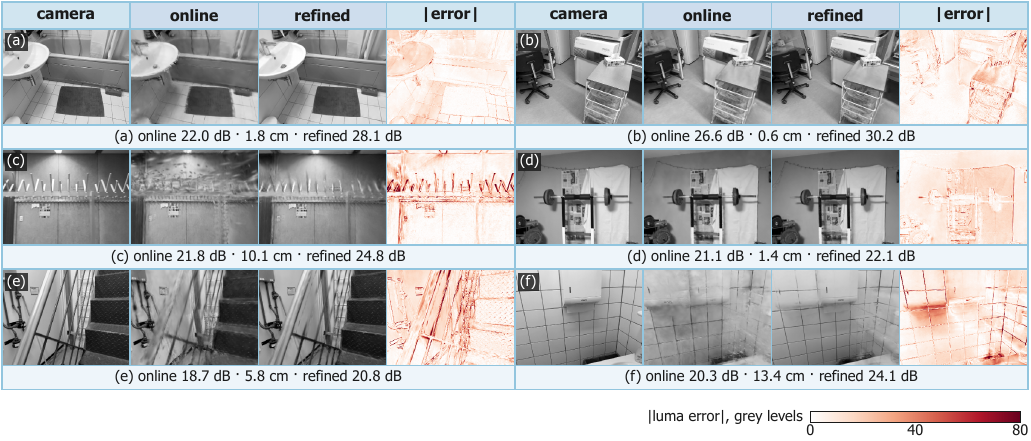}
\caption{All six VAL scenes, with camera image, CADENCE online and refined renders, refined luma error, scene PSNR and ATE.}
\label{fig:supp_contact_val}
\end{figure*}

Figures~\ref{fig:supp_contact} and~\ref{fig:supp_contact_val} present every scene of the second held-out split and of VAL, without frame selection.

\subsection{Runtime and Tracking Robustness}
This subsection extends \cref{sec:exp_runtime}.
Table~\ref{tab:supp_runtime} and Fig.~\ref{fig:exp_runtime} separate desktop checks of the real-time contract (pose-call p95 at most $33.3$\,ms, Eq.~\ref{eq:method_contract}) from Apple device measurements.
The real-time desktop configuration and the default M5 configuration (CPU mapping, no depth registration) meet the contract in every variant; the accuracy-first configuration does not.
Over 1660 deterministic ScanNet++ frames, M5 and desktop give identical tracked-frame, keyframe and job counts, and a tracking-only M5 run reproduces the M5 trajectory.
The $0.17$\,cm difference between the two devices therefore does not come from Gaussian optimization.
With matched schedules, the devices differ by $0.12$\,dB, and every tested Metal kernel meets its bound (Table~\ref{tab:supp_parity}) on M5 and iPhone.
Per $66.7$\,s sequence, evaluation included, the paced real-time run takes $4.3$\,min and the accuracy-first online run $213.5$\,min (Table~\ref{tab:supp_resources}c).
Real-time pacing of one VAL phone replay keeps tracking on time and trades map PSNR for it ($21.09$ to $17.14$\,dB), because mapping then gets less time per frame.
\label{sec:supp_runtime_detail}
Only the Replica tracker registers depth on the GPU; the phone engine does not.
When Metal mapping shares the M5 GPU with this registration, it exceeds the contract on two Replica sequences (Table~\ref{tab:supp_runtime}b).
On the A13 iPhone (Tables~\ref{tab:exp_device} and~\ref{tab:supp_device_runtime}a), the on-device benchmark (\cref{sec:supp_apple}) selects Metal mapping, which runs each job $1.76\times$ faster than the CPU optimizer in the 4\,GB profile, and the real-time replay ($661$ mapping jobs) stays under the tier's $150$\,k Gaussian cap.
In live stereo-depth sessions, the thermal governor lowers the input rate at the serious thermal state, and tracking continues.
The DINOv3-S/16 keyframe encoder (run offline in the evaluations) runs on the M5 Neural Engine, which tracking and mapping do not use, and occupies under $5$\,\% of it at the $5.7$ keyframes/s of the phone replay.
\begin{table*}[t]
\centering
\footnotesize
\setlength{\tabcolsep}{3.5pt}
\caption{On-device runtime. (a)~Phone engine. (b)~DINOv3-S/16 keyframe encoder with Core ML on the Apple M5. (c)~Single-agent configurations of the device measurements (\cref{sec:supp_profiles}).}
\label{tab:supp_device_runtime}
\label{tab:supp_profiles}
\begin{tabular}{@{}>{\raggedright\arraybackslash}p{0.22\textwidth}lccccc>{\raggedright\arraybackslash}p{0.12\textwidth}@{}}
\multicolumn{8}{@{}l}{\textbf{(a) Phone engine}} \\
\toprule
Run & Device & p50 (ms) & p95 (ms) & $>33.3$\,ms & Peak memory & Gaussians & Thermal \\
\midrule
Deterministic replay, full ScanNet++ sequence & M5 & -- & -- & -- & $2.4$\,GB & $383$\,k & -- \\
Real-time LiDAR replay, $600$ frames at $15$\,Hz & A13 & $14.3$ & $20.3$ & $1.0$\,\% & $1.0$\,GB & $144$\,k & -- \\
Six live stereo-depth sessions, $1.3$--$2.6$\,min ($5$ frames dropped in one) & A13 & $12.3$--$16.1$ & $19.5$--$32.2$ & $0.0$--$4.5$\,\% & -- & -- & serious in 4; reduced $3$--$88$\,s \\
Two live stereo-depth mapping sessions, iPhone~15, $0.6$--$1.1$\,min, no frame dropped & A16 & $7.1$--$7.4$ & $11.8$--$11.9$ & $0.0$\,\% & $1.7$\,GB & $230$--$237$\,k & fair \\
Three live two-phone sessions, $2$--$2.5$\,min, iPhone~15 & A16 & $6.5$--$6.9$ & $10.0$--$11.3$ & $0.0$--$0.3$\,\% & $2.6$\,GB & $579$--$590$\,k & serious \\
\quad same sessions, iPhone~11 Pro Max & A13 & $10.2$--$11.1$ & $17.0$--$26.7$ & $0.9$--$4.4$\,\% & $1.8$\,GB & $119$--$147$\,k & serious \\
\bottomrule
\end{tabular}

\vspace{0.5em}
\begin{tabular}{@{}l@{\hspace{2.5em}}l@{}}
\begin{tabular}[t]{@{}lcl@{}}
\multicolumn{3}{@{}l}{\textbf{(b) Keyframe encoder, Apple M5}} \\
\toprule
Compute unit & Time per keyframe (ms) & Note \\
\midrule
Neural Engine & $8.4$ p50 / $8.5$ p95 & not used by tracking or mapping \\
GPU & $5.8$ & -- \\
CPU & $19.7$ & -- \\
\bottomrule
\end{tabular}
&
\begin{tabular}[t]{@{}lccccc@{}}
\multicolumn{6}{@{}l}{\textbf{(c) Engine configurations of the device runs}} \\
\toprule
Profile & Color, $\lambda$ & $\kappa$ & Cap & $N_w{+}N_s$ & Tracker \\
\midrule
CPU mapping & RGB, 0 & 6 & -- & -- & -- \\
Replica & SH2, 0.2 & 3 & 1\,M & $4{+}4$ & R \\
TUM & RGB, 0 & 6 & 400\,k & $2{+}2$ & B \\
ScanNet++ & RGB, 0.2 & 6 & 400\,k & $4{+}4$ & -- \\
4\,GB tier & \multicolumn{4}{c}{overlay: cap 150\,k, residency 4} & -- \\
\bottomrule
\end{tabular}
\end{tabular}
\end{table*}

In Table~\ref{tab:supp_profiles}, part (a) reports pose-call times and the share of calls above the $33.3$\,ms contract. Part (b) uses an input of $336\times448$ pixels, which the evaluations process offline. In part (c), $\lambda$ denotes the SSIM weight, $\kappa$ denotes the seeding grid size in pixels and $N_w{+}N_s$ denotes the newest plus sampled views per job, where later Replica and phone profiles use $12{+}12$. The tracker additions R and B denote depth registration (\cref{eq:supp_reg}) and the combination of local model, inertial window, window bundle adjustment and loop closure. A dash marks inapplicable or unrecorded values.

Every iPhone run meets the pose-call contract, while mapping throughput limits phone map quality.
\begin{table*}[t]
\centering
\scriptsize
\setlength{\tabcolsep}{3.5pt}
\caption{Pose time and whole-run cost. (a)~Desktop, Replica-8 paced at $30$\,Hz. (b)~Apple M5, default configuration. (c)~Replica desktop resources of eight 2000-frame rooms on an RTX~5070.}
\label{tab:supp_runtime}
\label{tab:supp_resources}
\begin{tabular}{@{}l@{\hspace{2em}}r@{}}
\begin{tabular}[t]{@{}lrrr@{}}
\multicolumn{4}{@{}l}{\textbf{(a) Desktop, paced at $30$\,Hz}} \\
\toprule
Configuration & p50 (ms) & p95 (ms) & Late (\%) \\
\midrule
\oursrow CADENCE-RT\,$\cdot$\,RGB-D & 19.7 & 32.5 & 3.3 \\
\oursrow CADENCE-RT\,$\cdot$\,+ IMU (synth.) & 18.9 & 32.0 & 2.2 \\
\oursrow CADENCE\,$\cdot$\,RGB-D & 20.5 & 40.6 & 12.9 \\
\oursrow CADENCE\,$\cdot$\,+ IMU (synth.) & 19.5 & 35.4 & 8.8 \\
\bottomrule
\end{tabular}
&
\begin{tabular}[t]{@{}lrrr@{}}
\multicolumn{4}{@{}l}{\textbf{(b) Apple M5, p95 (ms)}} \\
\toprule
Mapping, WM & office3 & room0 & fr1\_desk \\
\midrule
CPU, off & 18.1 & 20.9 & 14.8 \\
CPU, on & 19.0 & 20.6 & 15.2 \\
Metal, off & 34.4 & 39.5 & 15.7 \\
Metal, on & 34.8 & 43.1 & 16.5 \\
\bottomrule
\end{tabular}
\\
\noalign{\vskip 0.4em}
\multicolumn{2}{@{}c@{}}{\setlength{\tabcolsep}{9pt}\begin{tabular}{@{}l@{}}
\textit{(c) Replica desktop resources} \\
\begin{tabular}{@{}lrrrrrr@{}}
\toprule
Configuration & Run time (min) & p50 (ms) & p95 (ms) & $1/$p50 (Hz) & Host RSS (GB) & Gaussians \\
\midrule
Accuracy / online & 213.5 & 43.8 & 57.8 & 22.8 & 10.47 & 803,994 \\
RT / paced & 4.3 & 19.7 & 30.6 & 50.7 & 5.73 & 284,373 \\
RT / drained & 14.2 & 25.0 & 41.4 & 40.0 & 5.65 & 216,741 \\
Accuracy / refined & 391.7 & 34.7 & 49.6 & 28.8 & 10.39 & 797,630 \\
\bottomrule
\end{tabular}
\end{tabular}}
\end{tabular}
\end{table*}

In Table~\ref{tab:supp_resources}, part (a) reports the mean p50, the worst-scene p95 and the late frames. Part (b) reports the p95 per sequence with the world memory (WM) off and on, where CPU denotes the default configuration without depth registration and Metal denotes the engine profile with GPU depth registration for Replica. Part (c) reports scene means: p95 is the mean per-scene percentile, and the run time is the wall-clock time per scene including mapping and evaluation, with the accuracy-first online and refined runs sharing one GPU. Host RSS is the resident CPU memory, and $1/$p50 is the pose-call rate, not the throughput.

\begin{figure*}[t]
\centering
\includegraphics[width=0.9\textwidth]{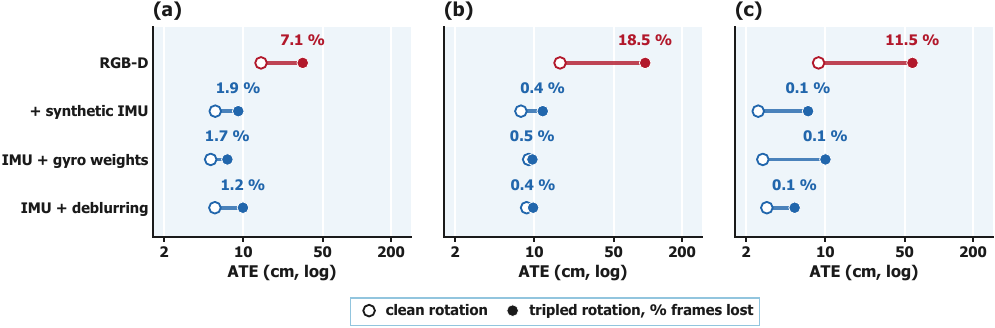}
\caption{Tracking under tripled rotation with synthetic IMU on Replica/TUM, as ATE with clean and tripled rotation on (a)~three DEV, (b)~three VAL and (c)~five held-out sequences.}
\label{fig:supp_rotation}
\end{figure*}

Figure~\ref{fig:supp_rotation} evaluates tracking under tripled rotation.
On five held-out sequences, synthetic IMU lowers ATE over tracked frames from $58.50$ to $7.04$\,cm ($60.02$ to $7.47$\,cm over all frames) and lost frames from $11.5$ to $0.1$\,\%, whereas gyroscope weighting and deblurring have only a minor effect.

In Fig.~\ref{fig:supp_rotation}, open markers denote clean rotation and filled markers denote tripled rotation, with labels for lost frames. Red denotes RGB-D input alone, whereas blue denotes the variants with inertial input.

\subsection{Independent Agents, Rigid Rigs and Multiple Floors}
\begin{figure*}[t]
\centering
\includegraphics[width=\textwidth]{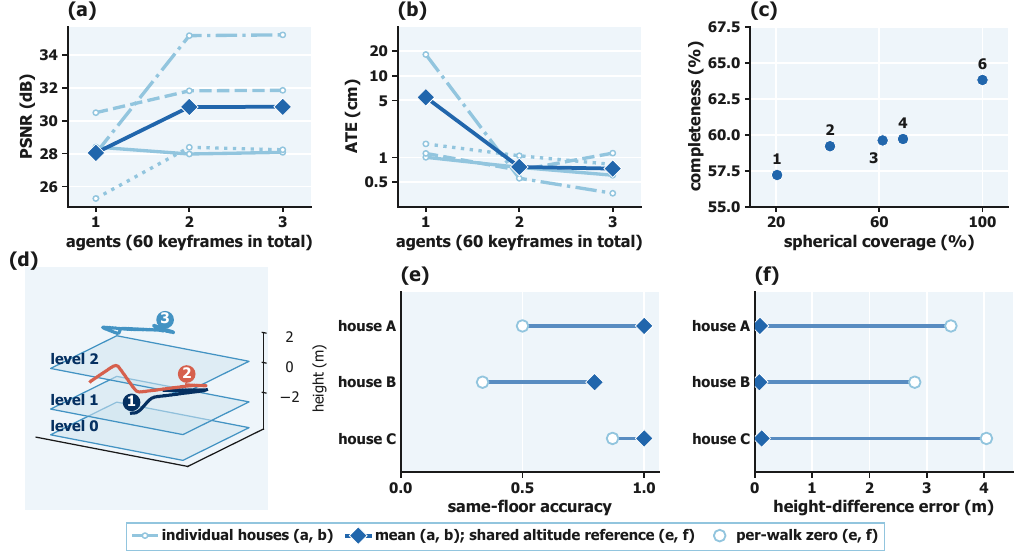}
\caption{Observation budget, camera layout and altitude reference. (a,b)~Independent agents at $60$ total keyframes on four MP3D houses. (c)~Synthetic Replica rigs. (d)~Three agents on three floors. (e,f)~Cross-agent floor metrics.}
\label{fig:supp_modes}
\label{fig:supp_floors}
\end{figure*}

Figure~\ref{fig:supp_modes} gathers three studies: independent agents at a fixed keyframe budget, a synthetic rigid camera rig, and a barometric altitude reference shared across agents.

In Fig.~\ref{fig:supp_modes}, panels (a) and (b) plot the PSNR and the ATE on a log scale for one to three independent agents without any accepted link, with light blue for the houses and dark blue for the mean. Panel (c) plots spherical coverage against completeness with labels for the camera count, and panel (d) presents the ground-truth paths in one simulated house, where slabs mark the floor levels. Panels (e) and (f) plot the cross-agent same-floor accuracy and the height-difference error, with open markers for per-walk zeros and filled markers for a shared altitude reference.

The agents with $60$ keyframes in total accept no link, so panels (a,b) measure the effect of more coverage, not of fusion.
A second agent adds $2.8$\,dB on average (one house gains $7.14$\,dB, one loses $0.42$\,dB), and much of the ATE reduction in (b) comes from one large single-agent drift.
Table~\ref{tab:supp_rig} lists every rig calibration arm, with each non-reference camera perturbed by $1^\circ$ and $10$\,mm.
For the two- and three-camera layouts, whose views do not overlap, refinement makes no update.
\begin{table*}[t]
\centering\footnotesize\setlength{\tabcolsep}{5pt}\renewcommand{\arraystretch}{1.0}
\caption{Synthetic rigid-rig controls on Replica room1 with known, perturbed and refined calibration.}
\label{tab:supp_rig}
\begin{tabular}{@{}llrrrrrrr@{}}
\toprule
Layout & Calibration & Cameras & Sphere (\%) & ATE (cm) & Complete (\%) & Ghost (\%) & Error ($^\circ$) & Error (mm) \\
\midrule
single & known & 1 & 20.4 & 1.07 & 57.2 & 17.5 & -- & -- \\
opposed2 & known & 2 & 40.9 & 0.44 & 59.2 & 15.2 & 0.00 & 0.00 \\
opposed2 & perturbed & 2 & 40.9 & 1.08 & 47.9 & 32.3 & 1.00 & 10.00 \\
opposed2 & refined & 2 & 40.9 & 1.08 & 47.9 & 32.3 & 1.00 & 10.00 \\
outward3 & known & 3 & 61.3 & 0.33 & 59.6 & 16.4 & 0.00 & 0.00 \\
outward3 & perturbed & 3 & 61.3 & 1.37 & 57.7 & 18.5 & 1.00 & 10.00 \\
outward3 & refined & 3 & 61.3 & 1.37 & 57.7 & 18.5 & 1.00 & 10.00 \\
cam4 & known & 4 & 69.2 & 0.46 & 59.7 & 15.7 & 0.00 & 0.00 \\
cam4 & perturbed & 4 & 69.2 & 1.13 & 50.1 & 31.5 & 1.00 & 10.00 \\
cam4 & refined & 4 & 69.2 & 0.74 & 59.0 & 15.8 & 0.07 & 3.30 \\
omni & known & 6 & 100.0 & 0.16 & 63.8 & 13.3 & 0.00 & 0.00 \\
omni & perturbed & 6 & 100.0 & 0.87 & 57.0 & 22.3 & 1.00 & 10.00 \\
omni & refined & 6 & 100.0 & 0.55 & 63.1 & 13.9 & 0.21 & 8.20 \\
\bottomrule
\end{tabular}
\end{table*}

In Table~\ref{tab:supp_rig}, Sphere denotes spherical coverage, Complete denotes the share of the ground-truth surface within the evaluation threshold, Ghost denotes the share of Gaussians beyond the threshold, and Error denotes the remaining extrinsic calibration error of the rig.

In three MP3D houses with agents on different levels, synthetic absolute barometers assign floors with accuracy $1.000$, $0.956$ and $0.990$.
A shared altitude reference cuts the height error between agents from $3.42$, $2.79$ and $4.04$\,m to $0.09$, $0.08$ and $0.12$\,m, and raises cross-agent same-floor accuracy from $0.500$, $0.336$ and $0.870$ to $1.000$, $0.796$ and $1.000$.
In one house, the full configuration (floors, physical checks and shared reference) accepts no link, whereas the floor estimator alone accepts $12$ links.

\section{Evaluation Protocol}
\label{sec:supp_review_protocol}
\paragraph{Data and Splits.}
Replica-8~\cite{straub2019replica,sucar2021imap} evaluates single-agent reconstruction. MA-Replica~\cite{hu2023cpslam} and three-agent Matterport3D walks~\cite{chang2017matterport3d} evaluate collaboration, replaying all agents in one process, interleaved in time.
ScanNet++ iPhone sequences~\cite{yeshwanth2023scannet} provide LiDAR depth, real IMU data, recorded ARKit poses and laser-registered ground truth over the first 2000 frames. All barometers, and all inertial streams outside ScanNet++, are synthetic, so the studies that use them test mechanisms rather than real multi-phone sensing.
Synthetic IMU samples ($200$\,Hz) follow an error model calibrated on $15$ real iPhone sequences of the ADVIO dataset~\cite{cortes2018advio}, with white noise of $6.7\times10^{-4}$\,rad\,s$^{-1}/\sqrt{\text{Hz}}$ for the gyroscope and $3.5\times10^{-3}$\,m\,s$^{-2}/\sqrt{\text{Hz}}$ for the accelerometer, bias random walks of $5\times10^{-5}$ and $5\times10^{-4}$ in the same units per second, turn-on biases of $\sigma=0.005$\,rad/s and $0.05$\,m/s$^2$, and a $0.5\,\%$ scale error per axis.
Synthetic barometers ($0.9$\,Hz) add $0.013$\,m white noise. Under a shared altitude reference, each agent's pressure follows its height in one building frame plus a per-device drift ($0.01$\,m/$\sqrt{\text{s}}$), because phones in the same building share the same weather. Under per-walk zeros, each agent's altitude instead starts at zero.
Each dataset serves non-commercial research under its own terms: the Replica Research License, the Matterport3D data-use agreement, the Microsoft Research license for 7-Scenes, CC~BY~4.0 for TUM RGB-D~\cite{sturm2012benchmark} and the non-redistributable ScanNet++ terms. We redistribute no raw ScanNet++ or Matterport3D data.
We tune on the DEV split (three scenes) and make design decisions on the VAL split (six scenes). Three held-out phone splits (TEST, TEST2, TEST3; $11$, $12$ and $12$ reported scenes) contain disjoint scenes from the official validation partition. Each split was evaluated once, with the configuration frozen at that time; only TEST3 tests the final configuration.
The collaboration studies use Replica room0/office3 (DEV), office0/office1 and four Matterport3D houses (VAL), and eight Matterport3D houses unseen in memory training. The unseen houses were evaluated with the first frozen configuration; an earlier version of the system had run once on them.
The memory trains on $29$ other Matterport3D houses, Replica room0, room1, office0, office1 and office2, five 7-Scenes scenes (chess, fire, office, pumpkin, redkitchen) and $29$ ScanNet++ scenes. It selects its epoch on the four VAL houses, Replica office3, 7-Scenes stairs and the six VAL phone scenes.
The eight unseen houses and all TEST, TEST2 and TEST3 phone scenes are disjoint from both sets. The Replica and 7-Scenes collaboration scenes, however, are training scenes, so those studies test verification and gating, not the generalization of the memory.

\paragraph{Operating Points and Metrics.}
Online is causal reconstruction without refinement. Deterministic runs complete all earned mapping work between frames, so an Online result alone is not a real-time claim. Refined adds 30\,000 iterations (Replica) or 1000 mapping jobs (phones) that also optimize poses, degree-2 spherical harmonics (SH) and an affine exposure model. Paced feeds frames in real time at $30$\,Hz, and CADENCE-RT is the configuration with a smaller map budget and RGB color. Merge integrates verified relations and peer points; Share also uses aligned peer keyframes.
ATE is the position RMSE (cm) after rigid alignment, over tracked frames, and $^\dagger$ marks coverage below $95$\,\%. The keyframe ATE (ATE-KF) uses every fifth frame per agent. Every method renders every tenth evaluation frame at its own estimated pose, mapping keyframes included, so the image metrics score training views. In Refined, the exposure affine of a frame is interpolated in time from the affines fitted at its neighboring keyframes, not fitted to the frame itself. Photo-SLAM saves its map with its final trajectory, and we report its scores on its own sparse keyframes as separate entries.

\paragraph{Statistics and Baselines.}
World-memory ablations use three seeds and paired cluster-bootstrap intervals. Phone comparisons use paired scene-bootstrap $95$\,\% intervals and treat differences below $0.05$\,dB or $0.05$\,cm as ties. Every SLAM result is a single deterministic run, and false-link bounds are exact Clopper-Pearson intervals.
The criterion fixed before the TEST3 evaluation requires at least ten paired wins out of twelve scenes (one-sided sign test, $p=0.019$), an interval excluding zero, and a mean effect of at least $1$\,dB PSNR, $0.02$ SSIM, $0.03$ LPIPS or $30$\,\% ATE reduction with a lower median.
Phone baselines run the official ORB-SLAM3~\cite{campos2021orb}, GS-ICP SLAM~\cite{ha2024rgbd} and Photo-SLAM~\cite{huang2024photoslam} code on the same replays, together with the recorded ARKit trajectory. Association controls use descriptor retrieval or raw DINOv3 similarity~\cite{simeoni2025dinov3}, followed by the CADENCE verifier.
Inertial ORB-SLAM3 completed $1$ of $6$ VAL scenes, and $4$ of $11$, $3$ of $12$ and $11$ of $12$ scenes on TEST, TEST2 and TEST3. Only TEST3 used a patched adapter, which interpolates the $60$\,Hz IMU to $240$\,Hz and adds numerical guards, so the completion rates mix scene and adapter effects.
Accuracy and cost measurements run on a desktop RTX~5070 and a Linux RTX~PRO~4000, and device latency measurements run on an Apple M5 and an iPhone~11 Pro Max. Published rows keep their own protocols and serve as context rather than as a ranking.

\section{Implementation Details}
\label{sec:supp_impl}

The following subsections give the mechanisms and constants in the order of \cref{sec:method}. Symbols are listed in Table~\ref{tab:supp_notation}, and values that a table already gives are not repeated in the text.

\subsection{Local Visual-Inertial SLAM}
\label{sec:supp_local}
\paragraph{Front End and Initialization.}
Pyramidal Kanade-Lucas-Tomasi (KLT) tracking follows at most 200 Features from Accelerated Segment Test (FAST) corners and keeps a track only if it passes a 1\,px forward-backward check.
The map initializes from depth when at least 25 landmarks have depth in $(0.1,25)$\,m, otherwise from a stereo pair, and otherwise by monocular triangulation; in the monocular case, the tracker trusts translation only once the scale is metric.
The world frame aligns with the gravity direction averaged over the first 20 accelerometer samples (static start). Without an IMU, the world frame is the first body frame and the tracker drops the rotation prior.

\paragraph{Inertial Prediction.}
Equation~\ref{eq:supp_predict} propagates rotation $R$, velocity $\mathbf v$ and position $\mathbf p$ with world-frame gravity $\mathbf g$ and increments $(\Delta R,\Delta\mathbf v,\Delta\mathbf p)$ preintegrated over $\Delta t$ (midpoint rule, first-order bias Jacobians~\cite{forster2017manifold}) as
\begin{equation}
\label{eq:supp_predict}
\begin{aligned}
R^{-}&=R\,\Delta R,\qquad
\mathbf v^{-}=\mathbf v+\mathbf g\,\Delta t+R\,\Delta\mathbf v,\\
\mathbf p^{-}&=\mathbf p+\mathbf v\,\Delta t+\tfrac12\mathbf g\,\Delta t^{2}+R\,\Delta\mathbf p .
\end{aligned}
\end{equation}
Fixed-gain filters correct velocity and biases (velocity 0.5, accelerometer bias 0.02, gyroscope bias 0.01) instead of estimating them jointly.

\paragraph{Pose Refinement.}
Starting from the propagated state, Gauss-Newton solves the IMU-prior pose objective of Eq.~\ref{eq:supp_pnp},
\begin{equation}
\label{eq:supp_pnp}
\begin{aligned}
\min_{T_{WB}}\;&\sum_i w_i\,\rho\big(\|\pi(T_{CB}T_{WB}^{-1}\mathbf X_i)-\mathbf u_i\|\big)\\
&+\lambda_R\big\|\operatorname{Log}\big(R\,{R^{-}}^{\top}\big)\big\|^{2}.
\end{aligned}
\end{equation}
Here $\mathbf X_i$ are world landmarks, $\mathbf u_i$ tracked pixels, $\rho$ a $2$\,px Huber penalty and $w_i=1$ ($0.7$ without initial depth). The solver uses $\lambda_R=10^{7}$ and eight iterations, and the IMU translation enters only through the initial pose.
The optional ARKit prior adds a translation penalty and replaces the motion prediction with ARKit's frame-to-frame motion. The tracker rejects this prior above $3$\,m/s or $500^\circ$/s, or when its displacement differs from the prediction by more than $5$\,cm. A second solve with the ordinary inertial prior replaces the ARKit-based solution when enough visual support disagrees with it by more than $5$\,cm.
A match is an inlier if its reprojection error is below 4\,px. A frame is tracked with at least 25 inliers and enters a degraded state with at least 10; otherwise it relocalizes against the three nearest keyframes. Three lost frames start a new segment.

\paragraph{Depth Registration.}
Equation~\ref{eq:supp_reg} registers dense depth against a keyframe-fused surfel model~\cite{keller2013real} with $1$\,cm voxels,
\begin{equation}
\label{eq:supp_reg}
\min_{T_{WB}}\;\sum_j \chi_j\,\rho\Big(\tfrac{1}{\sigma_p}\,\mathbf n_j^{\top}\big(T_{WB}\,\mathbf q_j-\mathbf m_j\big)\Big).
\end{equation}
In this point-to-plane objective~\cite{chen1992object}, calibrated depth points $\mathbf q_j$ lie in the body frame and surfels $(\mathbf m_j,\mathbf n_j)$ in world coordinates, with confidence $\chi_j$ and residual scale $\sigma_p=5$\,mm.
Weakly constrained directions stay at the prediction, and the registration result replaces the visual pose only if the overlap and residual tests pass and the result lies within 0.30\,m and $6^\circ$ of the prediction.
Point association and the $6\times6$ normal equations run on the selected backend, which is the CPU, Compute Unified Device Architecture (CUDA) or Metal; the solve itself runs on the CPU. While the background thread updates the surface, the tracker uses the previous surface instead of waiting.
On handheld TUM sequences, the tracker adds a local landmark model, a tight inertial window, window bundle adjustment and loop closure. Their corrections move the live state, not the stored keyframe poses.

\paragraph{Keyframes and Queues.}
The tracker creates a keyframe when the median track displacement exceeds 25\,px, fewer than 70\,\% of landmarks remain tracked, or 30 frames pass.
Mapping jobs enter a 256-job priority queue (64 in the low-resource preset) with a 64-job reserve for jobs that the mapper spawns. When the queue is full, a new job evicts the lowest-ranked low-priority job if it outranks that job; otherwise the queue refuses the new job and counts the refusal.
The 64-entry world-observation queue refuses new entries when full.
Since the tracker never reads the map, the pose call takes only constant-time hand-off locks.

\subsection{Observation Reliability}
\label{sec:supp_reliability}
\paragraph{Quality Cues.}
Sharpness uses a re-blur metric~\cite{crete2007blur}, and illumination is a piecewise function of mean intensity. Saturation falls linearly to zero at 40\,\% clipped pixels (intensity $\le5$ or $\ge250$), and the signal-to-noise ratio (SNR) uses a fast noise estimate~\cite{immerkaer1996fast}.
Motion-blur risk is $m=1-\exp(-b/6)$ with $b=\bar f\int_{t_s}^{t_e}\|\boldsymbol\omega(\tau)\|\,d\tau$ over the exposure and mean focal length $\bar f$. The estimator assumes a 10\,ms exposure when the exposure is unknown and sets $m=0.2$ without an IMU.
Texture coverage is the share of textured blocks (Laplacian variance above a threshold) in a $6\times6$ grid, and view quality is the mean cosine between sensed normals and viewing rays. Depth confidence is the valid-depth share times $\min(1,0.05\,\text{m}/\sigma_{\mathrm{depth}})$ (scaled by 0.3 if the depth is not metric), and pose confidence is the visual confidence of the tracker.
The estimator computes quality once per new keyframe, inside the pose call.

\paragraph{Confidence Combination.}
Equation~\ref{eq:method_conf} maps the nine cues to the four confidences of \cref{sec:method_reliability} as
\begin{equation}
\label{eq:method_conf}
\begin{aligned}
c^{p}&=\min\big(y_{\mathrm{cond}},\,y_{\mathrm{pose}},\,1-\tfrac{m}{2}\big),\\
c^{g}&=\max\big(\min(y_{\mathrm{sharp}},y_{\mathrm{view}},y_{\mathrm{cond}},1{-}m),\\
&\qquad\quad y_{\mathrm{depth}}(1{-}m)\big),\\
c^{a}&=\min\big(y_{\mathrm{sharp}},\,y_{\mathrm{illum}},\,y_{\mathrm{sat}},\,y_{\mathrm{snr}},\,1-m\big),\\
c^{w}&=\min\big(\tfrac12(y_{\mathrm{sharp}}+y_{\mathrm{illum}}),\,\max(y_{\mathrm{view}},y_{\mathrm{pose}})\big).
\end{aligned}
\end{equation}

\paragraph{Permissions and Peers.}
The thresholds are $(\theta_p,\theta_g,\theta_a,\theta_w)=(0.3,0.4,0.5,0.3)$.
Each keyframe packet carries its modality confidences. A peer's reliability is a moving average of these capped reports (rate $\nu$, prior $0.5$), not of verification outcomes, and an optional penalty for rejected map views, off by default, lowers geometry reliability.
The confidence of a contribution is the element-wise minimum of the running reliability and its capped report. The receiver takes this minimum only after updating the running reliability, so the initial prior cannot override a degraded current report.
The gate denies a peer keyframe without pose permission before verification.
The baseline policies, equal weights and frame rejection, use the same verification gate, and soft weighting uses the minimum of the four confidences as a single weight.

\subsection{Gaussian Mapping}
\label{sec:supp_mapping}
\paragraph{Representation and Rendering.}
A Gaussian $\mathcal G_n=(\boldsymbol\mu_n,\mathbf q_n,\mathbf s_n,o_n,\mathbf c_n)$ consists of mean, unit quaternion, scale, opacity and color coefficients, with covariance $\Sigma_n=R(\mathbf q_n)\operatorname{diag}(\mathbf s_n)^{2}R(\mathbf q_n)^{\top}$. Equation~\ref{eq:supp_render} composites the depth-sorted projected Gaussians from front to back,
\begin{equation}
\label{eq:supp_render}
\hat C=\sum_i \beta_i\alpha_i\mathbf c_i,\quad
\hat A=\sum_i \beta_i\alpha_i,\quad
\hat D=\frac{1}{\hat A}\sum_i \beta_i\alpha_i z_i.
\end{equation}
Here $\beta_i=\prod_{j<i}(1-\alpha_j)$, $\alpha_i=\min\big(0.99,\,o_i\exp(-\tfrac12\mathbf d_i^{\top}{\Sigma_i'}^{-1}\mathbf d_i)\big)$ and $\Sigma_i'=JR\Sigma R^{\top}J^{\top}+0.3\,I$, with offset $\mathbf d_i$ from the projected mean, projection Jacobian $J$, world-to-camera rotation $R$, camera-axis depth $z_i$ and $2\times2$ identity $I$. The renderer evaluates $\hat D$ only where $\hat A>0$.
A new submap opens when no submap origin of the agent lies within 6\,m. A graph correction $\Delta T$ of the submap origin node updates $T_{W_aS_k}\leftarrow\Delta T\,T_{W_aS_k}$, and the mapper leaves the coordinates of all Gaussians unchanged.

\paragraph{Seeding and Evidence.}
Each $\kappa\times\kappa$ pixel block whose accumulated opacity is below 0.5 or whose depth is off by more than 5\,cm seeds one Gaussian, back-projected with opacity 0.8, the pixel color and a surface-aligned orientation. The per-profile values of $\kappa$, the SSIM weight $\lambda$, the Gaussian cap and the views per job $N_w{+}N_s$ are listed in Table~\ref{tab:supp_profiles}c.
A Provisional Gaussian becomes Verified when the depth of another keyframe agrees within $\min(0.5,\max(0.08,3\sigma_d))$\,m, and becomes Stable after five or more observations.

\paragraph{Objective and Optimizer.}
The mean squared depth loss covers only sensed pixels with rendered opacity above $0.3$.
A zero permission removes the corresponding loss terms of a view; a zero appearance weight, for example, removes both the color-error and the SSIM terms. Permissions do not weight the seeding or validation of the agent's own keyframes.
Sparse Adam (moment decay rates $0.9$ and $0.999$) updates position, color, opacity, log-scales and a rotation increment. In the Replica profile, the degree-$l$ SH coefficients of a Gaussian start after $(l-1)\times30$ optimization steps of that Gaussian.
The render width starts at 320\,px and doubles every two jobs up to a maximum of 1200\,px.

\paragraph{Credit Scheduling.}
Each frame earns 3 iterations of credit, capped at 30 frames' worth, and the scheduler drops and reports the excess.
A job uses the $N_w$ newest keyframes plus $N_s$ older views, which the scheduler samples without replacement by the importance $\varphi_k$ of Eq.~\ref{eq:supp_importance},
\begin{equation}
\label{eq:supp_importance}
\varphi_k=\ell_k/\bar\ell+(1+n_k)^{-1}+0.5\cdot2^{-a_k/20}+0.05,
\end{equation}
with the last loss $\ell_k$ of the keyframe, mean loss $\bar\ell$, refinement count $n_k$ and age $a_k$ in keyframes.
Deterministic mode spends the whole credit in the tick that earned it, whereas paced mode spends at most the credit, and only in idle time.

\paragraph{Gaussian Budget.}
When the Gaussian count exceeds a cap after a job, the mapper evicts Gaussians down to 95\,\% of the cap: Provisional before Verified before Stable, and lowest opacity first.

\subsection{World Memory and Training}
\label{sec:supp_wcm}
\paragraph{Query and Network.}
The 95-d query of \cref{sec:method_wcm} holds the region token (32) and a presence flag (1), region-relative position (3), orientation (6), quality cues (9), confidences (4), inertial increments with uncertainties (11), altitude and gravity evidence (5, Eq.~\ref{eq:method_sensor_relative}), a floor hypothesis (2), evidence masks (6) and 16 similarity statistics against the committed embeddings. These statistics provide a bounded summary of the region history in place of raw embeddings, and the adopted model zeroes the inertial block.
The perceptron has $33{,}702$ parameters and outputs six logits and the 32-d increment. Its static/dynamic head is reserved for dynamic scenes; it is untrained and inactive here, because all evaluated scenes are static.
The embeddings are DINOv3 Vision Transformer (ViT-S/16)~\cite{simeoni2025dinov3} class-token (CLS) and generalized-mean (GeM) pooled features of the gray image at a 448\,px long side (768 values, float16), computed offline.

\paragraph{Region Memory and Scheduling.}
Regions form a grid centered on the origin of the agent.
The first query of a region returns no prediction, because the heads compare against a memory that is still empty. A full region still answers queries, and the memory drops a peer increment that lags more than eight proposals behind.
Pose confidence below 0.5 triggers a query. Once a peer is active, every own keyframe also issues a query, at most $\max(1,\lfloor f_{\mathrm{wm}}\rfloor)$ times per tick at world-model rate $f_{\mathrm{wm}}$ (\cref{tab:supp_levels}), and a peer announcement triggers one query per candidate region.
Queries run on the CPU, and a Core ML export of the network passes a parity test.

\paragraph{Training Data and Optimization.}
Training uses $2$\,m regions from Matterport3D~\cite{chang2017matterport3d}, Replica~\cite{straub2019replica}, 7-Scenes~\cite{shotton2013scene} and ScanNet++~\cite{yeshwanth2023scannet}, and each pseudo-agent is a separate tracker run in its own world frame.
The pipeline synthesizes inertial and barometric streams from ground truth except on ScanNet++ (real IMU, no barometer), with a per-sequence peer barometer offset ($\sigma=0.3$\,m).
To cover every alignment state, a peer pose arrives unaligned and masked (40\,\%), coarsely aligned (40\,\%, $20^\circ$/1\,m) or finely aligned (20\,\%, $5^\circ$/0.25\,m).
The memory visit spans four keyframes, and visibility is the co-visible depth share. Floor compatibility is same-level identity on Matterport3D, and an observation counts as reliable if its tracked motion over 0.5\,s lies within 2\,cm and $1^\circ$ of ground truth.
AdamW (learning rate $10^{-3}$, one-cycle schedule, batch 64, weight decay 0.05) trains the network with hidden dropout 0.2, weight averaging (0.999), input noise (0.1), evidence-group dropout (0.3) and log-variances clamped to $[-3,3]$. Training keeps the averaged weights with the best domain-averaged validation correspondence AUROC, here epoch 16 of the adopted run (seed 0). This run stopped after epoch 26 of at most 40 with a patience of 10 epochs.

\subsection{Collaborative Fusion}
\label{sec:supp_fusion}
\paragraph{Messages.}
Besides the fields of \cref{sec:method_fusion}, an announcement carries a place descriptor, pose variances, a floor hypothesis and a 16-d visual descriptor.
An evidence packet holds a gray thumbnail, depth points and FAST features with steered Binary Robust Independent Elementary Features (BRIEF) descriptors~\cite{calonder2010brief} (contents and sizes in Table~\ref{tab:supp_comm}a). The codec computes these descriptors in integer arithmetic, so every platform obtains identical bits.
For each peer, the system estimates only a clock offset from the request-response timestamps of its messages.

\paragraph{Candidates, Proposal and Gate.}
Embedding distance is $1-\cos$. Without embeddings, a descriptor screen ($\le0.55$) replaces the $0.45$ embedding screen. After alignment, up to two keyframes near the peer pose (latest relation) join the eight retrieved ones. In addition, for $10\,\%$ of peer keyframes the verifier checks the best candidate even below $p^{\mathrm{corr}}=0.5$ (exploration, logged).
The proposal uses mutual Hamming matches (ratio 0.8), 200 iterations of three-point RANSAC without scale~\cite{umeyama1991least} and PnP RANSAC at 3\,px, and requires 12 inliers.
The verifier takes medians at full resolution and requires its pose uncertainty to be below 0.5\,m, and a test that cannot be evaluated does not block a relation. Local keyframes count as fully confident, and memory deferral is off (threshold zero) in every reported run.
The height check uses the barometric altitude $h=44330.77\,(1-(p/101.325)^{0.190263})$ (pressure $p$ in kPa). Pairwise consistency rejects a relation that agrees (within 0.5\,m, $5^\circ$) with fewer than half of the accepted alignments of the peer~\cite{mangelson2018pairwise}.

\paragraph{Pose Graph and Map Integration.}
The graph solves Eq.~\ref{eq:supp_graph} over accepted relations and node priors,
\begin{equation}
\label{eq:supp_graph}
\min_{\{T_n\}}\sum_{(i,j)}\varrho\big(\|\sqrt{w^{p}_{ij}}\,\Lambda_{ij}^{1/2}\mathbf r_{ij}\|\big)+\sum_n\big\|\Lambda_0^{1/2}\mathbf r_n\big\|^2,
\end{equation}
with $\mathbf r_{ij}=[\operatorname{Log}(\bar R_{ij}^{\top}\hat R_{ij});\,\hat{\mathbf t}_{ij}-\bar{\mathbf t}_{ij}]$ comparing measured $(\bar R_{ij},\bar{\mathbf t}_{ij})$ and current $(\hat R_{ij},\hat{\mathbf t}_{ij})$ relative poses, information 100 (originated) or 50 (mirrored), $w^{p}_{ij}\in[0.05,1]$ and Huber kernel $\varrho$ ($\delta=0.5$).
The first local node fixes the gauge, and later nodes carry priors $\mathbf r_n$ (5\,cm and 0.035\,rad, or 5\,mm and 0.005\,rad in Replica and multi-agent runs). Without odometry or peer-to-peer factors, the graph only aligns peer frames and submap origins.
Levenberg-Marquardt runs at most eight iterations and 30\,ms per tick.
Merge inserts verified peer points that lie more than 5\,cm from existing geometry in gray, with opacity $0.1\,w^g$ and a position uncertainty that grows as $1/\sqrt{w^{g}}$.
Share also places peer keyframes at the medoid of up to 256 accepted relations and uses them at appearance weight $0.1\,w^a$, and memory updates require $w^w$.

\subsection{Controller}
\label{sec:supp_controller}
At every tick, the controller sets $\ell_t=\max(\ell_{\mathrm{lat}},\ell_{\mathrm{q}},\ell_{\mathrm{mem}},\ell_{\mathrm{th}},\ell_{\mathrm{bat}},\ell_{\mathrm{wm}})$, the highest level that any pressure demands, without hysteresis. Each level adds its action to those of the lower levels (\cref{tab:supp_levels}).
The latency ratio $r_t$ is the p95 of the last 128 pose times divided by $\tau^\star$, and the backlog counts jobs that are not low priority. The controller defers shed appearance jobs instead of discarding them.
A unit test enforces that the policy cannot change the tracker, the verification thresholds, the iterations per job or the objective. Of the viewer, residency and network budgets in \cref{tab:supp_levels}, the engine applies only the network budget itself and passes the viewer rate and residency target to the app.

\begin{table}[htb]
\centering
\footnotesize
\renewcommand{\arraystretch}{0.92}
\caption{Controller levels with their triggers and the action that each level adds.}
\label{tab:supp_levels}
\setlength{\tabcolsep}{3pt}
\begin{tabular}{@{}c>{\raggedright\arraybackslash}p{0.33\columnwidth}>{\raggedright\arraybackslash}p{0.51\columnwidth}@{}}
\toprule
$\ell$ & Trigger & Added action \\
\midrule
1 & thermal fair; $r_t>0.8$ & Viewer budget $30\to7.5$\,Hz \\
2 & backlog $\ge64$ jobs; low battery; world-model backlog $>8$ & World-model queries $2\to0.5$\,Hz (one query per tick) \\
3 & $r_t>1.0$; backlog $\ge256$; memory pressure $\ge0.8$ & Appearance refinement and densification deferred; refinement becomes geometry-only \\
4 & thermal serious; $r_t>1.2$ & Mapping jobs per tick $4\to1$ \\
5 & $r_t>1.5$; memory pressure $\ge0.95$ & Learned inference off; residency budget $200\,\text{k}\to100\,\text{k}$ Gaussians \\
6 & thermal critical & Background synchronization off; network budget $2\to0.2$\,MB/s \\
\midrule
fixed & -- & Tracker configuration, verification thresholds, iterations per job, loss weights, SH degree, seeding density \\
\bottomrule
\end{tabular}
\end{table}

In Table~\ref{tab:supp_levels}, a level applies when any of its triggers holds, $r_t$ is the latency ratio, and the fixed row lists the settings that no level changes.

\subsection{Backend Parity, Arithmetic and Refusals}
\label{sec:supp_backend}
Accelerated work passes through two interfaces, each with a CPU reference: the optimizer (forward pass, fused loss and backward pass with SSIM, sparse Adam) and registration (linearizing \cref{eq:supp_reg}).
The test suite also checks the references themselves: the tiled CPU rasterizer matches a per-pixel renderer (image error below $2\cdot10^{-5}$, depth below $10^{-4}$, gradient relative $L_2$ below $10^{-4}$, maximum relative gradient error below $10^{-3}$), and the reference backend is bit-identical to the CPU optimizer.
Every build tests each CUDA and Metal kernel against the bounds of \cref{tab:supp_parity} (Metal also on the phone). To confirm that the test can fail, a deliberately perturbed Adam step must exceed the Metal bound by more than $100\times$.

\begin{table*}[t]
\centering
\footnotesize
\renewcommand{\arraystretch}{0.9}
\caption{Parity bounds of the test suite for each accelerated operator against its CPU reference on CUDA and Metal.}
\label{tab:supp_parity}
\begin{tabular}{@{}>{\raggedright\arraybackslash}p{0.13\textwidth}>{\raggedright\arraybackslash}p{0.15\textwidth}>{\raggedright\arraybackslash}p{0.31\textwidth}>{\raggedright\arraybackslash}p{0.33\textwidth}@{}}
\toprule
Operator & Compared quantity & CUDA bound & Metal bound \\
\midrule
Forward composite & image; pair count & mean $|\Delta|<10^{-5}$, ${\le}0.1\,\%$ of pixels $>10^{-4}$; equal & same as CUDA \\
Loss & photometric $+$ depth & rel.\ $\le10^{-5}$ & rel.\ $\le10^{-4}$ (SSIM term $\le10^{-5}$) \\
Backward & gradients & rel.\ $L_2<10^{-3}$ & rel.\ $L_2<2\cdot10^{-3}$ \\
SSIM & loss; gradient & resident jobs$^\dagger$: loss $<2\cdot10^{-3}$, position $<2\cdot10^{-5}$\,m & loss rel.\ $<10^{-5}$; rel.\ $L_2<10^{-4}$, $\max|\Delta|\le10^{-4}\max|g|$ \\
Sparse Adam & parameters, moments & moments rel.\ $<10^{-12}$; values $\le10^{-6}$ abs. & rel.\ $\le4\cdot10^{-6}$ (SH degree 0--3) \\
Mapping job & losses; parameters & first loss $<10^{-3}$, last $<5\cdot10^{-2}$; position $<2$\,mm, color $<2\cdot10^{-2}$ & first loss $\le10^{-3}$; last $\le5\cdot10^{-2}$ (RGB, SH1), $\le10^{-3}$ (SH2, SH3); SH rel.\ $L_2\le3\cdot10^{-2}$ \\
Registration & sums; counts; pose & rel.\ $10^{-9}$; equal; $10^{-7}$\,m, $10^{-5}{}^\circ$; run-to-run identical & rel.\ $10^{-4}$; equal; $10^{-5}$\,m, $10^{-3}{}^\circ$; run-to-run identical \\
\bottomrule
\end{tabular}
\end{table*}

In Table~\ref{tab:supp_parity}, rel.\ denotes the normwise relative error, rel.\ $L_2$ denotes the relative $L_2$ error per gradient group, and abs.\ denotes the absolute error. The dagger marks the CUDA entry, because this backend checks SSIM inside whole resident-path jobs.

\begin{table*}[t]
\centering
\footnotesize
\setlength{\tabcolsep}{4pt}
\renewcommand{\arraystretch}{1.08}
\newcommand{\subhd}[1]{\textcolor{groupink}{\textit{#1}}}%
\caption{iPhone app profiles with (a)~memory tiers, (b)~depth sources and (c)~thermal governor levels.}
\label{tab:supp_apple}
\begin{tabular}{@{}>{\raggedright\arraybackslash}p{0.13\textwidth}>{\raggedright\arraybackslash}p{0.25\textwidth}>{\raggedright\arraybackslash}p{0.27\textwidth}>{\raggedright\arraybackslash}p{0.29\textwidth}@{}}
\toprule
\groupband{4}{(a) Memory tiers}
\subhd{Device memory} & \subhd{Gaussian cap} & \subhd{Residency window} & \\
$\le4.5$\,GiB & $150$\,k & $4$ & \\
$\le7$\,GiB & $250$\,k & $6$ & \\
$>7$\,GiB & $400$\,k & $8$ & \\
\groupband{4}{(b) Depth sources}
\subhd{Source} & \subhd{Selected when} & \subhd{Color} & \subhd{Depth} \\
LiDAR & ARKit offers scene depth in a supported 4:3 format & 4:3 from the wide camera, $1920\times1440$ (else $1280\times960$, else $640\times480$), up to 60\,fps & $256\times192$, resampled to $640\times480$ in millimeters, valid $0.05$--$20$\,m \\
Stereo & no LiDAR; rear dual-wide, else dual, else triple camera & focus and white balance lock after 1\,s & widest format ($320\times240$ on the iPhone~11 Pro Max), platform filtering off; relative accuracy only (iOS label), not rescaled \\
None & no depth camera & 30\,fps & none: tracking only, monocular initialization \\
\groupband{4}{(c) Thermal governor}
\subhd{Thermal state} & \subhd{Level} & \multicolumn{2}{@{}l}{\subhd{Action}} \\
nominal, fair & normal & \multicolumn{2}{>{\raggedright\arraybackslash}p{0.57\textwidth}@{}}{--} \\
serious, or Low Power Mode & reduced & \multicolumn{2}{>{\raggedright\arraybackslash}p{0.57\textwidth}@{}}{frame stride $+1$ ($30\to20$\,Hz with LiDAR); viewer 15\,fps at 0.75 scale} \\
critical & paused & \multicolumn{2}{>{\raggedright\arraybackslash}p{0.57\textwidth}@{}}{viewer and map uploads stop; controller receives the critical state; tracking and recording continue} \\
\bottomrule
\end{tabular}
\end{table*}

In Table~\ref{tab:supp_apple}, device memory sets the tier in part (a). In part (c), the governor escalates at once and relaxes one level per $30$\,s of cooler readings.

\paragraph{Arithmetic.}
Both backends keep results independent of thread scheduling.
On CUDA, the optimizer kernels compile without fused multiply-add, Adam runs in double precision, and the backward pass reduces in fixed-order trees without atomics, with losses and per-splat sums in double precision.
Apple GPUs lack 64-bit floats, so Metal runs in single precision with fixed-order reductions and safe math (no reassociation), while the host adds the partial sums in double precision.
Registration resolves surfel association by atomic minima over packed depth-index keys and remains run-to-run deterministic on both backends.
Without compensated summation, the backends agree within the bounds rather than exactly.

\paragraph{Refusals.}
A backend that cannot compute a term of \cref{eq:method_loss} refuses the job and leaves the map unchanged instead of silently dropping the term. Metal refuses joint pose refinement and the optional $L_1$ color loss, and the per-view path of CUDA refuses SSIM. No reported run uses the $L_1$ loss, and Eq.~\ref{eq:method_loss} uses squared error throughout.
Only registration falls back (CUDA, then Metal, then the CPU reference), because all three backends meet the same bounds.

\subsection{Deterministic and Paced Modes}
\label{sec:supp_modes}
A run $(T_{1:N},\mathcal G)=\Phi(x_{1:N};\theta,\sigma,b)$ depends on sensor input $x$, configuration $\theta$, schedule $\sigma$ (which background job runs after which packet) and backend $b$; the tracker does not access the Gaussian map.
Deterministic mode makes the schedule a function of input and configuration, $\sigma=\sigma(x,\theta)$: without a background thread or pacing, one tick of four mapping jobs follows each packet, the pose graph runs eight iterations without a time limit, the controller stays at level 0, and the run ends after three idle ticks (at most $10^{5}$).
As a result, poses and maps are bit-identical on one binary, and a test checks this property over 160 frames under simulated critical pressure. Across machines, we compare results empirically and use tracking-only runs to exclude Gaussian optimization.
Paced mode releases frame $t$ at the wall time of the first frame plus its timestamp offset and ticks the background thread every 10\,ms. The benchmark counts frames that arrive more than 10\,ms late, and a run lagging more than 200\,ms is invalid.
For example, with 25\,ms jobs, a 20\,ms live budget completes one job per tick, whereas deterministic mode completes four.
Reported p95 values cover all frames of a run, and controller-level shares are per-frame histograms.

\subsection{Apple Deployment}
\label{sec:supp_apple}
\paragraph{Metal.}
The portable C++20 core has no platform headers; CUDA and Metal form backend layers, and a C interface serves live and replay sessions to the iOS app.
The Metal optimizer projects, sorts and bins Gaussians on the CPU and runs compositing, SSIM, the backward pass and Adam on the GPU, encoding each pass into one command buffer with a single wait.
The app caches compiled pipelines on disk and rebuilds them when they are stale or corrupt, and a test checks that every cache state yields the same result. At launch, the app runs a GPU self-test that also warms the cache.

\paragraph{Capability Profile and Tiers.}
The app probes capabilities rather than the model name and picks the first available depth source in the order of \cref{tab:supp_apple}b. The Metal optimizer requires Apple GPU family 6 (A13) or newer.
Memory tiers (\cref{tab:supp_apple}a) change budgets only and leave resolution, objective and thresholds unchanged, although eviction under a cap changes which Gaussians exist.
The CPU optimizer uses $\max(2,\text{cores}-2)$ threads.
In automatic mode, an on-device benchmark of under 10\,s times short CPU and Metal calls on a seeded job, predicts a full job on each backend and picks the faster one. The app does not run this benchmark mid-session, when hot or in Low Power Mode. Without a benchmark record, GPU family 6 or newer selects Metal, and a forced setting overrides both rules.
For refused jobs, the app also keeps a retry profile without SSIM and with RGB color; because it changes the objective, it provides no evidence of backend parity.

\paragraph{Sensors, Memory, Thermal State and Viewer.}
Stereo depth provides relative accuracy only (\cref{tab:supp_apple}b), so live stereo sessions serve as feasibility checks; quantitative phone results come from recorded LiDAR sequences~\cite{yeshwanth2023scannet}.
Core Motion provides the IMU (100\,Hz) and barometer on the host clock.
A watchdog reports memory use to the controller every second and, at 0.90 of the limit, ends the session with results saved.
Input decimation by the thermal governor (\cref{tab:supp_apple}c) is the only adaptation that changes the tracker input, and this decimation is inactive in every evaluated replay.
The viewer draws Gaussians as instanced GPU quads at 30\,fps (60 during gestures, lower under load or heat as in \cref{tab:supp_apple}c) and scales resolution to about 10\,ms of GPU time per frame.

\subsection{Verification and Multi-View Settings}
\label{sec:supp_round2_settings}
Besides the inlier and uncertainty limits of \cref{sec:supp_fusion}, the verifier uses the geometric thresholds $\tau_r$ and $\tau_d$ (Eq.~\ref{eq:method_verify}, values in Table~\ref{tab:supp_notation}), at least eight samples per median and a pose-confidence floor of $0.2$.
Optional physical limits are $3^\circ$ gravity tilt and $1.5$\,m disagreement between geometric and barometric height under a shared altitude reference.

For multi-view linking, the sender uses only features tracked over keyframes within $\pm3$\,s, seen in at least three views and with reprojection error below $1.5$\,px in every view.
SIFT keypoints with hashed RootSIFT descriptors use the same $256$-bit packet format as single-frame binary features.
The forward and reverse pose estimates, expressed in the same direction, must agree within $0.30$\,m and $5^\circ$, and the multi-view path weights the gyroscope rotation prior $100\times$ more than the default path to stabilize the rotation estimate.

\subsection{Configurations}
\label{sec:supp_profiles}
All configurations share the tracker of \cref{sec:method_local}, plus profile-specific additions, and run at most ten iterations per mapping job. Scheduled profiles use the credits of \cref{sec:supp_mapping}, whereas the default ``CPU mapping'' configuration has neither credits nor depth registration.
Collaboration runs (Solo, Merge or Share; \cref{sec:supp_review_protocol}) replay all agents in one process over a loopback transport. Learned ranking is active whenever a model is loaded, and each experiment sets the physical checks, the floor estimator with shared altitude and the blur handling.
The device measurements use the configurations of Table~\ref{tab:supp_device_runtime}c on an Apple M5 MacBook Air (passively cooled, 8-core GPU, 16\,GB), an iPhone 11 Pro Max (A13, 4\,GB, no LiDAR) and a desktop RTX~5070 as CUDA reference.
Each run stores its configuration, power source and thermal state at start and end, so that it can be repeated.

\section{Scope and Protocol Qualifications}
\label{sec:supp_scope}
Published numbers serve only as context, because each was obtained under that system's own protocol.
On Replica-8, the eleven published RGB-D systems report a PSNR of $34.11$ to $44.87$\,dB, against our $38.50$\,dB. On MA-Replica, published collaborative Gaussian systems report a merged ATE-KF of $0.14$ to $0.25$\,cm and a PSNR of $34.26$ to $41.35$\,dB, against our $0.35$\,cm and $37.26$\,dB (Table~\ref{tab:supp_extended_results}).
The map-level depth check is not part of the default system; its improvement under noisy sent poses is concentrated in office0, and it misses IMU-corrupted peer poses (\cref{sec:supp_consistency}).
The memory is trained on $2$\,m regions but runs on $5$\,m regions at test time (sensitivity study in \cref{sec:supp_region}).
Four of the eight unseen houses remain unlinked under every policy, the MA-Replica map caps were chosen on the same scenes on which they are evaluated, and every SLAM result comes from a single deterministic run.
Finally, some Replica and 7-Scenes collaboration scenes are part of the memory training data, whereas the unseen Matterport3D houses and the held-out phone scenes are excluded from it (\cref{sec:supp_review_protocol}).

{\small\putbib[main]}
\end{bibunit}

\end{document}